\documentclass[11pt, letterpaper, onecolumn]{tau-class/tau}

\newif\ifinternal
\internalfalse

\usepackage{amsmath}
\usepackage{amssymb}
\usepackage[english]{babel}
\usepackage{graphicx}
\usepackage{xspace}
\usepackage{siunitx}
\usepackage{environ}
\usepackage{booktabs}
\usepackage{threeparttable}
\usepackage{array}
\usepackage{makecell}
\usepackage{multirow}
\usepackage{longtable}
\usepackage{placeins}
\usepackage{float}
\usepackage{needspace}

\newcommand{\internalnote}[1]{\ignorespaces}

\newcommand{\placeholdertag}[1]{{\color{red}\bfseries[#1 values]}}
\newcommand{\ppsetstatus}[2]{\expandafter\def\csname ppstatus-#1\endcsname{#2}}
\newcommand{\statustag}[1]{%
  \ifcsname ppstatus-#1\endcsname\csname ppstatus-#1\endcsname\else%
    \ifdefined\ppstatusloaded%
      \errmessage{statustag: no figure named `#1' (keys come from figures/render.py: plots.RENDERERS and tables.TABLES)}%
    \else\placeholdertag{no data}\fi\fi}
\newcommand{\ppvariant}{-public}
\InputIfFileExists{generated/status.tex}{\def\ppstatusloaded{}}{}

\newcommand{\pearl}{\textsc{Pearl}\xspace}
\newcommand{\ruby}{\textsc{Ruby}\xspace}
\newcommand{\genesispotency}{\texttt{GenesisPotency-2026}\xspace}
\newcommand{\boltz}[1]{\mbox{Boltz-#1\xspace}}

\newcommand{\nesso}{Nesso-1\xspace}
\newcommand{\ligunity}{LigUnity\xspace}
\newcommand{\aqaffinity}{AQAffinity\xspace}

\newcommand{\gnina}{GNINA\xspace}

\newcommand{\partnercohort}{Genesis--Incyte\xspace}

\newcommand{\openfe}{OpenFE\xspace}
\newcommand{\fepfour}{FEP+\,4\xspace}
\newcommand{\caspsixteen}{CASP16\xspace}
\newcommand{\openbind}{OpenBind\xspace}
\newcommand{\nef}[1]{nEF@#1\%\xspace}
\newcommand{\bedroc}{BEDROC\xspace}
\newcommand{\pX}{p$X$\xspace}

\newcommand{\genesisname}{Genesis Molecular AI\xspace}
\newcommand{\genesislogo}{genesis-logo-new.png}
\newcommand{\logoheight}{1.5cm}

\titleformat{\section}{\color{black}\sffamily\large\bfseries}{\thesection.}{0.5em}{#1}[]
\titleformat{name=\section,numberless}{\color{black}\sffamily\large\bfseries}{}{0em}{#1}[]

\makeatletter
\apptocmd{\ps@firststyle}{\fancyhead[R]{\includegraphics[height=\logoheight]{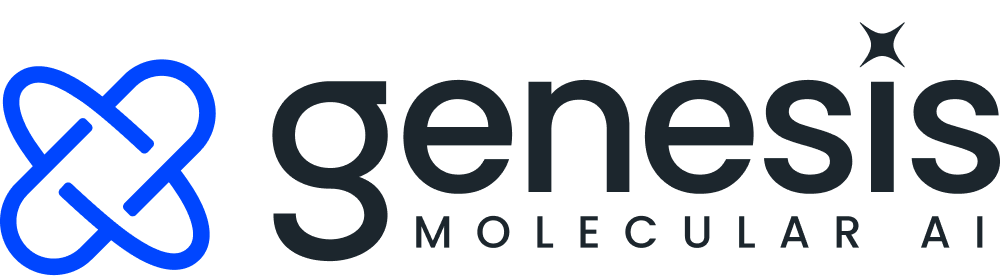}}}{}{}
\makeatother

\journalname{\genesisname\ -- Technical Report}
\renewcommand{\titlepos}{\raggedright}
\title{\ruby: State-of-the-Art Potency Prediction across Discovery Regimes Using Learned Structural Representations}
\author[]{Genesis Research Team}
\professor{\internalnote{}}
\institution{\genesisname}
\footinfo{\ruby}
\theday{November 2026}
\leadauthor{Genesis Research Team}

\renewcommand{\absfont}{\selectfont\small\color{black}\sffamily\bfseries}

\begin{abstract}
Small-molecule drug discovery involves repeated rounds of designing, making, testing, and analyzing compounds.
The goal is often to maximize potency while keeping other properties within the ranges required for a viable therapeutic candidate.
Compound prioritization places different demands on computational methods as drug discovery progresses from hit identification through hit-to-lead and lead optimization, with changing computational budgets and increasing amounts of on-target experimental data.
We introduce \ruby, a Robust Universal Binding affinitY predictor that uses intermediate representations from the structural foundation model \pearl without requiring an explicit binding pose for every compound.
\ruby is trained on proprietary data plus \genesispotency, an audited public-source potency training dataset that we are releasing for academic and commercial use under a CC BY-SA 4.0 license with row-level provenance and audit annotations.
\ruby achieves the highest aggregate early enrichment across seven public hit-identification targets among evaluated methods, with adjustable inference effort providing a trade-off between enrichment and computational cost over more than two orders of magnitude.
\ruby achieves the highest average correlations across four public benchmarks: \openfe, \caspsixteen, \fepfour, and \openbind.
\ruby significantly outperforms other methods in zero-shot prediction across eight biochemical and cellular assays from five ongoing \partnercohort programs, widening its performance advantage relative to the public benchmarks.
\ruby continues to improve more than other methods when all are given the opportunity to recalibrate using earlier assay measurements.
Together, these contributions provide an annotated training resource for the field and a state-of-the-art potency predictor for broad screening, prioritization within related chemical series, and assay-specific prediction as experimental measurements accumulate.

\end{abstract}

\begin{document}
\begin{hyphenrules}{nohyphenation}
\maketitle
\end{hyphenrules}
\thispagestyle{firststyle}
\tauabstract

\section{Introduction}
\label{sec:intro}

\subsection{Regimes of drug discovery}
The requirements for compound prioritization change with the available experimental data, structural knowledge, and computational budget across hit identification (HitID), hit-to-lead (H2L), and lead optimization (LeadOp).

\paragraph{Hit identification.}
HitID involves scoring enumerated libraries of tens of thousands to millions of compounds or combinatorial spaces up to 100 billion molecules~\cite{enamine2026realspace}.
Predictors must concentrate diverse and active compounds at the top of a ranking so that a small subset can be experimentally tested.
Prioritization at this scale requires computational methods fast enough to operate within practical time and compute budgets.

\paragraph{Hit-to-lead.}
H2L involves expanding a handful of confirmed active compounds into congeneric series around a common core scaffold, for example through R-group substitutions.
Experimental follow-up may involve only tens to hundreds of synthesized compounds, making accurate ranking among closely related candidates important.

\paragraph{Lead optimization.}
LeadOp usually begins after tens to hundreds of compounds have on-target assay measurements that can guide compound exploration and refinement across multiple parameters such as potency, selectivity, permeability, clearance, and bioavailability, as well as synthetic feasibility.
Learning from earlier on-target measurements and refining the feedback loop between prediction and experiment become more important as these properties are balanced against each other.

\paragraph{Assay heterogeneity.}
The experimental quantities predicted also vary across programs.
Binding affinity describes the strength of a protein--ligand interaction, whereas biochemical and cellular potency describe effects measured under particular assay conditions.
While related, these quantities are not interchangeable: an IC$_{50}$ can depend on assay format and conditions, and cellular activity additionally reflects factors such as compound permeability and the biological response being measured.
Even measurements against the same target can therefore differ across assays~\cite{kalliokoski2013mixedic50,postnikova2018cellassays}.
A useful potency predictor must consequently support compound ranking across binding affinity, biochemical potency, and cellular potency, as well as adaptation to the particular assay used to guide a program.

\subsection{Computational tools complement experimental measurement}
Computational methods help chemists decide which compounds to synthesize and test, with different approaches suited to different combinations of available data, structural knowledge, and computational budget.
We briefly summarize the main computational approaches to potency prediction and refer readers to broader reviews for more extensive discussion~\cite{sadybekov2023computational,lam2025structure,wang2025interactions}.

\paragraph{Quantitative structure--activity relationships.}
Quantitative structure--activity relationship (QSAR) models learn associations between molecular structure and measured activity.
Classical approaches use molecular descriptors or fingerprints, whereas neural models such as Chemprop learn representations directly from molecular graphs~\cite{heid2024chemprop}.
These ligand-based models can make inexpensive predictions and learn from a program's measured structure--activity relationships, but their usefulness depends on how well the available training measurements cover the compounds being considered.
Multi-task learning can share information across assays or targets, though transfer to a new target depends on how the model represents that target and on the relevance of the training tasks~\cite{allenspach2024multitask}.

\paragraph{Molecular docking.}
Docking uses an experimental or predicted protein structure to search for ligand conformations and orientations within a binding site.
Conventional methods usually hold the receptor fixed or allow limited flexibility, and they rank candidate poses using approximate scoring functions.
These scores may combine physical interaction terms, empirically fitted contributions, or statistical patterns learned from known complexes~\cite{bender2021docking}.
Docking can prioritize compounds at large scale and provide explicit binding hypotheses for interpreting interactions and designing analogs.
Recent large-library studies show that docking scores can contain useful affinity information, though they can also expose errors and artifacts among highly ranked compounds~\cite{liu2025scale,moesgaard2026sensitivity}.

\paragraph{Binding free energy calculations.}
Alchemical binding free-energy methods use molecular dynamics simulations and force fields to estimate thermodynamic differences between states.
Relative binding free-energy (RBFE) calculations estimate affinity differences by transforming one ligand into another in the bound and unbound states, making them particularly useful for comparisons within a series where chemical transformations are often modest.
Absolute binding free-energy (ABFE) calculations estimate the binding free energy of an individual ligand, allowing comparisons without requiring a closely related reference ligand.
Both approaches require suitable structural models, careful system preparation, and sufficient conformational sampling, and their very high computational cost generally limits the number of compounds that can be evaluated~\cite{muegge2023alchemical}.

\paragraph{Structure prediction.}
An experimentally determined complex is not always available when compound prioritization begins, and even an available protein structure may not capture the conformation relevant to a candidate ligand.
Binding can reorganize side chains and loops, and cryptic pockets may be absent from the observed conformation, complicating the use of a single fixed structure for screening~\cite{lam2025structure}.
Cofolding models such as \pearl predict protein--ligand complexes jointly, providing structural hypotheses for interpreting interactions and guiding compound design~\cite{genesis2025pearl}.

\paragraph{Potency prediction from structural representations.}
A plausible binding pose does not by itself establish whether a compound binds strongly or how it ranks against other compounds.
Boltz-2 addresses this additional task with an affinity module that uses learned pair representations and predicted coordinates~\cite{passaro2025boltz2}, while \nesso, \aqaffinity, and ApherisAffinity also
connect learned structural representations with affinity prediction~\cite{nesso2026,sandboxaq2026aqaffinity,bucher2026apherisaffinity}.
Other approaches include shared pocket--ligand embeddings in \ligunity and coarse structural representations in TerraBind~\cite{ligunity2025,rossi2026terrabind}, and proprietary systems such as IsoDDE report strong affinity prediction performance~\cite{isomorphic2026isodde}.

In \pearl and other cofolding models, intermediate pair representations encode relationships among protein and ligand tokens upstream of the module that generates atomic coordinates.
These representations offer an opportunity to predict potency directly, avoiding explicit pose generation for every compound while retaining information learned through structural training.
Reducing this computation could make structural representations useful across a wider range of screening budgets.

\paragraph{Deploying to partner programs.}
Public potency benchmarks enable reproducible comparisons, but their selected targets, chemical series, and assay endpoints capture only part of the prediction challenge in pharmaceutical discovery.
Ongoing programs pursue novel chemical matter, less-characterized therapeutic targets, or both, while generating measurements under assay conditions specific to their experimental objectives.
Models must therefore transfer learned relationships to new compounds and experimental contexts, making evaluation on active pharmaceutical programs an essential complement to public benchmarks.

Genesis has a strategic AI-focused research collaboration with \href{https://incyte.com/}{Incyte}, building and deploying state-of-the-art AI to accelerate the discovery of novel molecules for collaboration targets selected by Incyte.
The collaboration \href{https://investor.incyte.com/news-releases/news-release-details/incyte-and-genesis-therapeutics-announce-strategic-ai-focused}{started in 2025} and was \href{https://investor.incyte.com/news-releases/news-release-details/incyte-and-genesis-expand-molecular-ai-collaboration-accelerate}{expanded in 2026}.
The expanded agreement is one of the first major Pharma-AI collaborations to power large-scale foundation model training with a partner’s proprietary experimental data.

\subsection{\ruby: Robust Universal Binding affinitY across drug discovery regimes}
We introduce \ruby, a potency prediction model that uses \pearl's intermediate pair representation to predict activity without generating an explicit protein--ligand pose for each compound.
We evaluate its use for early enrichment in HitID, zero-shot potency ranking in H2L, and prediction after assay-specific recalibration and fine-tuning in LeadOp.
These settings test whether the same pretrained predictor can support both broad screening with limited on-target data and focused prioritization as experimental measurements accumulate.

\paragraph{Contributions.}
Our primary contributions are an annotated training corpus for the community and a robust, universal model applicable across discovery regimes:
\begin{itemize}
  \item \textbf{An annotated training dataset for community use.}
    We introduce \genesispotency, an audited public-source training dataset that we are releasing for academic and commercial use under a CC BY-SA 4.0 license with row-level provenance and audit annotations (\S\ref{sec:methods:corpus}).

  \item \textbf{Hit identification.}
    \ruby achieves the highest aggregate early enrichment among the evaluated methods across experimentally tested compounds from seven targets in four CACHE challenges and three published large-scale virtual screens~\cite{cache2022roadmap, herasymenko2026cache3,cachechallenge2026,lyu2019ultralarge, alon2021sigma2,liu2025scale}.
    We examine enrichment together with inference time to characterize its usefulness under different computational budgets (\S\ref{sec:results:hitid}).

  \item \textbf{Zero-shot potency ranking.}
    \ruby achieves leading correlations compared with both open models and established affinity prediction methods across the \openfe, \fepfour, \caspsixteen, and \openbind benchmarks (\S\ref{sec:results:public}).
    This advantage extends to ongoing \partnercohort drug-discovery programs, where \ruby achieves higher aggregate zero-shot correlations than all evaluated comparison methods across eight biochemical and cellular assays on five targets (\S\ref{sec:results:programs}).

  \item \textbf{Adaptation to ongoing discovery programs.}
    Using earlier measurements from each assay, \ruby achieves higher correlation and lower prediction error after rolling-temporal recalibration than the evaluated comparison methods on \partnercohort programs (\S\ref{sec:results:programs}).

\end{itemize}

\section{Methods}
\label{sec:methods}

\subsection{\genesispotency and training data}
\label{sec:methods:corpus}
\label{sec:methods:deleak}

\paragraph{\genesispotency and training sources.}
\ruby is trained on a combination of public-source and proprietary activity measurements.
The public-source portion, \genesispotency, assembles measurements from ChEMBL 37, BindingDB release 202607, and PubChem HTS (2025)~\cite{zdrazil2024chembl,gilson2016bindingdb,kim2025pubchem}.
We define ${\mathrm{p}X = -\log_{10}(X)}$, where $X$ is the numerical value of IC$_{50}$, $K_i$, or $K_d$ expressed in mol/L.
Higher \pX indicates stronger binding or greater potency, depending on the measured endpoint.
We restrict potency regression to IC$_{50}$, $K_i$, and $K_d$ measurements and retain EC$_{50}$ measurements for binary activity labels.
We group measurements by source document, target, and readout, and apply ligand- and assay-eligibility filters described in Appendix~\ref{app:corpus}.
Figure~\ref{fig:genesispotency}(c) summarizes the resulting training composition: 5.41 million public-source potency data points after the audit, compared with 1.42 million in AQAffinity's released training set.
Figure~\ref{fig:genesispotency}(d) breaks them down by drug-target family, and Appendix~\ref{app:coverage} compares target coverage with AQAffinity.
Following IsoDDE~\cite{isomorphic2026isodde}, we restrict \genesispotency to measurements dated before 1 January 2023 and apply the benchmark-specific target-exclusion rules detailed in Appendix~\ref{app:deleak}.

\begin{figure}[!t]
  \centering
  \makebox[\linewidth][c]{\includegraphics[width=1.05\linewidth]{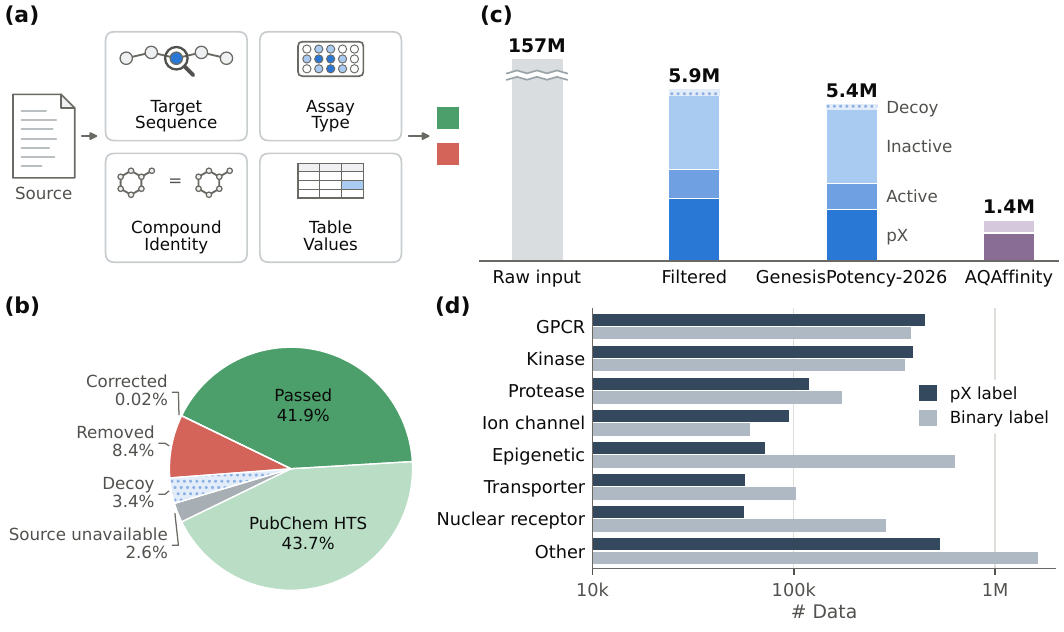}}
  \caption{Audit, composition, and target coverage of \genesispotency.
  \textbf{(a)} Two-pass source-document audit.
  \textbf{(b)} Audit status of the 5.90 million pre-audit potency data points.
  Of the passed data points, 32\% were reviewed against the full text and 68\% against the abstract or supplement only.
  PubChem HTS screening calls have no source document and pass without review.
  \textbf{(c)} Raw input rows (bar truncated) and potency data by label type before and after the audit, with AQAffinity's released training set~\cite{sandboxaq2026aqaffinity}.
  \textbf{(d)} \genesispotency's potency data points by drug-target family~\cite{zdrazil2024chembl}, on a log scale.
  Table~\ref{tab:coverage-groups} lists all target groups.}
  \label{fig:genesispotency}
\end{figure}

\paragraph{Source-document audit.}
Databases can share source records and propagate the same transcription errors, so agreement between databases does not necessarily provide independent confirmation of molecular identities, target assignments, endpoints, or units.
We audited available ChEMBL and BindingDB source documents in two passes (Figure~\ref{fig:genesispotency}(a)): assay-level checks of target identity, experimental context, endpoints, and units, followed by checks of up to 15 sampled compounds per document.
Source-supported corrections and exclusions were applied to the training corpus, with decision rules and handling of unresolved corrections detailed in Appendix~\ref{app:corpus}.
We could audit only the subset of records for which source material was openly accessible.

\paragraph{Audit outcomes.}
Of the 5.90 million pre-audit potency data points, 41.9\% passed source-document review and 8.4\% were removed or withheld (Figure~\ref{fig:genesispotency}(b)).
Appendix~\ref{app:audit-outcomes} reports the outcomes by source.
The audit identified errors that would change the biological meaning of a data point: for instance, the $K_i$ values of 885 FLAP inhibitors from one patent had been assigned to the unrelated protein flap endonuclease 1 in both databases.
In another patent, 10 of 15 sampled records depicted a synthetic intermediate in place of the tested compound (Figure~\ref{fig:audit-examples}).

\paragraph{Annotated release.}
We are releasing \genesispotency for academic and commercial use under a CC BY-SA 4.0 license with row-level provenance and audit annotations\footnote{Dataset and release documentation: \url{https://github.com/genesistherapeutics/genesispotency}.}.
Appendix~\ref{app:corpus} describes the audit coverage, decision rules, and distinction between source records and potency data points.

\subsection{Model}
\label{sec:methods:model}

\ruby predicts protein--ligand activity using learned representations from the structural foundation model \pearl~\cite{genesis2025pearl}.
It takes the same types of inputs: a protein amino acid sequence, a ligand represented as a SMILES string, and optional binding-site information and structural templates.
\ruby uses the \pearl trunk's protein--ligand pair representation without generating an explicit binding pose for each compound.
The model produces a binary activity score for distinguishing active from inactive compounds and a regression estimate on the \pX scale.

\subsection{Inference and conditioning}
\label{sec:methods:conditioning}

We use inference effort to describe the computational budget allocated to scoring each compound.
For hit identification, we evaluate four \ruby settings---Low, Medium, High and Max---using the same trained checkpoint.
These settings vary the computation devoted to protein--ligand representations, allowing a trade-off between early enrichment and inference time.
All four settings score the same protein--ligand pairs without retraining the model or generating explicit binding poses.

\ruby supports individual-compound and compound-list prediction.
In list mode, a compound's predicted potency can depend on the other compounds supplied alongside it, so this distinction affects predictions rather than only computational batching.
The hit-identification evaluation uses no binding-site conditioning or structural templates.
For public potency benchmarks, \ruby automatically selects among candidate input configurations for each target.
Partner-program evaluations use each program's existing pocket conditioning and structural template inputs across all methods.
Appendix~\ref{app:preparation} describes protein inputs, ligand preparation, and structural conditioning.

\subsection{Evaluation sets}
\label{sec:methods:data}

We evaluate \ruby across three complementary settings: hit identification on experimentally tested screening compounds, zero-shot potency ranking on public benchmarks, and prediction within ongoing pharmaceutical programs.
For the public benchmarks, we pursue aggressive train--test separation through both a pre-2023 measurement cutoff and benchmark-specific target exclusions based on protein sequence similarity.
Appendix~\ref{app:deleak} specifies the date and sequence-similarity thresholds, their application to each benchmark, and the corresponding policies reported for comparison methods.

\paragraph{Hit identification.}
We evaluate seven public sets containing 9{,}364 compounds and 1{,}008 positive activity labels: D$_4$, $\sigma_2$, and AmpC from large-library docking campaigns, and LRRK2 WD40, NSP13, Mac1, and CBLB TKB from the first four CACHE challenges~\cite{lyu2019ultralarge, alon2021sigma2,liu2025scale,cache2022roadmap, herasymenko2026cache3,cachechallenge2026}.
Our evaluation reranks the experimentally tested compounds from these campaigns and measures enrichment against their observed activity labels.
Appendix~\ref{app:evaluation} discusses other HitID benchmarks.

\paragraph{Public potency ranking.}
We evaluate 1{,}100 measurements from 64 source series across \openfe, \fepfour, and \caspsixteen (autotaxin and chymase), plus 494 compounds from \openbind~\cite{baumann2026openfe,wang2015fep,chen2023abfe,gilson2026casp16,nelen2026openbind}.
Appendix~\ref{app:public-cohorts} details cohort membership, label sources and conversion to \pX.

\paragraph{Partner programs.}
We evaluate potency prediction on eight biochemical and cellular potency assays across five \partnercohort targets, totalling 22{,}036 assay--compound measurements.
We report zero-shot performance on the full scored cohort and evaluate assay-specific recalibration using a rolling temporal split, building on prior time-split evaluation of industrial assay models~\cite{kearnes2024ordinal}.
At each weekly cutoff, recalibration uses earlier exact measurements from the same assay to predict compounds in a seven-day window beginning 28 days later.
Recalibrated evaluation requires at least 20 exact measurements in the preceding training history.
Base prediction models remain fixed, and comparator scores receive a separate assay-specific affine recalibration at each cutoff.
Appendices~\ref{app:partner-cohorts} and~\ref{app:rolling-temporal} describe cohort eligibility and the rolling temporal split.

\subsection{Metrics and uncertainty}
\label{sec:methods:metrics}
\begin{figure}[t!]
  \centering
  \includegraphics[width=\linewidth]{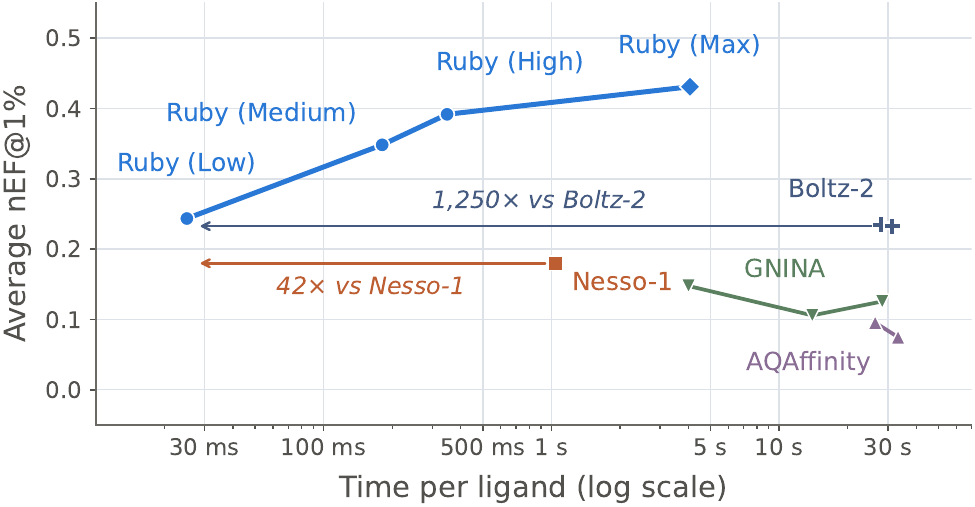}
\caption{Time per ligand versus equal-target average \nef{1} across seven screening targets.
\ruby Low, Medium, High and Max use the same checkpoint, with matched predictions and warmed timing measurements for each setting on Nvidia B200 hardware.
The connecting line identifies \ruby's effort settings.
AQAffinity points use one and five samples, and GNINA points use one, four and nine conformers.
The Boltz-2 line connects 1/1- and 2/5-sample enrichment settings, using RTX PRO 6000 timing for 1/1 and B200 timing measured at 5/5 samples for 2/5.
\statustag{cost}}
  \label{fig:cost}
\end{figure}

\paragraph{Hit identification.}
We measure early enrichment using a chance-corrected variant of the normalized enrichment factor of Liu et al.~\cite{liu2019nef}:
\[
\mathrm{nEF}_{q} = \frac{\mathrm{EF}_{q}-1}{\mathrm{EF}_{\max,q}-1},
\]
where $\mathrm{EF}_{q}$ is the active fraction among the top-ranked fraction $q$ of compounds divided by the active fraction in the full set, and $\mathrm{EF}_{\max,q}$ is its maximum achievable value.
The normalization assigns random ranking an expected score of 0 and the best achievable ranking a score of 1.
We average per-target values with equal weight so that larger screens do not dominate the aggregate.
Figure~\ref{fig:cost} uses \nef{1} for the inference-time comparison, and Figure~\ref{fig:hitid} reports mean nEF across selection fractions from 0.2\% to 10\%.
Appendix~\ref{app:hitid} reports per-target nEF from 0.2\% to 10\% and \bedroc~\cite{truchon2007bedroc}.

\paragraph{Public potency.}
For \openfe, \fepfour, and \caspsixteen, we report Pearson $r$ between predicted and measured potency within each source series, then average within each benchmark weighted by compound count.
We keep the 58 \openfe source series separate.
\fepfour and \caspsixteen each contain one series per target.
We use Pearson $r$ to support comparison with published model results.
For OpenBind, we report Spearman $\rho$ over the full 494-compound cohort, following its published evaluation~\cite{nelen2026openbind}.
Appendix~\ref{app:public} provides additional correlation metrics and detailed comparisons with published results.

\paragraph{Partner programs.}
We evaluate within-assay ranking using Spearman $\rho$ and Kendall $\tau$, linear association using Pearson $r$, and absolute accuracy after recalibration and fine-tuning using mean absolute error (MAE) in \pX units.
We report correlations in both zero-shot and recalibrated settings and MAE only after recalibration.
Correlations include censored measurements at their reported bounds, whereas MAE uses only uncensored measurements.
Aggregate results weight assays equally.
Appendix~\ref{app:partner-cohorts} details eligibility, including the requirement for finite fitted predictions from every method in the recalibrated setting.

\paragraph{Program chemical similarity.}
For the \partnercohort programs, we examine how prediction performance varies with chemical similarity to prior assay measurements.
We define similarity as the maximum Tanimoto similarity between each test compound's nonchiral Morgan fingerprint (radius 3, 2{,}048 bits) and those of compounds with exact measurements available before its rolling cutoff in the same assay.
Within each assay, we pool rolling-temporal recalibrated predictions across all test windows and group compounds by their similarity to measurements available at their respective cutoffs.
We compute Spearman $\rho$ and MAE within each similarity bin, using the same eligible compounds across methods for each metric and averaging assay-level metrics equally.
This analysis asks how performance varies with chemical familiarity, complementing the whole-assay correlations reported above.
Appendix~\ref{app:partner-similarity} gives fingerprint parameters, binning rules, and scoring details.

\paragraph{Uncertainty.}
We estimate uncertainty using 1{,}000 bootstrap resamples of compounds within each target or assay, using identical resampled compounds across methods for paired comparisons.
Error bars show one bootstrap standard error of the reported statistic, reflecting uncertainty from compound sampling with predictions held fixed.
Variability from model training or target selection is not captured by these estimates.
Appendix~\ref{app:bootstrap} provides resampling and aggregation details.

For each metric in the partner-program figures, we compare \ruby with the comparator having the best aggregate point estimate on the same eligible cohort.
Stars summarize one-sided paired bootstrap comparisons, with calculation details in Appendix~\ref{app:bootstrap} and results in Appendix Table~\ref{tab:partner-paired}.

\subsection{Baselines}
\label{sec:methods:baselines}
We compare \ruby with models and docking methods evaluated on the same benchmark compounds, published per-compound predictions re-scored under our evaluation protocol, and published aggregate results.
For published FEP+ and \openfe relative binding free-energy (RBFE) predictions, we use the same eligible compounds, metrics and bootstrap procedure as for the methods we run.
We identify cited aggregate results separately because they do not support matched uncertainty estimates or paired comparisons.
Appendices~\ref{app:published} and~\ref{app:comparators} describe prediction sources, comparator configurations and coverage.

\paragraph{Structure-based comparators.}
We run \boltz{2}~\cite{passaro2025boltz2}, \nesso~\cite{nesso2026}, \ligunity~\cite{ligunity2025} and \aqaffinity~\cite{sandboxaq2026aqaffinity} with their released code and weights, using the evaluation-specific settings documented in Appendix~\ref{app:comparators}.
\ligunity uses its screening checkpoint for hit identification and its ranking checkpoints for potency.

\paragraph{Docking and trivial baselines.}
For our docking runs, we use GNINA~\cite{mcnutt2025gnina13} and rank compounds by negated empirical docking energy.
Molecular weight serves as a size-only control.
Appendix~\ref{app:comparators} specifies receptor preparation, docking settings and coverage.

\paragraph{Cited results.}
For IsoDDE~\cite{isomorphic2026isodde}, ApherisAffinity~\cite{bucher2026apherisaffinity}, TerraBind~\cite{rossi2026terrabind}, and \nesso's OpenBind result~\cite{nesso2026}, we use published summary values because per-compound predictions are unavailable to us for these comparisons.
We evaluate \caspsixteen retrospectively and derive TerraBind's displayed aggregate from its published per-target values under our weighting.
Appendix~\ref{app:public} gives the matching, derivation and sensitivity checks.

\section{Results}
\label{sec:results}

\subsection{Inference cost and hit identification on seven public targets}
\label{sec:results:hitid}

Figure~\ref{fig:cost} shows increasing early enrichment with \ruby's inference effort across the seven public screening targets.
Mean \nef{1} rises from 0.24 at 25.11 ms per ligand for \ruby Low to 0.43 at 4.06 s for \ruby Max.
\ruby Medium and High provide intermediate settings, reaching 0.35 at 180.77 ms and 0.39 at 348.48 ms per ligand, respectively.

\ruby Max achieves a mean \nef{1} of 0.43 versus 0.23 for the \boltz{2} reference setting, and it scores compounds more than 7.5 times faster at the recorded inference rates.
Similarly, \ruby High reaches 0.39 versus 0.18 for \nesso while being 3 times faster.
Even \ruby Low exceeds both methods in mean enrichment while being 1250 and 42 times faster, respectively.

\begin{figure}[t!]
  \centering
  \begin{subfigure}[t]{.37\linewidth}
    \centering
    \includegraphics[width=\linewidth,trim=0 28bp 0 0,clip]{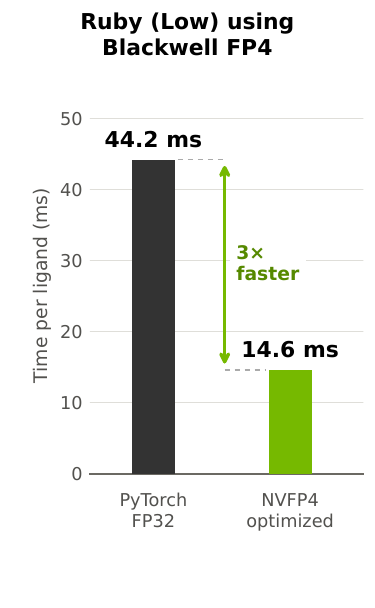}
    \caption{}
    \label{fig:ruby-fp4}
  \end{subfigure}\hfill
  \begin{subfigure}[t]{.60\linewidth}
    \centering
    \includegraphics[width=\linewidth,trim=0 28bp 0 0,clip]{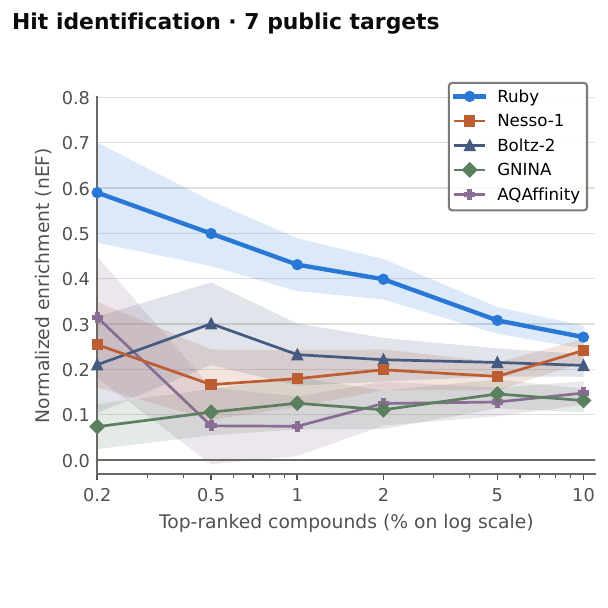}
    \caption{}
    \label{fig:hitid}
  \end{subfigure}
  \caption{Kernel efficiency and early enrichment.
  \textbf{(a)} Nvidia Blackwell kernel optimization at fixed \ruby Low effort on
  Nvidia B200 hardware over $\sigma_2$ compounds.
  The optimized stack combines NVFP4 kernels, Transformer Engine, cuBLAS BF16x9, CUDA Graphs and cuEquivariance attention.
  Bars show mean time per ligand.
  \textbf{(b)} Early enrichment on seven public targets.
  Lines give the equal-target mean nEF, and shading shows $\pm$ one bootstrap SE.
  Each model uses its highest screening setting evaluated here.
  Table~\ref{tab:a1} in Appendix~\ref{app:hitid} gives per-target results.
  \statustag{hitid}}
  \label{fig:efficiency-enrichment}
\end{figure}

Figure~\ref{fig:ruby-fp4} compares native and optimized GPU execution at fixed \ruby (Low) effort using the same checkpoint on Nvidia B200 hardware.
On $\sigma_2$, the optimized kernel stack reduces warm inference time from 44.2 to 14.6 ms per ligand, giving approximately threefold higher throughput at the same model effort.

Figure~\ref{fig:hitid} shows nEF across selection fractions from 0.2\% to 10\% averaged over the seven public screening targets.
\ruby (Max) has the highest mean nEF among the plotted methods at every evaluated selection fraction.
At 1\%, \ruby reaches 0.43 versus 0.23 for \boltz{2} ($p=0.007$), and at 5\%, it reaches 0.31 versus 0.22 ($p=0.011$).
Table~\ref{tab:a1} in Appendix~\ref{app:hitid} reports per-target results across all six selection fractions.

\begin{figure}[t!]
  \centering
  \includegraphics[width=\linewidth]{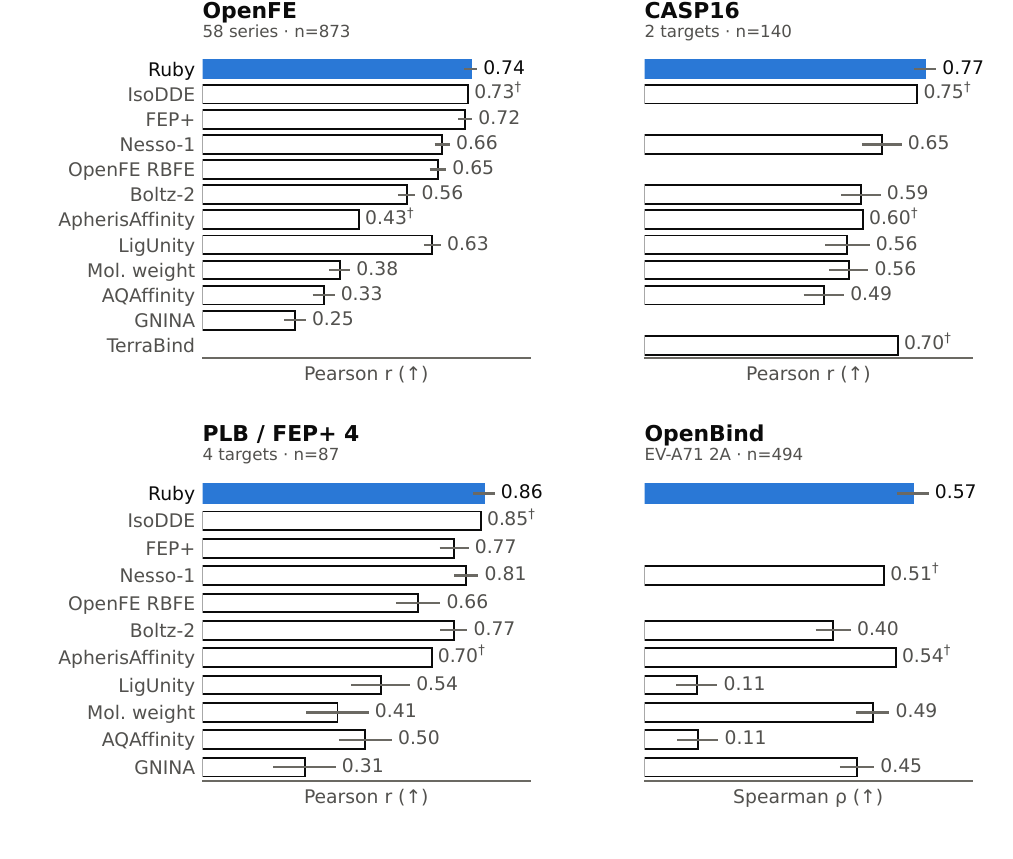}
  \caption{Potency ranking performance over four public benchmarks.
  For re-scored methods, Pearson correlations are computed within the 58 OpenFE source series and within each CASP16 and FEP+4 target, then averaged within each benchmark using compound-count weights.
  OpenBind uses Spearman correlation over its full 494-compound cohort.
  Daggered ($\dagger$) IsoDDE, ApherisAffinity and TerraBind have no released compound-level predictions here, so they have no matched error bars.
  Boltz-2 bars show our reruns, and its original report gives $r=0.62$, $0.65$ and $0.66$ on OpenFE, CASP16 and FEP+4, respectively~\cite{passaro2025boltz2}.
  \statustag{series}}
  \label{fig:series}
\end{figure}

\Needspace{6\baselineskip}
\subsection{Zero-shot potency ranking on four public benchmarks}
\label{sec:results:public}

Figure~\ref{fig:series} compares \ruby's zero-shot potency ranking in list-inference mode with other methods on the \openfe, \caspsixteen, \fepfour, and \openbind benchmarks.
\ruby has the highest average correlation among the methods.
On \openfe, \ruby reaches $r=0.74$ across 58 source series spanning 39 targets, compared to IsoDDE at $0.73$, FEP+ at $0.72$, and surpassing all other methods.
On \caspsixteen, \ruby reaches $r=0.77$, compared to IsoDDE at $0.75$, \nesso at $0.65$, and surpassing all other methods.
On the four-target \fepfour subset, \ruby reaches $r=0.86$, compared to IsoDDE at $0.85$, \nesso at $0.81$, and surpassing all other methods.
On \openbind, \ruby reaches $\rho=0.57$, compared to ApherisAffinity's $0.54$, \nesso's $0.51$, and surpassing all other methods.

Tables~\ref{tab:a2} and~\ref{tab:a2secondary} report per-series and per-target results and additional correlation metrics, while Appendix~\ref{app:public} documents prediction provenance, cohort matching, and sensitivity to aggregation.

\begin{figure}[t!]
  \centering
  \makebox[\linewidth][c]{\includegraphics[width=\linewidth]{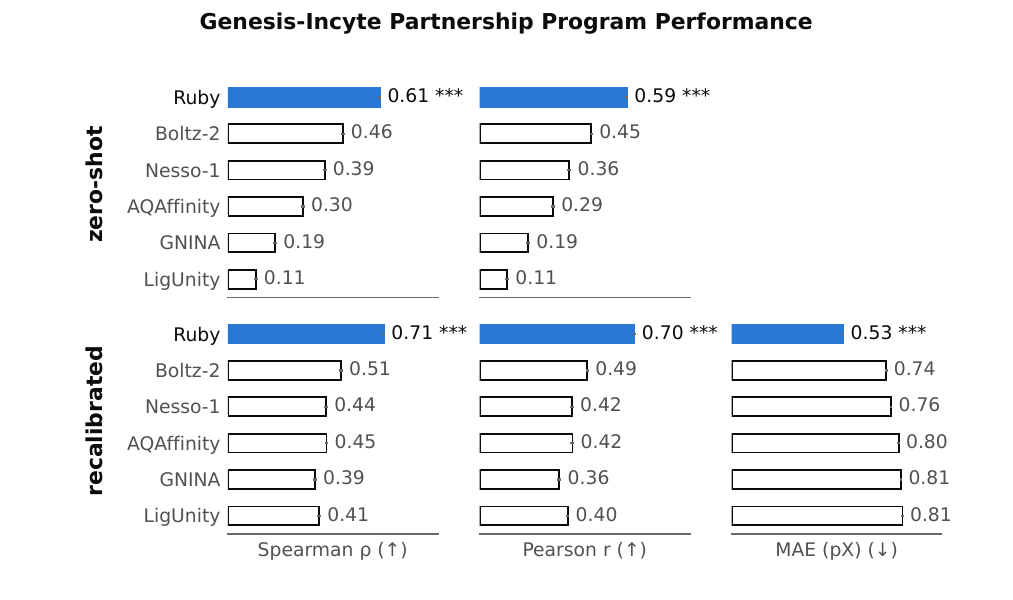}}
  \caption{Performance across eight \partnercohort assays on five targets.
  The top row shows zero-shot Spearman $\rho$ and Pearson $r$ across these assays.
  The bottom row shows correlations and MAE of rolling-temporal recalibrated predictions.
  Whiskers show $\pm$ one bootstrap standard error from compound resampling within the eight fixed assays.
  Three stars on each \ruby bar indicate a one-sided paired bootstrap $p<0.001$ against the best plotted comparator for that metric (Boltz-2, Appendix Table~\ref{tab:partner-paired}).
  \statustag{programs}}
  \label{fig:programs}
\end{figure}

\subsection{Lead optimization on \partnercohort programs}
\label{sec:results:programs}

Figure~\ref{fig:programs} tests \ruby on experimental measurements from ongoing \partnercohort drug-discovery programs, extending the evaluation to proprietary chemical series and program-specific biochemical and cellular assays.
Across these settings, \ruby achieves the highest aggregate zero-shot and recalibrated correlations among the evaluated methods.

The top row of Figure~\ref{fig:programs} shows zero-shot potency ranking across eight assays on five \partnercohort targets, with each assay weighted equally.
\ruby reaches mean Spearman $\rho=0.61$ and Pearson $r=0.59$, compared with $0.46$ and $0.45$ for \boltz{2}, the strongest aggregate comparator for both metrics.
Both \boltz{2} and \nesso achieve lower Pearson correlations here than on each of the three public benchmarks evaluated with that metric, emphasizing the importance of testing performance directly in active programs.

The bottom row of Figure~\ref{fig:programs} shows performance over rolling-temporal recalibration within the assay.
\ruby's mean MAE is 0.53 pX, compared with 0.74 pX for \boltz{2}, the lowest-MAE plotted comparator.
On the shared recalibrated cohort, \ruby reaches mean $\rho=0.71$ and $r=0.70$, compared with $0.51$ and $0.49$ for \boltz{2}.
Appendix Table~\ref{tab:incyte-detail} reports results for each assay.

\begin{figure}[t!]
  \centering
  \includegraphics[width=\linewidth]{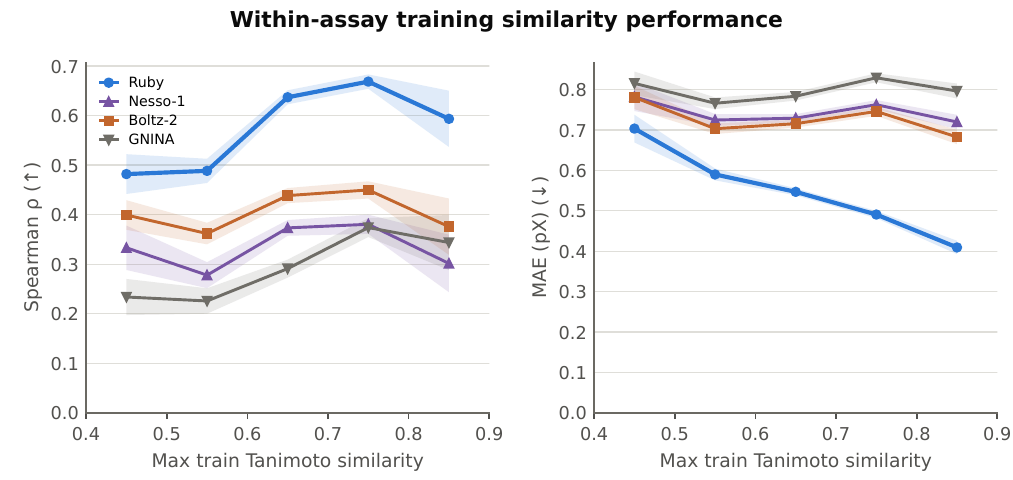}
  \caption{Spearman $\rho$ (left) and MAE (right) in 0.1-wide bins of maximum Tanimoto similarity to earlier exact same-assay measurements, across eight \partnercohort assays on five targets.
  Both metrics use rolling-temporal recalibrated predictions for every method.
  Shading shows assay-level bootstrap standard errors.
  Within-bin rank values omit between-bin comparisons and are not directly comparable with the whole-cohort values in Figure~\ref{fig:programs}.}
  \label{fig:partner-similarity}
\end{figure}

Figure~\ref{fig:partner-similarity} shows recalibrated ranking and prediction error as a function of maximum Tanimoto similarity to earlier compounds measured in the same assay.
Across bins represented in all eight assays, \ruby's MAE decreases more dramatically as similarity to prior training data increases.
Appendix~\ref{app:partner-similarity} gives the binning and rank-score details.

\internalnote{}

\section{Discussion}
\label{sec:discussion}

The discovery regimes introduced above place different demands on a potency predictor: early enrichment over large candidate sets, discrimination among related compounds, and adaptation to the assays that guide an ongoing program.
The \genesispotency training set addresses the experimental supervision on which such predictors depend. \ruby uses learned structural representations to make predictions across these settings.
Across these evaluations, \ruby's strong performance supports the use of a common pretrained predictor for HitID, H2L, and LeadOp, with computational effort and assay-specific recalibration adjusted to the available resources and experimental history.
Together, these results suggest that learned structural representations can support compound prioritization both before substantial on-target measurements are available and as those measurements accumulate.

Beyond supporting \ruby, \genesispotency provides a large, curated training resource that is publicly available to the community.
By combining public-source measurements with standardized annotations and source-level auditing, the release makes this experimental supervision more accessible for model development.

\paragraph{Screening at scale.}
For HitID, prediction quality and computational cost jointly determine how broadly a method can be applied.
Under a fixed computational budget, lower inference cost allows more candidates to be scored, while stronger early enrichment helps concentrate active compounds within the subset selected for experimental testing.
\ruby's effort sweep exposes this trade-off: lower-effort settings retain useful enrichment at substantially lower inference cost, whereas higher-effort settings provide stronger enrichment when additional computation is available.
At the recorded inference rates, the GPU time required to score 10{,}000 compounds with the \boltz{2} reference setting would support scoring approximately 12.5 million compounds with \ruby Low.
This throughput advantage could substantially expand the chemical space explored within a fixed inference budget.
Nvidia kernel optimization provides a complementary improvement, increasing warm inference throughput approximately threefold at fixed model effort on the example workload.
These retrospective results motivate prospective evaluation of how \ruby's combination of speed and enrichment translates into hit recovery under fixed computational and experimental budgets.

\paragraph{Ranking within chemical series.}
The public potency benchmarks in Figure~\ref{fig:series} test discrimination among compounds within a target, including the congeneric series relevant to H2L.
In this setting, a predictor must resolve activity differences among related molecules to help prioritize which analogs to synthesize and test.
On \openfe, \ruby achieves an aggregate Pearson correlation of $0.74$, compared with $0.72$ for FEP+, demonstrating strong ranking even before any assay-specific recalibration.
This result suggests that learned structural representations can capture information useful for compound prioritization alongside established free-energy methods.
Learning from the physical inductive biases of free-energy methods could further improve AI-based potency prediction while retaining the breadth and speed of learned models.

\paragraph{Prediction and adaptation in active programs.}
The \partnercohort results demonstrate why evaluation on active pharmaceutical programs matters alongside performance on established public benchmarks.
Both \boltz{2} and \nesso show lower aggregate Pearson correlations on these programs than on the public potency benchmarks, indicating that their public benchmark performance does not fully transfer to these program settings.
\ruby's stronger ranking on the same measurements supports its relevance to the compound-prioritization decisions faced in ongoing discovery.
With mean zero-shot Spearman $\rho=0.61$ and Pearson $r=0.59$, \ruby provides a stronger starting point for ranking compounds before assay-specific recalibration.

As a program advances into LeadOp, its experimental history becomes increasingly valuable for compound prioritization because earlier measurements describe local structure--activity relationships under the assay conditions used to guide the program.
This context matters because biochemical and cellular potency reflect different experimental quantities, and even assays against the same target can yield different measurements~\cite{kalliokoski2013mixedic50,postnikova2018cellassays}.
Learning from these observations is well established in QSAR~\cite{heid2024chemprop} and is also pursued by pretrained models such as ActFound~\cite{feng2024actfound} and recent work on project-specific fine-tuning of Boltz-2~\cite{amini2026boltzfinetuning}.
Beyond zero-shot accuracy, usefulness in LeadOp depends on combining knowledge learned across targets with earlier assay measurements to predict subsequent compounds.

Figure~\ref{fig:programs} evaluates this setting across eight biochemical and cellular assays on five \partnercohort targets.
At successive temporal cutoffs, recalibration uses earlier measurements from the same assay to predict compounds measured later, testing performance at multiple stages of each program.
\ruby's higher aggregate correlations and lower prediction error support its usefulness for both relative prioritization and estimation of assay-specific potency.
In ongoing programs, these capabilities can help prioritize compounds within a series and assess whether their predicted potency, alongside other properties, warrants further testing.

Figure~\ref{fig:partner-similarity} examines how recalibrated performance varies with chemical similarity to earlier compounds measured in the same assay.
As expected, \ruby's prediction error is lower in bins with greater similarity to those earlier measurements, and its error declines more across the reported bins than that of the comparison methods.
This pattern is consistent with the practical value of local experimental history during LeadOp, where successive compounds often refine an existing series.
Evaluating performance within familiar chemistry is therefore important alongside testing generalization to new scaffolds, and rolling temporal evaluation retains these relationships while restricting recalibration to earlier measurements.

\paragraph{Learning from structural representations.}
\ruby's strong performance across these regimes supports the use of intermediate cofolding representations for potency prediction.
By using information learned through structural training without generating an explicit pose for every compound, \ruby avoids the major portion of inference cost, helping make the approach practical for larger candidate sets.
The potency benchmarks demonstrate that useful information for ranking compounds is retained in this setting.
Explicit structural models remain valuable for interpreting interactions and developing hypotheses about how chemical modifications affect binding.
We therefore expect models like \pearl to increasingly complement potency predictors like \ruby that share intermediate representations, optionally providing structural hypotheses for interpreting structure--activity relationships only for the most interesting compounds.

\paragraph{Limitations.}
These evaluations are retrospective and measure predictive performance on compounds selected through existing experimental workflows.
The HitID benchmarks re-rank experimentally tested selections, so they do not directly establish enrichment over the very large libraries from which those compounds originated.
Similarly, rolling temporal evaluation preserves the ordering of assay measurements but does not establish how \ruby{}-guided choices would change the compounds synthesized or the subsequent course of a program.
The training-data audit was not able to access sources for every assay and so has partial source coverage.
It samples compounds within documents, so retained records could still contain some errors that were not detected.

\clearpage
\section*{Contributions}
\noindent\textbf{Writing:} Drausin Wulsin, Frank C. Pickard IV, Dominique Beaini, Ken Leidal, Steven Kearnes.

\noindent\textbf{Experiments:} Michael Atkin, Drausin Wulsin, Jules Berman, Dominique Beaini, Nina Jovic, Gabriel Monteiro da Silva, Allison Keys.

\noindent\textbf{\ruby model:} Han Wool Yoon, Jules Berman, Rebecca Boiarsky, Drausin Wulsin, Michael Atkin, Wesley Wei Qian, Will Harvey, William Wang, Ansh Sharma, Michael V. Levine, Aleksandra Faust.

\noindent\textbf{\pearl model:} Nate Gruver, Jules Berman, Richard Strong Bowen, Alejandro Dobles, Pranav Murugan, Gianluca Scarpellini, David Charles Williams.

\noindent\textbf{\genesispotency curation:} Han Wool Yoon, Rebecca Boiarsky, Wesley Wei Qian, Adi Nagle, Deidre Johns, Gerald Lelais, Indrawan McAlpine.

\noindent\textbf{Compute infrastructure:} Igor Morozov, Han Wang, Cyr Vetcher, Rohit Kundu, Jamin Chen, Wojtek Swiderski.

\noindent\textbf{Leadership:} Sergey Edunov, Carl Tilbury, Laura Tapias, Evan N. Feinberg.

Correspondence may be sent to \texttt{papers@genesis.ml}.

\section*{Acknowledgements}
We are grateful to Nvidia, including Roy Tal Dew and Kushal S. Shah, for their support in optimizing \pearl training and inference, since \ruby benefits from \pearl's structural foundation.
Nvidia's kernel optimizations and libraries such as cuEquivariance have also helped us improve \ruby's performance on B200 GPUs, allowing larger screens within a given computational budget.

We are also grateful to Incyte for our ongoing AI-focused research collaboration and for generating and providing the experimental data associated with the results in Section~\ref{sec:results:programs}.

We thank Pat Walters for providing feedback on an early draft of this report.

\clearpage
\printbibliography

@article{baumann2026openfe,
  title={Industry-scale benchmarking of relative binding free energy calculations},
  author={Baumann, Hannah M. and others},
  journal={J. Chem. Inf. Model.},
  volume={66},
  pages={6429--6452},
  year={2026},
  doi={10.1021/acs.jcim.6c00089}
}

@article{nelen2026openbind,
  title={The first {OpenBind} release: An open experimental structure--affinity dataset and benchmark for structure-based {AI}},
  author={Nelen, Jochem and Khan, Omeir and Adams, Etowah and others},
  journal={bioRxiv},
  year={2026},
  doi={10.64898/2026.08.27.747600}
}

@article{lyu2019ultralarge,
  title={Ultra-large library docking for discovering new chemotypes},
  author={Lyu, Jiankun and others},
  journal={Nature},
  volume={566},
  pages={224--229},
  year={2019},
  doi={10.1038/s41586-019-0917-9}
}

@article{alon2021sigma2,
  title={Structures of the $\sigma_2$ receptor enable docking for bioactive ligand discovery},
  author={Alon, Assaf and others},
  journal={Nature},
  volume={600},
  pages={759--764},
  year={2021},
  doi={10.1038/s41586-021-04175-x}
}

@article{liu2025scale,
  title={The impact of library size and scale of testing on virtual screening},
  author={Liu, Fangyu and others},
  journal={Nature Chemical Biology},
  volume={21},
  pages={1039--1045},
  year={2025},
  doi={10.1038/s41589-024-01797-w}
}

@article{cache2022roadmap,
  title={{CACHE} (Critical Assessment of Computational Hit-finding Experiments): a public--private partnership benchmarking initiative to enable the development of computational methods for hit-finding},
  author={Ackloo, Suzanne and others},
  journal={Nature Reviews Chemistry},
  volume={6},
  pages={287--295},
  year={2022},
  doi={10.1038/s41570-022-00363-z}
}

@article{herasymenko2026cache3,
  title={{CACHE} Challenge \#3: Targeting the {Nsp3} Macrodomain of {SARS-CoV-2}},
  author={Herasymenko, Oleksandra and Silva, Madhushika and Correy, Galen J. and others},
  journal={Journal of Chemical Information and Modeling},
  volume={66},
  pages={1566--1581},
  year={2026},
  doi={10.1021/acs.jcim.5c02441}
}

@misc{cachechallenge2026,
  title={{CACHE} challenges and results},
  author={{CACHE Challenge}},
  url={https://cache-challenge.org/node/},
  urldate={2026-11-03}
}

@online{enamine2026realspace,
  title={{Enamine REAL Space}},
  author={{Enamine}},
  url={https://enamine.net/compound-collections/real-compounds/real-space-navigator},
  urldate={2026-11-09}
}

@article{mysinger2012dude,
  title={Directory of Useful Decoys, Enhanced ({DUD-E}): better ligands and decoys for better benchmarking},
  author={Mysinger, Michael M. and Carchia, Michael and Irwin, John J. and Shoichet, Brian K.},
  journal={Journal of Medicinal Chemistry},
  volume={55},
  pages={6582--6594},
  year={2012},
  doi={10.1021/jm300687e}
}

@article{chen2019dudebias,
  title={Hidden bias in the {DUD-E} dataset leads to misleading performance of deep learning in structure-based virtual screening},
  author={Chen, Lieyang and Cruz, Anthony and Ramsey, Steven and Dickson, Callum J. and Duca, Jose S. and Hornak, Viktor and Koes, David R. and Kurtzman, Tom},
  journal={PLOS ONE},
  volume={14},
  number={8},
  pages={e0220113},
  year={2019},
  doi={10.1371/journal.pone.0220113}
}

@article{buterez2023mfpcba,
  title={{MF-PCBA}: multifidelity high-throughput screening benchmarks for drug discovery and machine learning},
  author={Buterez, David and others},
  journal={Journal of Chemical Information and Modeling},
  year={2023},
  doi={10.1021/acs.jcim.2c01569}
}

@article{passaro2025boltz2,
  title={Boltz-2: Towards Accurate and Efficient Binding Affinity Prediction},
  author={Passaro, Saro and Corso, Gabriele and Wohlwend, Jeremy and Reveiz, Mateo and Thaler, Stephan and Ram Somnath, Vignesh and Getz, Noah and Portnoi, Tally and Roy, Julien and Stark, Hannes and others},
  journal={bioRxiv},
  pages={2025.06.14.659707},
  year={2025},
  publisher={Cold Spring Harbor Laboratory}
}

@article{allenspach2024multitask,
  title={Neural multi-task learning in drug design},
  author={Allenspach, Stephan and Hiss, Jan A. and Schneider, Gisbert},
  journal={Nat. Mach. Intell.},
  volume={6},
  number={2},
  pages={124--137},
  year={2024},
  doi={10.1038/s42256-023-00785-4}
}

@article{bender2021docking,
  title={A practical guide to large-scale docking},
  author={Bender, Brian J. and Gahbauer, Stefan and Luttens, Andreas and Lyu, Jiankun and Webb, Chase M. and Stein, Reed M. and Fink, Elissa A. and Balius, Trent E. and Carlsson, Jens and Irwin, John J. and Shoichet, Brian K.},
  journal={Nat. Protoc.},
  volume={16},
  number={10},
  pages={4799--4832},
  year={2021},
  doi={10.1038/s41596-021-00597-z}
}

@article{heid2024chemprop,
  title={{Chemprop}: A Machine Learning Package for Chemical Property Prediction},
  author={Heid, Esther and Greenman, Kevin P. and Chung, Yunsie and Li, Shih-Cheng and Graff, David E. and Vermeire, Florence H. and Wu, Haoyang and Green, William H. and McGill, Charles J.},
  journal={J. Chem. Inf. Model.},
  volume={64},
  number={1},
  pages={9--17},
  year={2024},
  doi={10.1021/acs.jcim.3c01250}
}

@article{lam2025structure,
  title={Navigating structure-based drug discovery with emerging innovations in physics- and knowledge-based approaches},
  author={Lam, Jordy Homing and Katritch, Vsevolod},
  journal={npj Drug Discovery},
  volume={2},
  number={1},
  pages={29},
  year={2025},
  doi={10.1038/s44386-025-00031-4}
}

@article{moesgaard2026sensitivity,
  title={Modeling the Sensitivity of Large-Scale Virtual Screening to Scoring Function Accuracy, Artifacts, and Library Composition},
  author={Moesgaard, Laust and Shoichet, Brian K. and Mailhot, Olivier},
  journal={J. Chem. Inf. Model.},
  volume={66},
  number={14},
  pages={8325--8335},
  year={2026},
  doi={10.1021/acs.jcim.6c01066}
}

@article{muegge2023alchemical,
  title={Recent Advances in Alchemical Binding Free Energy Calculations for Drug Discovery},
  author={Muegge, Ingo and Hu, Yuan},
  journal={ACS Med. Chem. Lett.},
  volume={14},
  number={3},
  pages={244--250},
  year={2023},
  doi={10.1021/acsmedchemlett.2c00541}
}

@article{wang2025interactions,
  title={Modeling protein--ligand interactions for drug discovery in the era of deep learning},
  author={Wang, Yuzhe and Li, Yibo and Chen, Jiaxiao and Lai, Luhua},
  journal={Chem. Soc. Rev.},
  volume={54},
  number={23},
  pages={11141--11183},
  year={2025},
  doi={10.1039/d5cs00415b}
}

@article{sadybekov2023computational,
  title={Computational approaches streamlining drug discovery},
  author={Sadybekov, Anastasiia V and Katritch, Vsevolod},
  journal={Nature},
  volume={616},
  number={7958},
  pages={673--685},
  year={2023},
  publisher={Nature Publishing Group UK London}
}

@article{mcnutt2025gnina13,
  title={{GNINA} 1.3: the next increment in molecular docking with deep learning},
  author={McNutt, Andrew T. and Li, Yanjing and Meli, Rocco and Aggarwal, Rishal and Koes, David Ryan},
  journal={Journal of Cheminformatics},
  volume={17},
  pages={28},
  year={2025},
  doi={10.1186/s13321-025-00973-x}
}

@article{genesis2025pearl,
  title   = {Pearl: A Foundation Model for Placing Every Atom in the Right Location},
  author  = {{Genesis Research Team}},
  journal = {arXiv preprint arXiv:2510.24670},
  year    = {2025},
}

@techreport{isomorphic2026isodde,
  title       = {Accurate Predictions of Novel Biomolecular Interactions with {IsoDDE}},
  author      = {{Isomorphic Labs Team}},
  institution = {Isomorphic Labs},
  year        = {2026},
  month       = feb,
  url         = {https://storage.googleapis.com/isomorphiclabs-website-public-artifacts/isodde_technical_report.pdf},
}

@article{truchon2007bedroc,
  title   = {Evaluating Virtual Screening Methods: Good and Bad Metrics for the ``Early Recognition'' Problem},
  author  = {Truchon, Jean-Fran{\c{c}}ois and Bayly, Christopher I.},
  journal = {Journal of Chemical Information and Modeling},
  volume  = {47},
  number  = {2},
  pages   = {488--508},
  year    = {2007},
}

@article{wang2015fep,
  title={Accurate and Reliable Prediction of Relative Ligand Binding Potency in Prospective Drug Discovery by Way of a Modern Free-Energy Calculation Protocol and Force Field},
  author={Wang, Lingle and Wu, Yujie and Deng, Yuqing and Kim, Byungchan and Pierce, Levi and Krilov, Goran and Lupyan, Dmitry and Robinson, Shaughnessy and Dahlgren, Markus K. and Greenwood, Jeremy and others},
  journal={Journal of the American Chemical Society},
  volume={137},
  number={7},
  pages={2695--2703},
  year={2015},
  doi={10.1021/ja512751q}
}

@article{chen2023abfe,
  title={Enhancing Hit Discovery in Virtual Screening through Absolute Protein--Ligand Binding Free-Energy Calculations},
  author={Chen, Wei and Cui, Di and Jerome, Steven V. and Michino, Mayako and Lenselink, Eelke B. and Huggins, David J. and Beautrait, Alexandre and Vendome, Jeremie and Abel, Robert and Friesner, Richard A. and others},
  journal={Journal of Chemical Information and Modeling},
  volume={63},
  number={10},
  pages={3171--3185},
  year={2023},
  doi={10.1021/acs.jcim.3c00013}
}

@article{ross2023maximal,
  title={The maximal and current accuracy of rigorous protein-ligand binding free energy calculations},
  author={Ross, Gregory A. and Lu, Chao and Scarabelli, Guido and Albanese, Steven K. and Houang, Evelyne and Abel, Robert and Harder, Edward D. and Wang, Lingle},
  journal={Communications Chemistry},
  volume={6},
  pages={222},
  year={2023},
  doi={10.1038/s42004-023-01019-9}
}

@article{anew2026anewdde,
  title={{AnewDDE}: An Agentic Drug Discovery Engine for Biomolecular Interaction Modelling and Closed-Loop Design},
  author={{Anew Labs Team}},
  journal={bioRxiv},
  year={2026},
  doi={10.64898/2026.09.20.752967}
}

@article{nesso2026,
  title={{Nesso-1}: Accelerating Open-Source Binding Affinity Predictions},
  author={Shenoy, Nikhil and Errington, David and Bengio, Emmanuel and
          Kapu{\'s}niak, Kacper and Klaeser, Kerstin and Pang, Yui Tik and
          Radenkovic, Vladimir and Tossou, Prudencio and Bois, Therence and
          Wedlake, Andrew and {Di Giovanni}, Francesco},
  journal={bioRxiv},
  year={2026},
  doi={10.64898/2026.08.01.742196},
  url={https://www.biorxiv.org/content/10.64898/2026.08.01.742196v1},
  note={Preprint, posted 2026-08-03}
}

@online{bucher2026apherisaffinity,
  title={Grounding {ApherisAffinity} predictions in binding structure},
  author={Bucher, Alwin Otto and Javer, Avelino and Triendl, Hagen and Hales, Ian},
  year={2026},
  month={10},
  day={29},
  url={https://www.apheris.com/resources/grounding-apherisaffinity-predictions-in-binding-structure},
  note={Updated 29 October 2026}
}

@article{ligunity2025,
  title={Hierarchical affinity landscape navigation through learning a shared pocket-ligand space},
  author={Feng, Bin and Liu, Zijing and Li, Hao and Yang, Mingjun and Zou, Junjie and
          Cao, He and Li, Yu and Zhang, Lei and Wang, Sheng},
  journal={Patterns},
  volume={6},
  number={10},
  pages={101371},
  year={2025},
  doi={10.1016/j.patter.2025.101371},
  note={The model is named {LigUnity}. Preprint bioRxiv doi:10.1101/2025.02.17.638554}
}

@misc{sandboxaq2026aqaffinity,
  title        = {{AQAffinity}: Open-Source Structure-to-Affinity Built on {OpenFold3} by {SandboxAQ}},
  author       = {{SandboxAQ}},
  year         = {2026},
  month        = jan,
  howpublished = {SandboxAQ Blog},
  url          = {https://www.sandboxaq.com/post/aqaffinity-open-source-structure-to-affinity-built-on-openfold3-by-sandboxaq},
  note         = {Model and code: \url{https://huggingface.co/SandboxAQ/AQAffinity}}
}

@article{zdrazil2024chembl,
  title={The {ChEMBL} Database in 2023: a drug discovery platform spanning multiple bioactivity data types and time periods},
  author={Zdrazil, Barbara and Felix, Eloy and Hunter, Fiona and Manners, Emma J. and Blackshaw, James and Corbett, Sybilla and de Veij, Marleen and Ioannidis, Harris and Mendez Lopez, David and Mosquera, Juan F. and Magari{\~n}os, Maria P. and Bosc, Nicolas and Arcila, Ricardo and K{\i}z{\i}l{\"o}ren, Tevfik and Gaulton, Anna and Bento, A. Patr{\'i}cia and Adasme, Melissa F. and Monecke, Peter and Landrum, Gregory A. and Leach, Andrew R.},
  journal={Nucleic Acids Research},
  volume={52},
  number={D1},
  pages={D1180--D1192},
  year={2024},
  doi={10.1093/nar/gkad1004}
}

@article{gilson2016bindingdb,
  title={{BindingDB} in 2015: a public database for medicinal chemistry, computational chemistry and systems pharmacology},
  author={Gilson, Michael K. and Liu, Tiqing and Baitaluk, Michael and Nicola, George and Hwang, Linda and Chong, Jenny},
  journal={Nucleic Acids Research},
  volume={44},
  number={D1},
  pages={D1045--D1053},
  year={2016},
  doi={10.1093/nar/gkv1072}
}

@article{kim2025pubchem,
  title={{PubChem} 2025 update},
  author={Kim, Sunghwan and Chen, Jie and Cheng, Tiejun and Gindulyte, Asta and He, Jia and He, Siqian and Li, Qingliang and Shoemaker, Benjamin A. and Thiessen, Paul A. and Yu, Bo and Zaslavsky, Leonid and Zhang, Jian and Bolton, Evan E.},
  journal={Nucleic Acids Research},
  volume={53},
  number={D1},
  pages={D1516--D1525},
  year={2025},
  doi={10.1093/nar/gkae1059}
}

@article{rossi2026terrabind,
  title={{TerraBind}: fast and accurate binding affinity prediction through coarse structural representations},
  author={Rossi, Matteo and Pederson, Ryan and others},
  journal={arXiv},
  pages={2602.07735},
  year={2026},
  doi={10.48550/arXiv.2602.07735},
}

@article{gilson2026casp16,
  title={Assessment of Pharmaceutical Protein--Ligand Pose and Affinity Predictions in {CASP16}},
  author={Gilson, Michael K. and Eberhardt, Jerome and {\v{S}}krinjar, Petar and Durairaj, Janani and Robin, Xavier and Kryshtafovych, Andriy},
  journal={Proteins: Structure, Function, and Bioinformatics},
  volume={94},
  number={1},
  pages={249--266},
  year={2026},
  doi={10.1002/prot.70061},
}

@online{casp16pharmaarchive,
  title={CASP16 Pharmaceutical Ligand Data Archive},
  author={{CASP Prediction Center}},
  year={2024},
  url={https://predictioncenter.org/download_area/CASP16/extra_experiments/pharma_ligands/},
  urldate={2026-11-03},
}

@article{kalliokoski2013mixedic50,
  title={Comparability of Mixed {IC50} Data: A Statistical Analysis},
  author={Kalliokoski, Tuomo and Kramer, Christian and Vulpetti, Anna and Gedeck, Peter},
  journal={PLOS ONE},
  volume={8},
  number={4},
  pages={e61007},
  year={2013},
  doi={10.1371/journal.pone.0061007},
}

@article{liu2019nef,
  title={Practical Model Selection for Prospective Virtual Screening},
  author={Liu, Shengchao and Alnammi, Moayad and Ericksen, Spencer S. and Voter, Andrew F. and Ananiev, Gene E. and Keck, James L. and Hoffmann, F. Michael and Wildman, Scott A. and Gitter, Anthony},
  journal={Journal of Chemical Information and Modeling},
  volume={59},
  number={1},
  pages={282--293},
  year={2019},
  doi={10.1021/acs.jcim.8b00363},
}

@article{feng2024actfound,
  title={A bioactivity foundation model using pairwise meta-learning},
  author={Feng, Bin and Liu, Zequn and Huang, Nanlan and Xiao, Zhiping and Zhang, Haomiao and Mirzoyan, Srbuhi and Xu, Hanwen and Hao, Jiaran and Xu, Yinghui and Zhang, Ming and Wang, Sheng},
  journal={Nature Machine Intelligence},
  volume={6},
  number={8},
  pages={962--974},
  year={2024},
  doi={10.1038/s42256-024-00876-w},
}

@article{amini2026boltzfinetuning,
  title={Affinity Fine-Tuning of {Boltz-2}: An Open Framework for Protein-Ligand Potency Prediction in Drug Discovery},
  author={Amini, Samad and Sciabola, Simone and Wang, Ye},
  journal={bioRxiv},
  year={2026},
  doi={10.64898/2026.05.26.727958},
  note={Preprint},
}

@article{postnikova2018cellassays,
  title={Testing therapeutics in cell-based assays: Factors that influence the apparent potency of drugs},
  author={Postnikova, Elena and Cong, Yu and DeWald, Lisa Evans and others},
  journal={PLOS ONE},
  volume={13},
  number={3},
  pages={e0194880},
  year={2018},
  doi={10.1371/journal.pone.0194880},
}

@article{bento2020chembl,
  title={An open source chemical structure curation pipeline using {RDKit}},
  author={Bento, A. Patr{\'i}cia and Hersey, Anne and F{\'e}lix, Eloy and Landrum, Greg and Gaulton, Anna and Atkinson, Francis and Bellis, Louisa J. and De Veij, Marleen and Leach, Andrew R.},
  journal={Journal of Cheminformatics},
  volume={12},
  pages={51},
  year={2020},
  doi={10.1186/s13321-020-00456-1}
}

@article{kearnes2024ordinal,
  title={Ordinal Confidence Level Assignments for Regression Model Predictions},
  author={Kearnes, Steven and Riley, Patrick},
  journal={Journal of Chemical Information and Modeling},
  volume={64},
  number={24},
  pages={9299--9305},
  year={2024},
  doi={10.1021/acs.jcim.4c01755},
}

\appendix

\clearpage
\section*{Appendix contents}
\pdfbookmark[0]{Appendix contents}{appendix-contents}
\startcontents[appendices]
\begingroup
\hypersetup{linktoc=all}
\printcontents[appendices]{}{1}{\setcounter{tocdepth}{2}}
\endgroup

\clearpage
\section{\genesispotency: source curation and audit}
\label{app:corpus}

\subsection{Sources, eligibility, and label preparation}
\genesispotency covers the public-source portion of the potency-data curation, drawing on ChEMBL 37, BindingDB release 202607, and PubChem HTS (2025)~\cite{zdrazil2024chembl,gilson2016bindingdb,kim2025pubchem}.
The full \ruby training index additionally includes proprietary measurements outside this public-source corpus.
For the training build used here, we retain measurements dated before 2023-01-01 and group them by document, target, and readout.
\ruby and IsoDDE both restrict supervised affinity training to measurements dated before 2023, imposing an explicit temporal restriction absent from the reported training criteria of the other comparison methods in Table~\ref{tab:deleak}~\cite{isomorphic2026isodde}.
Those methods do not report an affinity-data date cutoff, so their temporal training restrictions cannot be established from the available descriptions.
We convert IC$_{50}$, $K_i$, and $K_d$ to \pX, while EC$_{50}$ contributes binary activity labels only.
For numerical potency measurements used for binary supervision, we assign activity at $\pX \geq 6.70$.
Eligibility filters exclude ligands with more than 50 heavy atoms, PAINS-flagged compounds, covalent assays, and assays with fewer than 10 measurements.
Quality control and sampling reduce the 157 million input rows (Table~\ref{tab:training-audit}) to 5.9 million potency data points.
Most of the reduction is in PubChem HTS, where we keep every active and nine inactives per active, following \boltz{2}~\cite{passaro2025boltz2}.
Source records retain their reported relation qualifiers, and censored regression labels are used at their reported bounds in model training.
Benchmark-specific target exclusions are described in Appendix~\ref{app:deleak}.
Measurements may also supply more than one task, so row counts do not represent independent experiments.

\subsection{Source-document audit protocol}
We reviewed source documents in two passes.
The document-level pass checked each assay group's target, species, construct, origin of reported values, endpoint, and units.
The compound-level pass examined up to 15 compounds per document, comparing structures with drawings, systematic names, and substituent tables, and comparing values and qualifiers with the reported measurements.
In total, this pass examined 115{,}235 sampled compounds from 8{,}995 documents.
The two passes took approximately 160 and 700 agent-hours, respectively.

Full text was available for 5{,}329 patents and 3{,}666 articles.
For 19{,}746 paywalled articles, only abstracts were available.
Rejection decisions required a contradictory source passage.
Code checked the quotation against the document, and an independent reviewer assessed the decision without seeing the first review.
Structured reports recorded the fields checked and the proposed actions.
We removed the affected assay group if we found a document-level contradiction or errors in at least 20\% of sampled compounds.
Records were marked for correction when the source supported a specific replacement and for removal when it did not.
The audit does not cover PubChem HTS records, and it does not imply that every compound within an audited document was inspected.

\subsection{Audit outcomes and remaining work}
\label{app:audit-outcomes}
\begin{figure}[t!]
  \centering
  \includegraphics[width=\linewidth]{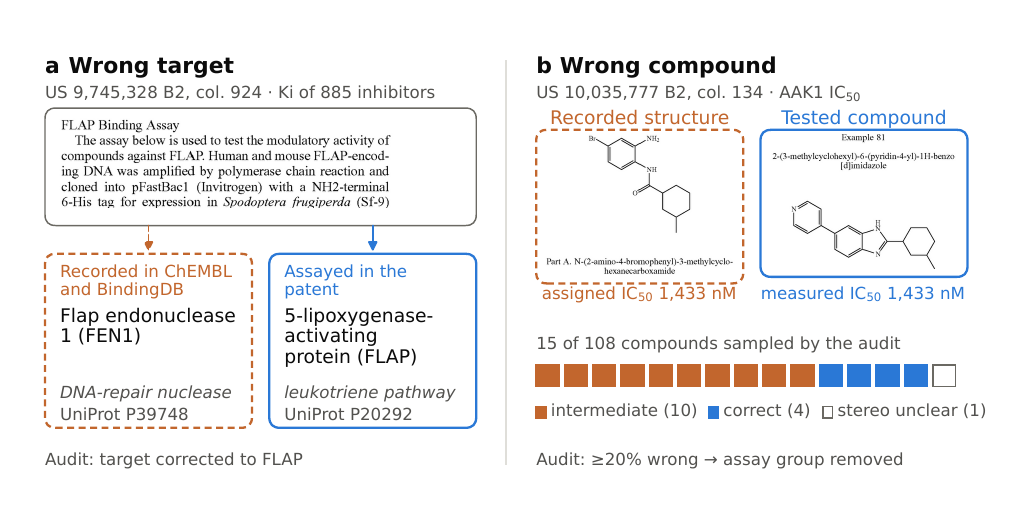}
  \caption{Source-document evidence for the two audit examples, cropped from the issued patents.
  In each panel, the database record (orange) is shown left of the source (blue).
  (a) Patent US9745328 assays FLAP, but ChEMBL and BindingDB assigned the $K_i$ values of its 885 inhibitors to flap endonuclease 1 (FEN1).
  (b) For Example~81 of patent US10035777, both databases record the synthetic intermediate drawn below the tested compound.
  Ten of the 15 compounds sampled from this patent had the same error.}
  \label{fig:audit-examples}
\end{figure}

Of 2.51 million ChEMBL and BindingDB source records in scope, 1.24 million belonged to auditable documents.
The audit retained 699{,}742 unchanged, marked 52{,}575 for correction, and removed 486{,}600 (Table~\ref{tab:training-audit}).
Of the 5.90 million pre-audit potency data points (Figure~\ref{fig:genesispotency}(b)), 8.4\% were removed or withheld, 0.02\% were corrected, and 41.9\% passed source-document review, including partial passes: 13.6\% against the full text and 28.3\% against the abstract or supplement alone.
Another 2.6\% are ChEMBL and BindingDB data points without an accessible source or with an unresolved review.
PubChem HTS screening calls (43.7\%), which screening centers deposit directly, and synthetic decoys (3.4\%) have no source document and pass without review.
Of 486{,}600 removed records, 338{,}718 belonged to assay groups excluded because at least 20\% of sampled compounds had errors.
Other reasons included values not measured in the cited document, activity classes or thresholds treated as measured values, and mismatched targets, constructs, endpoints, or units.
Figure~\ref{fig:audit-examples} shows the two examples from the main text in the issued patents.
In patent \texttt{US9745328}, the binding assay measures 5-lipoxygenase-activating protein (FLAP), but both databases assigned the $K_i$ values of its 885 inhibitors to flap endonuclease 1.
The audit remapped these records to FLAP.
These records are withheld from the frozen training build because the correction requires a new protein input.
In patent \texttt{US10035777}, the record for Example~81 depicts the synthetic intermediate drawn below the tested title compound and carries the title compound's AAK1 IC$_{50}$.
Ten of the 15 sampled records had the same error, so the audit removed the assay group.
Both errors originated in BindingDB and reached ChEMBL through its import of BindingDB patent data.

The audit covered the ChEMBL and BindingDB records of a broader curation build without the 2023 date cutoff, so its scope includes records outside the training eligibility filters.

Of 52{,}575 source records marked for correction, 49{,}575 require new protein or ligand inputs: 31{,}585 target remappings, 17{,}384 sequence changes for stated point mutations, and 606 ligand-structure corrections.
The frozen training build applies supported corrections where the required inputs are available and withholds records requiring new inputs pending their preparation.
Records from inaccessible documents, partial reviews, and documents that were ambiguous on a deciding point were retained, so the training corpus contains both source-checked and unverified records.
The public-source corpus is distinct from proprietary training records and from the papers and patents consulted during the audit.

\begin{table}[!t]
\centering\small
\caption{Public-source inputs and source-document audit outcomes for \genesispotency.
The middle block counts the potency data shown in Figure~\ref{fig:genesispotency}(c).
Synthetic decoys count toward the source of their assay.
The lower block counts ChEMBL and BindingDB source-record audit decisions, a different population and unit from the input rows and potency data.
The audit covered the ChEMBL and BindingDB records of a broader curation build without the 2023 date cutoff, so its scope includes records outside the training eligibility filters.
Partial or unavailable review includes records for which only abstract-level checks were possible.
Retained unchanged includes 21{,}844 records whose source document was read but was ambiguous on a deciding point.
PubChem HTS was outside the source-document audit (---), so audit totals cover ChEMBL and BindingDB only.}
\label{tab:training-audit}
\begin{tabular}{@{}lrrrr@{}}
\toprule
Stage & ChEMBL & BindingDB & PubChem HTS & Total \\
\midrule
Available input rows & 24{,}527{,}044 & 3{,}228{,}554 & 129{,}334{,}526 & 157{,}090{,}124 \\
\midrule
Potency data before audit & 1{,}356{,}596 & 1{,}965{,}221 & 2{,}583{,}180 & 5{,}904{,}997 \\
Potency data after audit & 1{,}167{,}510 & 1{,}659{,}926 & 2{,}583{,}180 & 5{,}410{,}616 \\
\midrule
Records in audit scope & 1{,}004{,}926 & 1{,}504{,}456 & --- & 2{,}509{,}382 \\
Partial or unavailable review & 576{,}473 & 693{,}992 & --- & 1{,}270{,}465 \\
Auditable source records & 428{,}453 & 810{,}464 & --- & 1{,}238{,}917 \\
\midrule
Retained unchanged & 240{,}362 & 459{,}380 & --- & 699{,}742 \\
Marked for correction & 17{,}516 & 35{,}059 & --- & 52{,}575 \\
Removed & 170{,}575 & 316{,}025 & --- & 486{,}600 \\
\bottomrule
\end{tabular}
\end{table}

\subsection{Coverage comparison with \aqaffinity}
\label{app:coverage}
We compare \genesispotency with the 1{,}415{,}051 training rows of \aqaffinity's release, excluding its validation rows~\cite{sandboxaq2026aqaffinity}, and process both sets identically.
Compounds are identified by the InChIKey of their ChEMBL Structure Pipeline parent~\cite{bento2020chembl}, and proteins are identified by the UniProt accession of their sequence.
PubChem HTS constructs without a full-length match are resolved through UniParc, PDB chain mappings, or UniProt identifier mapping.
Each (protein, compound) pair is counted once across assays, sources, and label types, with PubChem HTS included in both sets.
Target groups follow the ChEMBL protein classification~\cite{zdrazil2024chembl}, with our own family rollup for the 13\% of proteins it does not classify.
An \aqaffinity row matches an audit outcome when its compound, protein, and label agree with a pre-audit \genesispotency data point within 0.01 log units.
Because the audit removes whole assay groups, a match marks a measurement from a rejected group, not a verified error.

Compared with \aqaffinity, \genesispotency has 1.16 times as many proteins (3{,}110 versus 2{,}691), 1.23 times as many compounds (1.00 versus 0.82 million), and 2.78 times as many (protein, compound) pairs, mostly from PubChem HTS.
For pairs with a \pX label, the ratio is 1.24.
Ion channels gain the most proteins (2.5 times), and kinases have 1.5 times as many pairs (Table~\ref{tab:coverage-groups}).

\begin{table}[!t]
\centering\small
\caption{Proteins and (protein, compound) pairs per ChEMBL target group in \aqaffinity and \genesispotency.
Rows follow Figure~\ref{fig:genesispotency}(d), whose ``Other'' bar combines the last five target groups listed here.
The last column counts proteins present in both datasets.}
\label{tab:coverage-groups}
\begin{tabular}{@{}lrrrrr@{}}
\toprule
 & \multicolumn{2}{c}{\aqaffinity} & \multicolumn{2}{c}{\genesispotency} & Both\\
\cmidrule(lr){2-3}\cmidrule(lr){4-5}\cmidrule(l){6-6}
Target group & Proteins & Potency data & Proteins & Potency data & Proteins\\
\midrule
GPCR & 456 & 300{,}203 & 481 & 315{,}592 & 381\\
Kinase & 439 & 216{,}064 & 475 & 324{,}379 & 387\\
Protease & 250 & 117{,}614 & 276 & 170{,}472 & 194\\
Ion channel & 68 & 14{,}078 & 169 & 62{,}766 & 62\\
Epigenetic & 103 & 35{,}646 & 102 & 638{,}164 & 80\\
Transporter & 93 & 36{,}437 & 122 & 101{,}345 & 76\\
Nuclear receptor & 65 & 54{,}577 & 63 & 241{,}087 & 46\\
Other enzyme & 237 & 95{,}807 & 336 & 248{,}811 & 184\\
Hydrolase \& PDE & 334 & 99{,}718 & 323 & 270{,}245 & 205\\
Oxidoreductase & 237 & 47{,}927 & 274 & 622{,}846 & 180\\
Transferase & 206 & 79{,}641 & 253 & 200{,}767 & 147\\
Other & 203 & 152{,}051 & 236 & 274{,}150 & 94\\
\midrule
All groups & 2{,}691 & 1{,}249{,}763 & 3{,}110 & 3{,}470{,}624 & 2{,}036\\
\bottomrule
\end{tabular}

\end{table}
Neither set contains the other: 655 proteins, 0.25 million compounds, and 0.32 million \pX pairs appear only in \aqaffinity, partly because \genesispotency excludes benchmark targets and data dated after 2022.
About 5\% of \aqaffinity's data points (68{,}678) match a measurement our agentic audit removed or marked for correction, so auditing \aqaffinity the same way may be worthwhile.

\subsection{Annotated dataset release}
We are releasing the audited public-source training dataset as \genesispotency for academic and commercial use under the \href{https://creativecommons.org/licenses/by-sa/4.0/}{Creative Commons Attribution-ShareAlike 4.0 International (CC BY-SA 4.0) license}, with annotations attached to its rows.
The license permits reuse, redistribution, and adaptation, subject to attribution, a link to the license, and identification of changes.
Adapted material that is shared must be distributed under the same or a compatible license.
Release documentation and version history will be maintained in the \href{https://github.com/genesistherapeutics/genesispotency}{GenesisPotency GitHub repository}, with dataset artifacts hosted on \href{https://huggingface.co/datasets/genesisml/GenesisPotency}{Hugging Face}.
Each released row identifies its source provenance, audit status, and applied corrections, distinguishing document-level review from direct checking of a sampled compound and identifying partial reviews or unavailable source evidence.
The annotations link rows to their source records and audit decisions, since a source record can contribute to more than one task.
Original endpoints, values, units, relation qualifiers, and normalized labels preserve the context needed to interpret the supervision.
Removed rows and corrections awaiting input preparation are not represented as usable, corrected data points.
The release contains the redistributable public-source portion of the corpus with source attribution.
\internalnote{}

\FloatBarrier
\section{Benchmark-specific training exclusions}
\label{app:deleak}
Following \boltz{2}~\cite{passaro2025boltz2} and IsoDDE~\cite{isomorphic2026isodde}, we
exclude a training target when its sequence aligns to an \openfe, \fepfour or
hit-identification target with at least 90\% identity and the alignment covers
at least 50\% of either sequence.
\caspsixteen and \openbind are deleaked implicitly by the date cutoff. We remove no program targets.

\begin{table}[htbp]
  \centering\scriptsize
  \setlength{\tabcolsep}{2pt}
  \caption{Reported benchmark-specific eligibility of supervised affinity training observations.
  Protein $<90\%$ means no qualifying high-identity match under that method's
  coverage rule. ``No exclusion'' means no panel-specific target exclusion,
  not confirmed training exposure. Label dates refer to supervised affinity or activity
  records, not protein structures. A dash means the criterion was not specified.}
  \label{tab:deleak}
  \begin{tabular}{@{}p{0.12\linewidth}p{0.12\linewidth}p{0.11\linewidth}p{0.20\linewidth}p{0.12\linewidth}p{0.12\linewidth}p{0.13\linewidth}@{}}
    \toprule
    \multirow{2}{*}{Model} & \multirow{2}{*}{Label date} &
      \multicolumn{5}{c}{Target similarity} \\
    \cmidrule(lr){3-7}
    & & HitID (7) & \openfe{} / \fepfour & \caspsixteen & \openbind & \partnercohort \\
    \midrule
    \ruby & Before 2023 & Protein $<90\%$ & Protein $<90\%$ & No exclusion & No exclusion & No exclusion \\
    IsoDDE~\cite{isomorphic2026isodde} & Before 2023 & --- & Protein $<90\%$ & No exclusion & --- & --- \\
    \boltz{2}~\cite{passaro2025boltz2} & --- & --- & Outside $\geq90\%$ test clusters & No exclusion & --- & --- \\
    \nesso~\cite{nesso2026} & --- & --- & Protein $<90\%$ & --- & --- & --- \\
    TerraBind~\cite{rossi2026terrabind} & --- & --- & --- & --- & --- & --- \\
    \aqaffinity~\cite{sandboxaq2026aqaffinity} & --- & --- & --- & --- & --- & --- \\
    \ligunity~\cite{ligunity2025} & --- & --- & --- & --- & --- & --- \\
    \bottomrule
  \end{tabular}
\end{table}

The published descriptions behind Table~\ref{tab:deleak} state:
\begin{quote}\small
\textbf{IsoDDE:} ``cutoff date of 01-January-2023'' and ``we removed any target with a sequence identity of 90\% or greater to either of the sets.''~\cite[Secs.~2.2 and 4.5]{isomorphic2026isodde}
\end{quote}
\begin{quote}\small
\textbf{Boltz-2:} ``training protein that falls into a cluster shared with a validation or test protein'' and ``except for CASP16 and the Recursion internal assays''~\cite[App.~D.2.1]{passaro2025boltz2}.
\end{quote}
\begin{quote}\small
\textbf{Nesso-1:} ``removing any training datapoint with $\geq90\%$ protein sequence similarity to the FEP+ benchmark''~\cite[Sec.~2.1]{nesso2026}
\end{quote}
\begin{quote}\small
\textbf{TerraBind (structure data):} ``All structures from the Protein Data Bank (PDB)'' and ``released before 2021-09-30''~\cite[Sec.~2.2.2]{rossi2026terrabind}
\end{quote}
\begin{quote}\small
\textbf{AQAffinity release:} ``the L3000 target has high similarity with at least one protein target in the training set (i.e. single point mutation).''~\cite{sandboxaq2026aqaffinity}
\end{quote}
\begin{quote}\small
\textbf{LigUnity:} ``assays similar to those included in the FEP benchmarks are excluded in the pre-training stage'' and ``Ligands in the FEP benchmarks are also removed''~\cite{ligunity2025}
\end{quote}

CASP16 released the L1000 and L3000 ligand targets in May 2024, and the archive dates
their experimental affinity files to November--December 2024~\cite{casp16pharmaarchive}.
This post-cutoff benchmark release does not preclude earlier measurements on the same
protein or compounds in training.

For \boltz{2}, MMseqs clusters training and test proteins using at least 90\% identity
over at least 1\% of the longer sequence, then removes training proteins in a cluster
containing a test protein~\cite[App.~D.2.1]{passaro2025boltz2}. The authors exempt \caspsixteen from this protein filter,
stating that its data were released after training, but they do not give a complete affinity-label date cutoff. \ruby instead removes
direct alignment matches covering at least half of either sequence.
TerraBind's quoted cutoff applies to structural PDB data, but its paper does not give a corresponding cutoff or CASP16 target-exclusion rule for affinity labels~\cite{rossi2026terrabind}.
\ligunity's benchmark-ligand and matched-assay holdouts apply to its narrower
JACS/Merck panel~\cite{ligunity2025}. No global
benchmark-ligand holdout is reported for the other methods. \aqaffinity reports a training
target one residue away from CASP16 L3000~\cite{sandboxaq2026aqaffinity}.

\FloatBarrier

\clearpage

\section{Supporting evaluation methods}
\label{app:evaluation}

\subsection{Input preparation and scientific identity}
\label{app:preparation}
We key benchmark identity to the source compound's InChIKey and carry the prepared
molecular state for inference separately. This avoids treating two
protonation states as independent experimental compounds. Frozen prepared-source
records bind each scored observation to its protein input, ligand state, and cached
representation. Joins require exact identity coverage and reject conflicting
structures or nonfinite scores. Protein sequences and any selected pocket constraints
or structural templates come from each benchmark's prepared inputs, and programs use their production input configurations.

\subsection{Hit-identification benchmark choice}
DUD-E pairs known actives with computationally selected decoys, whose selection can introduce chemical biases that models exploit without learning target-specific interactions~\cite{mysinger2012dude,chen2019dudebias}.
MF-PCBA uses experimental HTS measurements~\cite{buterez2023mfpcba}, but compounds recurring across public screening datasets may also occur in a model's training data, even after target-level exclusions, motivating an explicit audit of compound overlap.
Enrichment within a fixed HTS library can also have limited practical value when the collection is screened in bulk, since ranking compounds does not reduce the number tested.
Prioritization is more directly actionable when experimental capacity limits which compounds can be tested or advanced to follow-up.

\subsection{Public potency cohorts and labels}
\label{app:public-cohorts}
\paragraph{Cohort membership.}
\openfe contains 873 observations from 58 source series spanning 39 targets, and each series is scored separately.
Compound identity is unique within each source series, and observations of the same compound in different source series are retained.
\fepfour contains 87 observations across CDK2, JNK1, p38 and TYK2, with one series per target.
\caspsixteen contains 123 autotaxin and 17 chymase observations, giving 1{,}100 observations across 64 source series in total~\cite{baumann2026openfe,wang2015fep,chen2023abfe,gilson2026casp16}.
Correlations are computed within source series and averaged within each benchmark using compound-count weights, as specified in Section~\ref{sec:methods:metrics}.
Table~\ref{tab:a2} gives series-level counts and results.

\paragraph{Label sources and conversion.}
For \openfe and \fepfour, we use the 873 and 87 \texttt{pchembl\_value} labels, respectively, from the pinned IsoDDE benchmark extraction~\cite{isomorphic2026isodde}.
The extraction is fixed at revision \href{https://github.com/bamattsson/paper-identifying_and_addressing_data_leakage/tree/7e6077bd36ca610b0e5d4d9814bbfab2ec874f7e/data/out}{\texttt{7e6077bd}} of the benchmark data repository, and source-series and compound identities are matched before scoring.
These supplied \pX values are retained without an additional energy-to-concentration conversion.
The \fepfour values are verified against $\pX=-\log_{10}(C)$ using the recorded concentrations and units.
For \caspsixteen, we instead use the organizers' original IC$_{50}$ measurements, with $\pX=6-\log_{10}(\mathrm{IC}_{50}/\si{\micro M})$.
The nine autotaxin observations at the 10\,\si{\micro M} assay ceiling are marked as censored and enter correlation calculations at the boundary value $\pX=5$.
The autotaxin protein input corresponds to the rat construct.

\paragraph{OpenBind.}
We retain the full 494-entry affinity benchmark after the source's measurement quality control, using the released experimental p$K_D$ values, where $\mathrm{p}K_D=-\log_{10}(K_D)$~\cite{nelen2026openbind}.
These measurements were obtained against CVA16 2A protease, the experimental surrogate for EV-A71 2A protease.
Released comparator predictions are matched by structure identifier to the same 494 benchmark entries, and their experimental labels are checked against our frozen cohort.
For methods with per-compound predictions, Spearman $\rho$ and its bootstrap uncertainty use this full cohort.
Cited aggregate results are identified separately in Table~\ref{tab:a2secondary}.

\subsection{Partner-program cohorts and recalibration}
\label{app:partner-cohorts}
The \partnercohort evaluation uses shared scored rolling cohorts totalling 22{,}036 assay--compound records across eight assays and five targets.
Biochemical and cellular measurements remain separate tasks, and a compound measured in both can contribute to each task.
Aggregate results give every assay equal weight.

Correlations include censored measurements at their reported \pX bounds.
Zero-shot correlations use the full scored cohorts.
Recalibrated correlations use the intersection of finite rolling predictions across all methods.
\ruby and comparison methods are scored on the same subset within each setting.

At each rolling cutoff, assay-specific recalibration uses only earlier exact measurements from that assay, and the base prediction models remain fixed.
We give comparison methods a separate affine fit at each cutoff.
For every method, recalibrated Pearson $r$, Spearman $\rho$ and Kendall $\tau_b$ are computed after pooling the eligible rolling predictions within each assay.
Kendall $\tau_b$ accounts for ties in both measured and predicted values.
Fold-specific slopes and intercepts can change the pooled correlations, even when each individual fit has a positive slope.

\subsection{Rolling temporal evaluation window}
\label{app:rolling-temporal}
\begin{figure}[htbp]
  \centering
  \includegraphics[width=\linewidth]{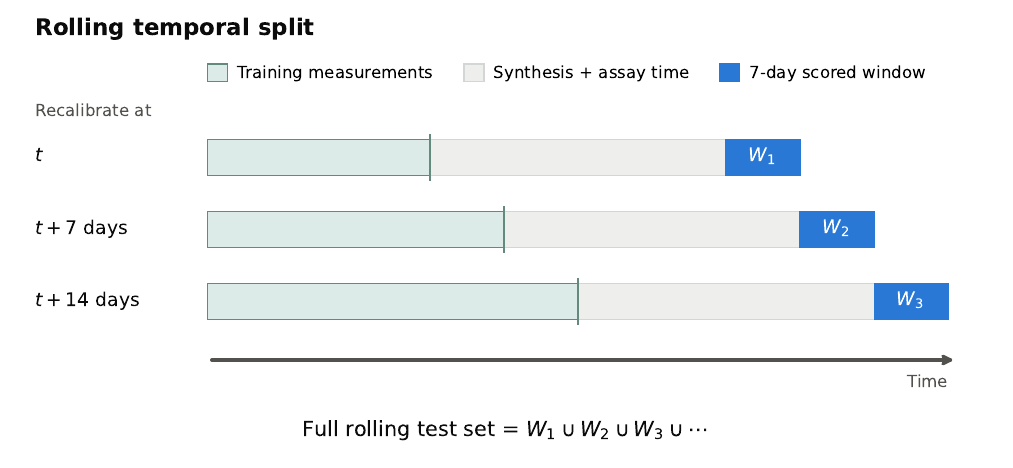}
  \caption{Rolling temporal evaluation within one assay.
  Each row shows a fold: recalibration is refit using only exact training measurements available at its cutoff, while the base prediction model remains fixed.
  A 28-day synthesis and assay interval separates the cutoff from the seven-day scored window.
  The cutoff advances by seven days, expanding the training history.
  The full rolling temporal test set is the union of the scored windows, $\bigcup_k W_k$.
  Each window uses its own recalibration fit, and each compound contributes once.
  Three consecutive eligible folds are shown schematically, with earlier history truncated.
  Bar widths indicate elapsed time, not compound counts.}
  \label{fig:rolling-temporal}
\end{figure}

Related time-split validation has been used to evaluate prediction confidence on internal pharmaceutical assay data~\cite{kearnes2024ordinal}.
We construct folds separately for each of the eight \partnercohort assays.
Figure~\ref{fig:rolling-temporal} illustrates the split and per-window recalibration.
\internalnote{}
Let $t_k$ be a fold cutoff, advancing in seven-day steps from the assay's earliest exact compound date.
We assign each compound the date of its earliest exact
measurement, or its earliest censored measurement if none is exact. The fold's
training history contains compounds measured by $t_k$, and its scored window
contains compounds whose assigned date falls in
$[t_k+28\,\mathrm{d},\,t_k+35\,\mathrm{d})$. Thus each prediction uses a
28-day gap after its training cutoff, and the seven-day windows do not overlap.
We begin scoring only once at least 20 compounds have dated, uncensored
measurements in the training history.
The fold grid is determined from uncensored dates.
Compounds with only censored measurements may enter a scored window, but they do not enter the exact-label recalibration fits.

We pool the eligible scored windows for each assay and evaluate every method on a shared eligible cohort for each metric and setting.
Each compound contributes once, at its first eligible window.
At each cutoff, models are recalibrated using only the assay history available by that date.
Zero-shot correlations use the full shared scored cohort.

\subsection{Bootstrap details}
\label{app:bootstrap}
For each partner-program metric, we select the comparator with the best aggregate point estimate and retain that comparator throughout the paired bootstrap.
We draw 1{,}000 bootstrap samples with replacement within each target or assay, retaining its observed sample size.
For paired comparisons, we share compound draws across methods.
For public targets containing several source series, we resample compounds within each source series before pooling, keeping series sizes fixed.
This preserves pairing with historical source-series summaries.
Predictions, including rolling recalibration fits, remain fixed throughout resampling.
The resulting uncertainty is conditional on the evaluated targets, assays and fitted models.
The \ruby (Max) setting used for the hit-identification enrichment curve was selected after inspecting its \nef{1} score.
The associated paired bootstrap $p$-values condition on this selection and do not adjust for it.

Within each paired bootstrap draw, we define \ruby's advantage as \ruby minus comparator for correlations and comparator minus \ruby for MAE.
We report the one-sided bootstrap tail fraction $p$ as the fraction of valid draws with negative advantage, counting ties as half.
Stars indicate $p \leq 0.05$ ($^{*}$), $p \leq 0.01$ ($^{**}$), and $p \leq 0.001$ ($^{***}$).
Appendix Table~\ref{tab:partner-paired} reports the paired comparisons.

For aggregates computed from bootstrap draws, we apply the point estimate's fixed membership and weights to each draw: equal target weights for hit identification, compound-count weights for public-potency correlations, and equal assay weights for partner programs.
The bootstrap standard error is the sample standard deviation of the valid aggregate draws.
An aggregate draw is invalid whenever any member correlation is undefined.
We change neither membership nor weights to rescue a draw.
Where aggregate error bars are propagated from independently bootstrapped assay estimates, we use $\mathrm{SE}_{\mathrm{mean}}=\sqrt{\sum_{a=1}^{A}\mathrm{SE}_a^2}/A$ for $A$ equally weighted assays.
This propagation assumes independence across assays and does not account for dependence when assays share compounds.

For the chemical-similarity analysis in Figure~\ref{fig:partner-similarity}, we resample compounds separately within each assay and similarity bin, using the metric-specific eligible cohort.
These draws are not paired across methods.
For each assay and bin, we estimate one standard error as the width of the central 95\% bootstrap percentile interval divided by $2\times1.96$.
Undefined correlation draws are excluded, and an uncertainty estimate requires at least 500 valid draws.
We propagate these assay-level estimates to the equal-assay mean using the formula above, and the shading shows the mean plus or minus this estimated standard error.

\subsection{Comparator recipes and coverage}
\label{app:comparators}

\paragraph{Where every comparator number comes from.}

\noindent \gnina has no complete \caspsixteen aggregate because the frozen benchmark carries no receptor geometry for its 17 chymase compounds.
We omit that aggregate rather than report autotaxin alone as the full benchmark.
FEP+ and \openfe RBFE appear only on the two benchmarks for which published predictions are available.
We assume that IsoDDE's \caspsixteen cohort matches ours from its stated use of the \boltz{2} protocol, but the available report does not establish an exact compound match (Appendix~\ref{app:published}).
TerraBind's \caspsixteen aggregate is our compound-weighted derivation from its two published target values.
ApherisAffinity contributes cited results for all four public-potency benchmarks, and \nesso's OpenBind value is also cited rather than re-scored.
OpenBind's other comparator prediction sources are detailed in Table~\ref{tab:a2secondary}.
All partner-program comparator predictions were generated within the partner evaluations.

\paragraph{Configuration.}
Table~\ref{tab:comparator-config} gives the versions and settings of our public screening and \openfe, \fepfour and \caspsixteen runs.
The OpenBind results re-scored from published predictions retain their source inference settings.
Partner-specific preparation and inference records are linked from \texttt{paper/RUN\_PROVENANCE.md}.

\begin{table}[htbp]
\centering\footnotesize
\caption{Comparator versions and inference settings for runs generated here on the seven screening targets and the \openfe, \fepfour and \caspsixteen benchmarks.
A dash in the Method column continues the method above.
Settings may differ from those used for published OpenBind predictions or partner evaluations.}
\label{tab:comparator-config}
\begin{tabular}{@{}p{0.19\linewidth}p{0.16\linewidth}p{0.23\linewidth}p{0.34\linewidth}@{}}
\toprule
Method & Version & Setting & Value \\
\midrule
\boltz{2}~\cite{passaro2025boltz2} & 2025 release & Models & Structure and affinity checkpoints \\
-- & & Hit identification & 3 recycles and 2 structure samples \\
-- & & Public potency & 5 recycles and 5 structure samples \\
-- & & Shared sampling & 200 structure steps and 5 affinity samples \\
-- & & Output & Molar \pX conversion and molecular-weight correction \\
\midrule
\nesso~\cite{nesso2026} & v1.0.0 & Model assets & Released affinity ensemble \\
\midrule
\aqaffinity~\cite{sandboxaq2026aqaffinity} & v0.0.1 (2026) & Models & OpenFold3 and affinity checkpoints \\
-- & & Sampling & 1 seed and 5 diffusion samples \\
-- & & Conditioning & Templates enabled \\
\midrule
\ligunity~\cite{ligunity2025} & 2025 release & Hit identification & Dedicated screening checkpoint \\
-- & & Potency and programs & 5 pocket-ranking and 5 protein-ranking checkpoints \\
\midrule
\gnina~\cite{mcnutt2025gnina13} & v1.3.3, locally modified & Search & 9 starting conformers, 10 poses, exhaustiveness 1, seed 42 \\
-- & & Receptor box & Reference-ligand autobox with 6\,\AA{} padding \\
-- & & Scoring & Negative empirical docking energy, best pose/state per compound \\
-- & & Limits & 600-s docking and 150-s conformer timeouts and 0.25\,\AA{} pose deduplication \\
-- & & Clash handling & Weight 1, severe penetration 0.2, multiplier 3, maximum 0.5 and polar-hydrogen refinement \\
\bottomrule
\end{tabular}
\end{table}

\FloatBarrier

\section{Hit-identification and runtime results}
\label{app:hitid}
Table~\ref{tab:cost-claims} gives relevant published reference timings for different workloads and hardware, with a citation for each claim.
\begingroup\def\pptablecaption{Published timing claims for other workloads and hardware.
These values are provided as context and are not points in Figure~\ref{fig:cost}.}%
{\footnotesize
\begin{longtable}{@{}p{1.0in}p{1.4in}p{3.85in}@{}}
\caption{\pptablecaption{}}\label{tab:cost-claims}\\
\toprule
Method & Reported timing & Workload and source\\
\midrule
\endfirsthead
\multicolumn{3}{@{}l}{\textit{continued}}\\
\toprule
Method & Reported timing & Workload and source\\
\midrule
\endhead
\bottomrule
\endfoot
Boltz-2 & 20 GPU-s / compound & FEP+ 4-target subset. GPU unstated. Boltz-2 paper, Supp. Table 12~\cite{passaro2025boltz2}\\
Nesso-1 & 1.0-2.7 s / compound & One H100, 300-1000 residues. Embedding load excluded. Nesso-1 paper\\
LigUnity & 2.4 ms / compound & RTX A6000, 10k ligands. Ligand embeddings included. LigUnity paper, Fig. S7\\
AnewAffinity & 3.4 s / ligand & Amortized over the FEP+ 4-target set. GPU unstated. AnewAffinity paper\\
GNINA 1.3 & 16.8 s / compound & RTX 3080, default CNN ensemble, PDBbind core. GNINA 1.3 paper\\
OpenFE / FEP+ & 3-12 GPU-h / edge & one RBFE transformation. OpenFE / FEP+ published per-edge costs\\
\end{longtable}
}
\endgroup

\subsection{Per-target enrichment}
\begingroup\def\pptablecaption{Per-target hit identification: $n$ and $n_{\mathrm{active}}$ count scored compounds and positive labels.
nEF at 0.2\%, 0.5\%, 1\%, 2\%, 5\%, and 10\%, and \bedroc at $\alpha = 80.5$ and $\alpha = 20$ show point estimates $\pm$ one bootstrap standard error. The opening block gives equal-target means of the per-target nEF values, and \bedroc is reported per target only. Bold marks the best method in each metric row, compared as printed.}%
{\scriptsize
\setlength{\tabcolsep}{2pt}
\begin{longtable}{@{}lrrlrrrrrrr@{}}
\caption{\pptablecaption{} \statustag{hitid}}\label{tab:a1}\\
\toprule
Target & $n$ & $n_{\mathrm{active}}$ & Metric & \rotatebox{70}{\ruby} & \rotatebox{70}{Nesso-1} & \rotatebox{70}{Boltz-2} & \rotatebox{70}{GNINA} & \rotatebox{70}{LigUnity} & \rotatebox{70}{AQAffinity} & \rotatebox{70}{Mol. weight}\\
\midrule
\endfirsthead
\multicolumn{11}{@{}l}{\textit{continued}}\\
\toprule
Target & $n$ & $n_{\mathrm{active}}$ & Metric & \rotatebox{70}{\ruby} & \rotatebox{70}{Nesso-1} & \rotatebox{70}{Boltz-2} & \rotatebox{70}{GNINA} & \rotatebox{70}{LigUnity} & \rotatebox{70}{AQAffinity} & \rotatebox{70}{Mol. weight}\\
\midrule
\endhead
\bottomrule
\endfoot
\multirow{8}{*}{\textit{equal-target mean}} & 9{,}364 & 1{,}008 & nEF@0.2\% & \phantom{-}\textbf{0.59}{\scriptsize$\pm$0.11} & \phantom{-}0.25{\scriptsize$\pm$0.09} & \phantom{-}0.21{\scriptsize$\pm$0.11} & \phantom{-}0.07{\scriptsize$\pm$0.05} & -0.16{\scriptsize\phantom{$\pm$0.00}} & \phantom{-}0.32{\scriptsize$\pm$0.13} & -0.01{\scriptsize\phantom{$\pm$0.00}}\\*
 &  &  & nEF@0.5\% & \phantom{-}\textbf{0.50}{\scriptsize$\pm$0.07} & \phantom{-}0.17{\scriptsize$\pm$0.08} & \phantom{-}0.30{\scriptsize$\pm$0.09} & \phantom{-}0.11{\scriptsize$\pm$0.05} & -0.08{\scriptsize\phantom{$\pm$0.00}} & \phantom{-}0.08{\scriptsize$\pm$0.08} & \phantom{-}0.15{\scriptsize\phantom{$\pm$0.00}}\\*
 &  &  & nEF@1\% & \phantom{-}\textbf{0.43}{\scriptsize$\pm$0.06} & \phantom{-}0.18{\scriptsize$\pm$0.06} & \phantom{-}0.23{\scriptsize$\pm$0.07} & \phantom{-}0.13{\scriptsize$\pm$0.06} & -0.03{\scriptsize$\pm$0.06} & \phantom{-}0.07{\scriptsize$\pm$0.07} & \phantom{-}0.15{\scriptsize$\pm$0.06}\\*
 &  &  & nEF@2\% & \phantom{-}\textbf{0.40}{\scriptsize$\pm$0.04} & \phantom{-}0.20{\scriptsize$\pm$0.05} & \phantom{-}0.22{\scriptsize$\pm$0.05} & \phantom{-}0.11{\scriptsize$\pm$0.04} & \phantom{-}0.01{\scriptsize\phantom{$\pm$0.00}} & \phantom{-}0.13{\scriptsize$\pm$0.05} & \phantom{-}0.15{\scriptsize\phantom{$\pm$0.00}}\\*
 &  &  & nEF@5\% & \phantom{-}\textbf{0.31}{\scriptsize$\pm$0.03} & \phantom{-}0.18{\scriptsize$\pm$0.03} & \phantom{-}0.22{\scriptsize$\pm$0.03} & \phantom{-}0.15{\scriptsize$\pm$0.03} & \phantom{-}0.00{\scriptsize$\pm$0.03} & \phantom{-}0.13{\scriptsize$\pm$0.03} & \phantom{-}0.19{\scriptsize$\pm$0.03}\\*
 &  &  & nEF@10\% & \phantom{-}\textbf{0.27}{\scriptsize$\pm$0.03} & \phantom{-}0.24{\scriptsize$\pm$0.03} & \phantom{-}0.21{\scriptsize$\pm$0.02} & \phantom{-}0.13{\scriptsize$\pm$0.03} & \phantom{-}0.01{\scriptsize\phantom{$\pm$0.00}} & \phantom{-}0.15{\scriptsize$\pm$0.03} & \phantom{-}0.16{\scriptsize\phantom{$\pm$0.00}}\\*
 &  &  & BEDROC$_{80.5}$ & -- & -- & -- & -- & -- & -- & --\\*
 &  &  & BEDROC$_{20}$ & -- & -- & -- & -- & -- & -- & --\\
\midrule
\multirow{8}{*}{\texttt{ampc}} & 1{,}447 & 298 & nEF@0.2\% & \phantom{-}0.58{\scriptsize$\pm$0.40} & \phantom{-}\textbf{1.00}{\scriptsize$\pm$0.26} & \phantom{-}0.58{\scriptsize$\pm$0.36} & -0.26{\scriptsize$\pm$0.02} & -0.26{\scriptsize$\pm$0.07} & \phantom{-}0.16{\scriptsize$\pm$0.39} & -0.26{\scriptsize$\pm$0.02}\\*
 &  &  & nEF@0.5\% & \phantom{-}\textbf{0.82}{\scriptsize$\pm$0.19} & \phantom{-}0.60{\scriptsize$\pm$0.20} & \phantom{-}0.46{\scriptsize$\pm$0.24} & -0.26{\scriptsize$\pm$0.05} & -0.26{\scriptsize$\pm$0.12} & -0.08{\scriptsize$\pm$0.18} & -0.26{\scriptsize$\pm$0.08}\\*
 &  &  & nEF@1\% & \phantom{-}\textbf{0.73}{\scriptsize$\pm$0.15} & \phantom{-}0.62{\scriptsize$\pm$0.15} & \phantom{-}0.46{\scriptsize$\pm$0.17} & -0.17{\scriptsize$\pm$0.08} & -0.17{\scriptsize$\pm$0.09} & -0.08{\scriptsize$\pm$0.11} & -0.08{\scriptsize$\pm$0.15}\\*
 &  &  & nEF@2\% & \phantom{-}0.48{\scriptsize$\pm$0.12} & \phantom{-}0.50{\scriptsize$\pm$0.13} & \phantom{-}\textbf{0.57}{\scriptsize$\pm$0.11} & -0.22{\scriptsize$\pm$0.06} & -0.09{\scriptsize$\pm$0.07} & -0.04{\scriptsize$\pm$0.10} & \phantom{-}0.17{\scriptsize$\pm$0.11}\\*
 &  &  & nEF@5\% & \phantom{-}\textbf{0.49}{\scriptsize$\pm$0.07} & \phantom{-}0.43{\scriptsize$\pm$0.07} & \phantom{-}0.41{\scriptsize$\pm$0.08} & -0.03{\scriptsize$\pm$0.06} & -0.15{\scriptsize$\pm$0.04} & \phantom{-}0.06{\scriptsize$\pm$0.07} & \phantom{-}0.14{\scriptsize$\pm$0.07}\\*
 &  &  & nEF@10\% & \phantom{-}0.44{\scriptsize$\pm$0.05} & \phantom{-}\textbf{0.46}{\scriptsize$\pm$0.05} & \phantom{-}0.29{\scriptsize$\pm$0.05} & -0.02{\scriptsize$\pm$0.04} & -0.07{\scriptsize$\pm$0.04} & \phantom{-}0.08{\scriptsize$\pm$0.04} & \phantom{-}0.11{\scriptsize$\pm$0.05}\\*
 &  &  & BEDROC$_{80.5}$ & \phantom{-}\textbf{0.67}{\scriptsize$\pm$0.08} & \phantom{-}0.63{\scriptsize$\pm$0.08} & \phantom{-}0.58{\scriptsize$\pm$0.08} & \phantom{-}0.08{\scriptsize$\pm$0.04} & \phantom{-}0.09{\scriptsize$\pm$0.04} & \phantom{-}0.19{\scriptsize$\pm$0.06} & \phantom{-}0.25{\scriptsize$\pm$0.06}\\*
 &  &  & BEDROC$_{20}$ & \phantom{-}\textbf{0.57}{\scriptsize$\pm$0.04} & \phantom{-}0.56{\scriptsize$\pm$0.04} & \phantom{-}0.48{\scriptsize$\pm$0.04} & \phantom{-}0.17{\scriptsize$\pm$0.03} & \phantom{-}0.14{\scriptsize$\pm$0.03} & \phantom{-}0.25{\scriptsize$\pm$0.03} & \phantom{-}0.32{\scriptsize$\pm$0.04}\\
\midrule
\multirow{8}{*}{\texttt{d4\_drd4}} & 548 & 80 & nEF@0.2\% & \phantom{-}\textbf{1.00}{\scriptsize$\pm$0.50} & \phantom{-}\textbf{1.00}{\scriptsize$\pm$0.34} & -0.17{\scriptsize$\pm$0.36} & -0.17{\scriptsize$\pm$0.14} & -0.17{\scriptsize$\pm$0.49} & \phantom{-}\textbf{1.00}{\scriptsize$\pm$0.55} & -0.17{\scriptsize$\pm$0.35}\\*
 &  &  & nEF@0.5\% & \phantom{-}\textbf{0.61}{\scriptsize$\pm$0.32} & \phantom{-}\textbf{0.61}{\scriptsize$\pm$0.31} & \phantom{-}0.22{\scriptsize$\pm$0.33} & -0.17{\scriptsize$\pm$0.18} & \phantom{-}0.22{\scriptsize$\pm$0.33} & \phantom{-}0.22{\scriptsize$\pm$0.35} & \phantom{-}0.22{\scriptsize$\pm$0.29}\\*
 &  &  & nEF@1\% & \phantom{-}\textbf{0.77}{\scriptsize$\pm$0.24} & \phantom{-}0.59{\scriptsize$\pm$0.25} & \phantom{-}0.30{\scriptsize$\pm$0.27} & \phantom{-}0.06{\scriptsize$\pm$0.18} & \phantom{-}0.30{\scriptsize$\pm$0.25} & \phantom{-}0.30{\scriptsize$\pm$0.26} & \phantom{-}0.06{\scriptsize$\pm$0.22}\\*
 &  &  & nEF@2\% & \phantom{-}\textbf{0.68}{\scriptsize$\pm$0.21} & \phantom{-}0.41{\scriptsize$\pm$0.17} & \phantom{-}0.15{\scriptsize$\pm$0.17} & -0.06{\scriptsize$\pm$0.12} & \phantom{-}0.15{\scriptsize$\pm$0.16} & \phantom{-}0.25{\scriptsize$\pm$0.17} & \phantom{-}0.04{\scriptsize$\pm$0.14}\\*
 &  &  & nEF@5\% & \phantom{-}\textbf{0.31}{\scriptsize$\pm$0.11} & \phantom{-}0.26{\scriptsize$\pm$0.10} & \phantom{-}0.26{\scriptsize$\pm$0.11} & \phantom{-}0.00{\scriptsize$\pm$0.09} & \phantom{-}0.09{\scriptsize$\pm$0.09} & \phantom{-}0.09{\scriptsize$\pm$0.10} & \phantom{-}0.09{\scriptsize$\pm$0.09}\\*
 &  &  & nEF@10\% & \phantom{-}\textbf{0.25}{\scriptsize$\pm$0.07} & \phantom{-}\textbf{0.25}{\scriptsize$\pm$0.07} & \phantom{-}0.23{\scriptsize$\pm$0.07} & \phantom{-}0.00{\scriptsize$\pm$0.05} & \phantom{-}0.11{\scriptsize$\pm$0.06} & \phantom{-}0.06{\scriptsize$\pm$0.06} & \phantom{-}0.04{\scriptsize$\pm$0.06}\\*
 &  &  & BEDROC$_{80.5}$ & \phantom{-}\textbf{0.62}{\scriptsize$\pm$0.13} & \phantom{-}0.52{\scriptsize$\pm$0.13} & \phantom{-}0.31{\scriptsize$\pm$0.13} & \phantom{-}0.11{\scriptsize$\pm$0.08} & \phantom{-}0.27{\scriptsize$\pm$0.12} & \phantom{-}0.33{\scriptsize$\pm$0.13} & \phantom{-}0.18{\scriptsize$\pm$0.11}\\*
 &  &  & BEDROC$_{20}$ & \phantom{-}\textbf{0.45}{\scriptsize$\pm$0.07} & \phantom{-}0.39{\scriptsize$\pm$0.07} & \phantom{-}0.34{\scriptsize$\pm$0.07} & \phantom{-}0.15{\scriptsize$\pm$0.05} & \phantom{-}0.24{\scriptsize$\pm$0.06} & \phantom{-}0.24{\scriptsize$\pm$0.06} & \phantom{-}0.19{\scriptsize$\pm$0.05}\\
\midrule
\multirow{8}{*}{\texttt{sigma2\_tmem97}} & 495 & 128 & nEF@0.2\% & \phantom{-}\textbf{1.00}{\scriptsize$\pm$0.15} & -0.35{\scriptsize$\pm$0.39} & \phantom{-}\textbf{1.00}{\scriptsize$\pm$0.42} & \phantom{-}\textbf{1.00}{\scriptsize$\pm$0.14} & -0.35{\scriptsize$\pm$0.18} & -0.35{\scriptsize$\pm$0.24} & -0.35{\scriptsize$\pm$0.61}\\*
 &  &  & nEF@0.5\% & \phantom{-}\textbf{1.00}{\scriptsize$\pm$0.16} & -0.01{\scriptsize$\pm$0.36} & \phantom{-}\textbf{1.00}{\scriptsize$\pm$0.42} & \phantom{-}\textbf{1.00}{\scriptsize$\pm$0.20} & -0.35{\scriptsize$\pm$0.23} & -0.35{\scriptsize$\pm$0.30} & \phantom{-}0.33{\scriptsize$\pm$0.49}\\*
 &  &  & nEF@1\% & \phantom{-}\textbf{0.73}{\scriptsize$\pm$0.21} & \phantom{-}0.12{\scriptsize$\pm$0.30} & \phantom{-}0.46{\scriptsize$\pm$0.32} & \phantom{-}\textbf{0.73}{\scriptsize$\pm$0.28} & -0.08{\scriptsize$\pm$0.31} & -0.08{\scriptsize$\pm$0.31} & \phantom{-}0.46{\scriptsize$\pm$0.31}\\*
 &  &  & nEF@2\% & \phantom{-}\textbf{0.87}{\scriptsize$\pm$0.13} & \phantom{-}0.39{\scriptsize$\pm$0.21} & \phantom{-}0.33{\scriptsize$\pm$0.22} & \phantom{-}0.60{\scriptsize$\pm$0.20} & \phantom{-}0.19{\scriptsize$\pm$0.22} & \phantom{-}0.33{\scriptsize$\pm$0.23} & \phantom{-}0.19{\scriptsize$\pm$0.22}\\*
 &  &  & nEF@5\% & \phantom{-}\textbf{0.78}{\scriptsize$\pm$0.11} & \phantom{-}0.37{\scriptsize$\pm$0.14} & \phantom{-}0.46{\scriptsize$\pm$0.12} & \phantom{-}0.46{\scriptsize$\pm$0.14} & \phantom{-}0.14{\scriptsize$\pm$0.13} & \phantom{-}0.41{\scriptsize$\pm$0.13} & \phantom{-}0.35{\scriptsize$\pm$0.13}\\*
 &  &  & nEF@10\% & \phantom{-}\textbf{0.70}{\scriptsize$\pm$0.08} & \phantom{-}0.45{\scriptsize$\pm$0.09} & \phantom{-}0.46{\scriptsize$\pm$0.09} & \phantom{-}0.27{\scriptsize$\pm$0.09} & \phantom{-}0.19{\scriptsize$\pm$0.09} & \phantom{-}0.30{\scriptsize$\pm$0.09} & \phantom{-}0.35{\scriptsize$\pm$0.09}\\*
 &  &  & BEDROC$_{80.5}$ & \phantom{-}\textbf{0.89}{\scriptsize$\pm$0.08} & \phantom{-}0.45{\scriptsize$\pm$0.14} & \phantom{-}0.59{\scriptsize$\pm$0.14} & \phantom{-}0.73{\scriptsize$\pm$0.12} & \phantom{-}0.30{\scriptsize$\pm$0.12} & \phantom{-}0.39{\scriptsize$\pm$0.14} & \phantom{-}0.46{\scriptsize$\pm$0.14}\\*
 &  &  & BEDROC$_{20}$ & \phantom{-}\textbf{0.79}{\scriptsize$\pm$0.05} & \phantom{-}0.56{\scriptsize$\pm$0.07} & \phantom{-}0.60{\scriptsize$\pm$0.07} & \phantom{-}0.53{\scriptsize$\pm$0.07} & \phantom{-}0.38{\scriptsize$\pm$0.07} & \phantom{-}0.49{\scriptsize$\pm$0.07} & \phantom{-}0.50{\scriptsize$\pm$0.07}\\
\midrule
\multirow{8}{*}{\texttt{cache1\_lrrk2\_wd40}} & 1{,}618 & 101 & nEF@0.2\% & \phantom{-}\textbf{0.64}{\scriptsize$\pm$0.35} & -0.07{\scriptsize$\pm$0.01} & -0.07{\scriptsize$\pm$0.12} & -0.07{\scriptsize$\pm$0.15} & -0.07{\scriptsize$\pm$0.01} & -0.07{\scriptsize$\pm$0.18} & -0.07{\scriptsize$\pm$0.20}\\*
 &  &  & nEF@0.5\% & \phantom{-}\textbf{0.20}{\scriptsize$\pm$0.19} & -0.03{\scriptsize$\pm$0.04} & \phantom{-}0.07{\scriptsize$\pm$0.12} & \phantom{-}0.07{\scriptsize$\pm$0.12} & -0.07{\scriptsize$\pm$0.01} & \phantom{-}\textbf{0.20}{\scriptsize$\pm$0.16} & \phantom{-}0.07{\scriptsize$\pm$0.13}\\*
 &  &  & nEF@1\% & \phantom{-}\textbf{0.13}{\scriptsize$\pm$0.11} & \phantom{-}0.00{\scriptsize$\pm$0.06} & \phantom{-}0.00{\scriptsize$\pm$0.07} & \phantom{-}0.07{\scriptsize$\pm$0.09} & -0.07{\scriptsize$\pm$0.01} & \phantom{-}\textbf{0.13}{\scriptsize$\pm$0.11} & \phantom{-}0.07{\scriptsize$\pm$0.10}\\*
 &  &  & nEF@2\% & \phantom{-}\textbf{0.17}{\scriptsize$\pm$0.08} & \phantom{-}0.03{\scriptsize$\pm$0.05} & \phantom{-}0.07{\scriptsize$\pm$0.06} & \phantom{-}0.13{\scriptsize$\pm$0.08} & -0.03{\scriptsize$\pm$0.03} & \phantom{-}0.13{\scriptsize$\pm$0.07} & \phantom{-}0.07{\scriptsize$\pm$0.07}\\*
 &  &  & nEF@5\% & \phantom{-}\textbf{0.16}{\scriptsize$\pm$0.05} & \phantom{-}0.04{\scriptsize$\pm$0.03} & \phantom{-}0.07{\scriptsize$\pm$0.04} & \phantom{-}0.09{\scriptsize$\pm$0.04} & -0.05{\scriptsize$\pm$0.01} & \phantom{-}0.08{\scriptsize$\pm$0.04} & \phantom{-}0.14{\scriptsize$\pm$0.05}\\*
 &  &  & nEF@10\% & \phantom{-}\textbf{0.16}{\scriptsize$\pm$0.04} & \phantom{-}0.06{\scriptsize$\pm$0.04} & \phantom{-}0.14{\scriptsize$\pm$0.04} & \phantom{-}0.14{\scriptsize$\pm$0.05} & -0.09{\scriptsize$\pm$0.02} & \phantom{-}0.14{\scriptsize$\pm$0.04} & \phantom{-}0.15{\scriptsize$\pm$0.05}\\*
 &  &  & BEDROC$_{80.5}$ & \phantom{-}\textbf{0.22}{\scriptsize$\pm$0.07} & \phantom{-}0.08{\scriptsize$\pm$0.03} & \phantom{-}0.11{\scriptsize$\pm$0.05} & \phantom{-}0.16{\scriptsize$\pm$0.06} & \phantom{-}0.01{\scriptsize$\pm$0.01} & \phantom{-}0.18{\scriptsize$\pm$0.06} & \phantom{-}0.16{\scriptsize$\pm$0.05}\\*
 &  &  & BEDROC$_{20}$ & \phantom{-}\textbf{0.23}{\scriptsize$\pm$0.04} & \phantom{-}0.14{\scriptsize$\pm$0.03} & \phantom{-}0.19{\scriptsize$\pm$0.03} & \phantom{-}0.20{\scriptsize$\pm$0.04} & \phantom{-}0.02{\scriptsize$\pm$0.01} & \phantom{-}0.20{\scriptsize$\pm$0.04} & \phantom{-}0.21{\scriptsize$\pm$0.04}\\
\midrule
\multirow{8}{*}{\texttt{cache2\_nsp13}} & 1{,}954 & 155 & nEF@0.2\% & -0.09{\scriptsize$\pm$0.11} & -0.02{\scriptsize$\pm$0.08} & -0.09{\scriptsize$\pm$0.14} & \phantom{-}\textbf{0.19}{\scriptsize$\pm$0.23} & -0.09{\scriptsize$\pm$0.01} & \phantom{-}\textbf{0.19}{\scriptsize$\pm$0.21} & \phantom{-}\textbf{0.19}{\scriptsize$\pm$0.22}\\*
 &  &  & nEF@0.5\% & \phantom{-}0.02{\scriptsize$\pm$0.12} & \phantom{-}0.00{\scriptsize$\pm$0.06} & \phantom{-}\textbf{0.24}{\scriptsize$\pm$0.16} & \phantom{-}0.13{\scriptsize$\pm$0.13} & -0.09{\scriptsize$\pm$0.01} & \phantom{-}0.02{\scriptsize$\pm$0.11} & \phantom{-}0.13{\scriptsize$\pm$0.13}\\*
 &  &  & nEF@1\% & \phantom{-}0.08{\scriptsize$\pm$0.09} & \phantom{-}0.00{\scriptsize$\pm$0.06} & \phantom{-}\textbf{0.24}{\scriptsize$\pm$0.11} & \phantom{-}0.02{\scriptsize$\pm$0.08} & -0.09{\scriptsize$\pm$0.01} & \phantom{-}0.02{\scriptsize$\pm$0.08} & \phantom{-}0.13{\scriptsize$\pm$0.11}\\*
 &  &  & nEF@2\% & \phantom{-}0.05{\scriptsize$\pm$0.06} & \phantom{-}0.03{\scriptsize$\pm$0.05} & \phantom{-}\textbf{0.25}{\scriptsize$\pm$0.08} & \phantom{-}0.11{\scriptsize$\pm$0.07} & -0.09{\scriptsize$\pm$0.01} & \phantom{-}0.05{\scriptsize$\pm$0.06} & \phantom{-}\textbf{0.25}{\scriptsize$\pm$0.08}\\*
 &  &  & nEF@5\% & \phantom{-}0.00{\scriptsize$\pm$0.03} & \phantom{-}0.12{\scriptsize$\pm$0.04} & \phantom{-}0.18{\scriptsize$\pm$0.05} & \phantom{-}0.16{\scriptsize$\pm$0.04} & -0.08{\scriptsize$\pm$0.01} & \phantom{-}0.06{\scriptsize$\pm$0.04} & \phantom{-}\textbf{0.20}{\scriptsize$\pm$0.05}\\*
 &  &  & nEF@10\% & \phantom{-}0.01{\scriptsize$\pm$0.03} & \phantom{-}0.17{\scriptsize$\pm$0.03} & \phantom{-}0.18{\scriptsize$\pm$0.03} & \phantom{-}0.14{\scriptsize$\pm$0.04} & -0.10{\scriptsize$\pm$0.01} & \phantom{-}0.11{\scriptsize$\pm$0.03} & \phantom{-}\textbf{0.24}{\scriptsize$\pm$0.04}\\*
 &  &  & BEDROC$_{80.5}$ & \phantom{-}0.11{\scriptsize$\pm$0.05} & \phantom{-}0.12{\scriptsize$\pm$0.04} & \phantom{-}\textbf{0.26}{\scriptsize$\pm$0.06} & \phantom{-}0.18{\scriptsize$\pm$0.05} & \phantom{-}0.00{\scriptsize$\pm$0.00} & \phantom{-}0.13{\scriptsize$\pm$0.05} & \phantom{-}0.25{\scriptsize$\pm$0.06}\\*
 &  &  & BEDROC$_{20}$ & \phantom{-}0.11{\scriptsize$\pm$0.03} & \phantom{-}0.22{\scriptsize$\pm$0.03} & \phantom{-}0.27{\scriptsize$\pm$0.03} & \phantom{-}0.24{\scriptsize$\pm$0.03} & \phantom{-}0.02{\scriptsize$\pm$0.01} & \phantom{-}0.18{\scriptsize$\pm$0.03} & \phantom{-}\textbf{0.31}{\scriptsize$\pm$0.04}\\
\midrule
\multirow{8}{*}{\texttt{cache3\_mac1}} & 1{,}661 & 229 & nEF@0.2\% & \phantom{-}\textbf{1.00}{\scriptsize$\pm$0.20} & \phantom{-}0.23{\scriptsize$\pm$0.31} & \phantom{-}0.23{\scriptsize$\pm$0.32} & -0.16{\scriptsize$\pm$0.06} & -0.16{\scriptsize$\pm$0.13} & \phantom{-}0.61{\scriptsize$\pm$0.36} & \phantom{-}0.61{\scriptsize$\pm$0.35}\\*
 &  &  & nEF@0.5\% & \phantom{-}\textbf{0.86}{\scriptsize$\pm$0.19} & \phantom{-}0.02{\scriptsize$\pm$0.14} & \phantom{-}0.13{\scriptsize$\pm$0.20} & -0.01{\scriptsize$\pm$0.12} & -0.01{\scriptsize$\pm$0.14} & \phantom{-}0.28{\scriptsize$\pm$0.22} & \phantom{-}0.57{\scriptsize$\pm$0.22}\\*
 &  &  & nEF@1\% & \phantom{-}\textbf{0.59}{\scriptsize$\pm$0.15} & -0.06{\scriptsize$\pm$0.08} & \phantom{-}0.18{\scriptsize$\pm$0.14} & \phantom{-}0.11{\scriptsize$\pm$0.13} & -0.09{\scriptsize$\pm$0.08} & \phantom{-}0.11{\scriptsize$\pm$0.14} & \phantom{-}0.39{\scriptsize$\pm$0.14}\\*
 &  &  & nEF@2\% & \phantom{-}\textbf{0.51}{\scriptsize$\pm$0.10} & \phantom{-}0.04{\scriptsize$\pm$0.07} & \phantom{-}0.16{\scriptsize$\pm$0.09} & \phantom{-}0.12{\scriptsize$\pm$0.09} & -0.02{\scriptsize$\pm$0.07} & \phantom{-}0.05{\scriptsize$\pm$0.09} & \phantom{-}0.37{\scriptsize$\pm$0.10}\\*
 &  &  & nEF@5\% & \phantom{-}\textbf{0.34}{\scriptsize$\pm$0.06} & \phantom{-}0.02{\scriptsize$\pm$0.05} & \phantom{-}0.06{\scriptsize$\pm$0.05} & \phantom{-}0.15{\scriptsize$\pm$0.06} & \phantom{-}0.08{\scriptsize$\pm$0.05} & \phantom{-}0.08{\scriptsize$\pm$0.05} & \phantom{-}0.30{\scriptsize$\pm$0.07}\\*
 &  &  & nEF@10\% & \phantom{-}\textbf{0.25}{\scriptsize$\pm$0.04} & \phantom{-}0.10{\scriptsize$\pm$0.03} & \phantom{-}0.07{\scriptsize$\pm$0.04} & \phantom{-}0.17{\scriptsize$\pm$0.04} & \phantom{-}0.04{\scriptsize$\pm$0.03} & \phantom{-}0.13{\scriptsize$\pm$0.04} & \phantom{-}0.13{\scriptsize$\pm$0.04}\\*
 &  &  & BEDROC$_{80.5}$ & \phantom{-}\textbf{0.57}{\scriptsize$\pm$0.08} & \phantom{-}0.14{\scriptsize$\pm$0.05} & \phantom{-}0.25{\scriptsize$\pm$0.07} & \phantom{-}0.23{\scriptsize$\pm$0.06} & \phantom{-}0.13{\scriptsize$\pm$0.05} & \phantom{-}0.24{\scriptsize$\pm$0.07} & \phantom{-}0.46{\scriptsize$\pm$0.08}\\*
 &  &  & BEDROC$_{20}$ & \phantom{-}\textbf{0.40}{\scriptsize$\pm$0.04} & \phantom{-}0.21{\scriptsize$\pm$0.03} & \phantom{-}0.22{\scriptsize$\pm$0.03} & \phantom{-}0.28{\scriptsize$\pm$0.04} & \phantom{-}0.18{\scriptsize$\pm$0.03} & \phantom{-}0.24{\scriptsize$\pm$0.03} & \phantom{-}0.33{\scriptsize$\pm$0.04}\\
\midrule
\multirow{8}{*}{\texttt{cache4\_cblb\_tkb}} & 1{,}641 & 17 & nEF@0.2\% & -0.01{\scriptsize$\pm$0.00} & -0.01{\scriptsize$\pm$0.00} & -0.01{\scriptsize$\pm$0.00} & -0.01{\scriptsize$\pm$0.02} & -0.01{\scriptsize$\pm$0.00} & \phantom{-}\textbf{0.66}{\scriptsize$\pm$0.35} & -0.01{\scriptsize$\pm$0.00}\\*
 &  &  & nEF@0.5\% & -0.01{\scriptsize$\pm$0.00} & -0.01{\scriptsize$\pm$0.00} & -0.01{\scriptsize$\pm$0.00} & -0.01{\scriptsize$\pm$0.06} & -0.01{\scriptsize$\pm$0.00} & \phantom{-}\textbf{0.24}{\scriptsize$\pm$0.17} & -0.01{\scriptsize$\pm$0.00}\\*
 &  &  & nEF@1\% & -0.01{\scriptsize$\pm$0.00} & -0.01{\scriptsize$\pm$0.00} & -0.01{\scriptsize$\pm$0.03} & \phantom{-}0.05{\scriptsize$\pm$0.07} & -0.01{\scriptsize$\pm$0.00} & \phantom{-}\textbf{0.12}{\scriptsize$\pm$0.09} & -0.01{\scriptsize$\pm$0.00}\\*
 &  &  & nEF@2\% & \phantom{-}0.04{\scriptsize$\pm$0.05} & -0.01{\scriptsize$\pm$0.02} & \phantom{-}0.04{\scriptsize$\pm$0.07} & \phantom{-}\textbf{0.10}{\scriptsize$\pm$0.09} & -0.02{\scriptsize$\pm$0.00} & \phantom{-}\textbf{0.10}{\scriptsize$\pm$0.08} & -0.02{\scriptsize$\pm$0.03}\\*
 &  &  & nEF@5\% & \phantom{-}0.07{\scriptsize$\pm$0.09} & \phantom{-}0.06{\scriptsize$\pm$0.07} & \phantom{-}0.07{\scriptsize$\pm$0.08} & \phantom{-}\textbf{0.20}{\scriptsize$\pm$0.11} & -0.05{\scriptsize$\pm$0.00} & \phantom{-}0.13{\scriptsize$\pm$0.10} & \phantom{-}0.07{\scriptsize$\pm$0.09}\\*
 &  &  & nEF@10\% & \phantom{-}0.09{\scriptsize$\pm$0.11} & \phantom{-}\textbf{0.22}{\scriptsize$\pm$0.11} & \phantom{-}0.09{\scriptsize$\pm$0.10} & \phantom{-}\textbf{0.22}{\scriptsize$\pm$0.12} & \phantom{-}0.02{\scriptsize$\pm$0.09} & \phantom{-}\textbf{0.22}{\scriptsize$\pm$0.12} & \phantom{-}0.09{\scriptsize$\pm$0.11}\\*
 &  &  & BEDROC$_{80.5}$ & \phantom{-}0.02{\scriptsize$\pm$0.02} & \phantom{-}0.02{\scriptsize$\pm$0.01} & \phantom{-}0.05{\scriptsize$\pm$0.03} & \phantom{-}0.10{\scriptsize$\pm$0.06} & \phantom{-}0.00{\scriptsize$\pm$0.00} & \phantom{-}\textbf{0.18}{\scriptsize$\pm$0.10} & \phantom{-}0.02{\scriptsize$\pm$0.01}\\*
 &  &  & BEDROC$_{20}$ & \phantom{-}0.10{\scriptsize$\pm$0.05} & \phantom{-}0.12{\scriptsize$\pm$0.05} & \phantom{-}0.11{\scriptsize$\pm$0.06} & \phantom{-}\textbf{0.20}{\scriptsize$\pm$0.08} & \phantom{-}0.04{\scriptsize$\pm$0.02} & \phantom{-}\textbf{0.20}{\scriptsize$\pm$0.09} & \phantom{-}0.10{\scriptsize$\pm$0.05}\\
\end{longtable}
}
\endgroup

\ligunity's screening checkpoint is retained in Table~\ref{tab:a1}: its
seven-target equal-target \nef{1} is $-0.029 \pm 0.058$ and \nef{5}
is $-0.005 \pm 0.025$. Its ensemble inference averaged $0.000829$ seconds
per ligand across the seven seeded 32-ligand timing samples, weighted by the
full target cohorts, and this includes ligand embedding time.

\FloatBarrier

\section{Per-series and per-target potency results}
\label{app:public}
\paragraph{Choice of correlation metric.}
Figure~\ref{fig:series} uses Pearson $r$ for \openfe, \fepfour, and \caspsixteen to support comparison with published results: IsoDDE reports Pearson correlations for all three benchmarks, ApherisAffinity plots Pearson correlations for all three, and TerraBind reports per-target Pearson correlations for \caspsixteen~\cite{isomorphic2026isodde,bucher2026apherisaffinity,rossi2026terrabind}.
The \caspsixteen organizers instead assess affinity using a compound-count-weighted mean of per-target Kendall $\tau$~\cite{gilson2026casp16}, and we report this metric alongside the individual target values in Table~\ref{tab:a2secondary}.
For OpenBind, Figure~\ref{fig:series} and Table~\ref{tab:a2secondary} use Spearman $\rho$, following the published ranking benchmark~\cite{nelen2026openbind}.
We do not use OpenBind's RMSE metric for this zero-shot comparison because the evaluated methods do not all produce calibrated predictions on a common p$K_D$ scale.
Absolute errors would therefore reflect differences in score scale and offset as well as predictive performance.
Spearman $\rho$ assesses ranking without requiring assay-specific calibration.

\subsection{Published per-compound predictions}\label{app:published}
We join FEP+~\cite{ross2023maximal} and \openfe RBFE~\cite{baumann2026openfe} predictions to frozen rows by system, series and ligand name, then re-score them with the same paired bootstrap as re-run methods.
Neither covers \caspsixteen.
The source authors shift their relative networks to each series' experimental mean.
Within-series correlations are invariant to these shifts, whereas pooling independently anchored series can credit experimental offsets as predicted between-series differences.
We therefore compute correlations within the original source series for every re-scored method and report no MAE for the relative free-energy methods.
The \openfe means are $r=0.72$ for FEP+ (published $0.72$) and $0.65$ for RBFE (published $0.63$).
Our pinned RBFE predictions include the \href{https://github.com/OpenFreeEnergy/IndustryBenchmarks2024/commit/829b4f94ee86716a23da1340f5bf68b0cc2f9512}{September 2025 upstream correction} to the 40 PFKFB3 predictions.
The earlier release reproduces $r=0.63$ under source-series aggregation.

\paragraph{Boltz-2 reruns.}
Our Boltz-2 runs give $r=0.56$, $0.59$ and $0.77$ on \openfe, \caspsixteen and \fepfour, respectively, compared with the original report's $0.62$, $0.65$ and $0.66$~\cite{passaro2025boltz2}.
These are separate inference runs scored on our frozen cohorts, and we have not reconciled the remaining differences with the published aggregates.
Changing the OpenFE aggregation alone does not explain them.

\paragraph{Boltz-2 sampling check.}
On the three public-potency benchmarks, increasing structure recycles from three to five and structure samples from two to five produced small changes in compound-weighted Pearson $r$.
The paired differences (higher minus lower sampling) were $-0.004$ on \openfe (95\% bootstrap interval $[-0.017, 0.010]$), $-0.001$ on \fepfour ($[-0.023, 0.020]$), and $-0.006$ on \caspsixteen ($[-0.037, 0.023]$).
Inputs, checkpoints, affinity sampling and molecular-weight correction were unchanged.
This sensitivity analysis retains the historical 39-target grouping for \openfe.
The main evaluation instead computes correlations within its 58 source series.
We used 1{,}000 paired compound bootstrap draws, of which 842 were valid for \openfe and all 1{,}000 for the other two benchmarks.
Each setting had one inference run, so these intervals reflect compound resampling rather than inference-seed variability.
We detected no difference on these benchmarks, but the comparison does not establish equivalence or test enrichment on the screening targets.
The current public-potency figures use five recycles and five structure samples, and the screening figures retain three recycles and two structure samples.

\paragraph{OpenFE bootstrap validity.}
Fourteen source series contain only three to five compounds, so resampling can produce constant labels or predictions and an undefined correlation.
An aggregate draw is retained only when every constituent series has a defined correlation.
We do not drop series or renormalize their weights within a draw.
For Pearson $r$, 455 of 1,000 draws are valid for \ruby and all re-scored comparators except molecular weight, which retains 426.
The reported standard errors are conditional on these valid draws.

\paragraph{Cited and derived results.}
IsoDDE~\cite{isomorphic2026isodde} reports 873 observations over 39 \openfe targets, matching our counts, but supplies no target manifest.
It says these public sets follow the \boltz{2} protocol~\cite{passaro2025boltz2}, which weights cross-assay regression averages by compound count, but does not state the correlation units or weights used for IsoDDE's own reported Pearson means.
The count of 39 unique proteins does not establish whether source series sharing a protein were pooled before computing correlations.
Its \fepfour set names our four targets and 87 compounds.
It does not name its \caspsixteen compounds.
\boltz{2} calls the set 140 pairs across two targets, and the \caspsixteen assessment lists autotaxin (123) and chymase (17) as the affinity targets~\cite{gilson2026casp16}, so we assume the same cohort.
IsoDDE takes the other methods' plotted values from \boltz{2}, so agreement between those bars and \boltz{2}'s reported aggregates cannot establish IsoDDE's own weighting.
No per-compound IsoDDE predictions are available here for matched bootstrap errors or an equal-target sensitivity check.

TerraBind~\cite{rossi2026terrabind} reports Pearson 0.73 on autotaxin (123 compounds) and 0.45 on chymase (17), read from figure labels, but no pooled \caspsixteen value or pooling rule.
Our compound-weighted derivation, $(0.73\times123+0.45\times17)/140=0.696$, is not a quoted aggregate.

AnewAffinity~\cite{anew2026anewdde} explicitly weights system-level metrics by ligand count on the eight-system, 199-ligand \textsc{jacs} subset of \openfe, but does not report on the full \openfe panel or the other two series sources.
Its reported Pearson value is a weighted mean of within-system $r^2$, which cannot be converted to our mean Pearson $r$ by taking its square root.

\begingroup\def\pptablecaption{Per-series Pearson $r$ for each method with per-compound predictions, $\pm$ one bootstrap standard error, with each benchmark's compound-weighted mean first. \openfe rows are its 58 original source series, grouped by source collection.
\fepfour and \caspsixteen carry one series per target. Pearson is undefined for a series with fewer than three compounds or constant labels or predictions. Bold marks the best method in each row, compared as printed. Significance marks are omitted from the public-potency figure and this table.}%
{\footnotesize
\setlength{\tabcolsep}{1.2pt}
\begin{longtable}{@{}lrrrrrrrrrr@{}}
\caption{\pptablecaption{} \statustag{series}}\label{tab:a2}\\
\toprule
Series / target & $n$ & \multicolumn{1}{c}{\rotatebox{70}{\ruby}} & \multicolumn{1}{c}{\rotatebox{70}{FEP+$^\ddagger$}} & \multicolumn{1}{c}{\rotatebox{70}{Nesso-1}} & \multicolumn{1}{c}{\rotatebox{70}{OpenFE RBFE$^\ddagger$}} & \multicolumn{1}{c}{\rotatebox{70}{Boltz-2}} & \multicolumn{1}{c}{\rotatebox{70}{LigUnity}} & \multicolumn{1}{c}{\rotatebox{70}{Mol. weight}} & \multicolumn{1}{c}{\rotatebox{70}{AQAffinity}} & \multicolumn{1}{c}{\rotatebox{70}{GNINA}}\\
\midrule
\endfirsthead
\multicolumn{11}{@{}l}{\textit{continued}}\\
\toprule
Series / target & $n$ & \multicolumn{1}{c}{\rotatebox{70}{\ruby}} & \multicolumn{1}{c}{\rotatebox{70}{FEP+$^\ddagger$}} & \multicolumn{1}{c}{\rotatebox{70}{Nesso-1}} & \multicolumn{1}{c}{\rotatebox{70}{OpenFE RBFE$^\ddagger$}} & \multicolumn{1}{c}{\rotatebox{70}{Boltz-2}} & \multicolumn{1}{c}{\rotatebox{70}{LigUnity}} & \multicolumn{1}{c}{\rotatebox{70}{Mol. weight}} & \multicolumn{1}{c}{\rotatebox{70}{AQAffinity}} & \multicolumn{1}{c}{\rotatebox{70}{GNINA}}\\
\midrule
\endhead
\bottomrule
\endfoot
\multicolumn{11}{@{}l}{\textbf{OpenFE}}\\
\hspace{0.4em}\textit{compound-weighted} & 873 & \phantom{-}\textbf{0.74}{\scriptsize$\pm$0.02} & \phantom{-}0.72{\scriptsize$\pm$0.02} & \phantom{-}0.66{\scriptsize$\pm$0.02} & \phantom{-}0.65{\scriptsize$\pm$0.02} & \phantom{-}0.56{\scriptsize$\pm$0.02} & \phantom{-}0.63{\scriptsize$\pm$0.02} & \phantom{-}0.38{\scriptsize$\pm$0.03} & \phantom{-}0.33{\scriptsize$\pm$0.03} & \phantom{-}0.25{\scriptsize$\pm$0.03}\\
\multicolumn{11}{@{}l}{\hspace{0.4em}\textit{Charge annihilation}}\\*
\hspace{1.2em}\texttt{\scriptsize cdk2} & 3 & \phantom{-}0.97{\scriptsize$\pm$0.01} & \phantom{-}0.98{\scriptsize$\pm$0.01} & \phantom{-}0.92{\scriptsize$\pm$0.04} & \phantom{-}\textbf{1.00}{\scriptsize$\pm$0.00} & -0.10{\scriptsize$\pm$0.81} & -0.70{\scriptsize$\pm$0.13} & -0.47{\scriptsize$\pm$0.88} & \phantom{-}0.91{\scriptsize$\pm$0.04} & \phantom{-}0.57{\scriptsize$\pm$0.19}\\
\hspace{1.2em}\texttt{\scriptsize dlk} & 5 & \phantom{-}0.86{\scriptsize$\pm$0.26} & \phantom{-}\textbf{0.99}{\scriptsize$\pm$0.01} & \phantom{-}0.45{\scriptsize$\pm$0.49} & \phantom{-}0.64{\scriptsize$\pm$0.51} & \phantom{-}0.44{\scriptsize$\pm$0.51} & \phantom{-}0.95{\scriptsize$\pm$0.08} & \phantom{-}0.26{\scriptsize$\pm$0.57} & \phantom{-}0.83{\scriptsize$\pm$0.31} & \phantom{-}0.16{\scriptsize$\pm$0.62}\\
\hspace{1.2em}\texttt{\scriptsize egfr} & 5 & -\textbf{0.13}{\scriptsize$\pm$0.52} & -0.47{\scriptsize$\pm$0.42} & -0.42{\scriptsize$\pm$0.67} & -0.37{\scriptsize$\pm$0.47} & -0.20{\scriptsize$\pm$0.54} & -0.51{\scriptsize$\pm$0.68} & -0.18{\scriptsize$\pm$0.63} & -0.40{\scriptsize$\pm$0.50} & -0.23{\scriptsize$\pm$0.55}\\
\hspace{1.2em}\texttt{\scriptsize ephx2} & 4 & \phantom{-}0.89{\scriptsize$\pm$0.47} & \phantom{-}0.93{\scriptsize$\pm$0.47} & \phantom{-}\textbf{1.00}{\scriptsize$\pm$0.00} & \phantom{-}0.82{\scriptsize$\pm$0.48} & \phantom{-}\textbf{1.00}{\scriptsize$\pm$0.00} & \phantom{-}0.39{\scriptsize$\pm$0.65} & \phantom{-}0.94{\scriptsize$\pm$0.03} & \phantom{-}0.96{\scriptsize$\pm$0.47} & \phantom{-}0.93{\scriptsize$\pm$0.04}\\
\hspace{1.2em}\texttt{\scriptsize irak4\_s2} & 5 & \phantom{-}0.82{\scriptsize$\pm$0.31} & \phantom{-}0.85{\scriptsize$\pm$0.41} & \phantom{-}0.88{\scriptsize$\pm$0.34} & \phantom{-}0.83{\scriptsize$\pm$0.36} & \phantom{-}0.91{\scriptsize$\pm$0.45} & \phantom{-}\textbf{0.93}{\scriptsize$\pm$0.29} & \phantom{-}0.63{\scriptsize$\pm$0.75} & \phantom{-}0.22{\scriptsize$\pm$0.57} & \phantom{-}0.76{\scriptsize$\pm$0.71}\\
\hspace{1.2em}\texttt{\scriptsize irak4\_s3} & 4 & \phantom{-}0.75{\scriptsize$\pm$0.45} & \phantom{-}0.42{\scriptsize$\pm$0.50} & -0.22{\scriptsize$\pm$0.67} & \phantom{-}0.67{\scriptsize$\pm$0.82} & \phantom{-}0.53{\scriptsize$\pm$0.48} & \phantom{-}\textbf{0.89}{\scriptsize$\pm$0.07} & -0.86{\scriptsize$\pm$0.09} & \phantom{-}0.64{\scriptsize$\pm$0.46} & -0.78{\scriptsize$\pm$0.45}\\
\hspace{1.2em}\texttt{\scriptsize itk} & 4 & \phantom{-}0.63{\scriptsize$\pm$0.58} & \phantom{-}0.37{\scriptsize$\pm$0.65} & \phantom{-}\textbf{0.99}{\scriptsize$\pm$0.07} & \phantom{-}0.23{\scriptsize$\pm$0.59} & \phantom{-}0.65{\scriptsize$\pm$0.43} & \phantom{-}0.46{\scriptsize$\pm$0.65} & \phantom{-}0.12{\scriptsize$\pm$0.73} & \phantom{-}0.13{\scriptsize$\pm$0.65} & \phantom{-}0.49{\scriptsize$\pm$0.55}\\
\hspace{1.2em}\texttt{\scriptsize jak1} & 6 & \phantom{-}0.51{\scriptsize$\pm$0.37} & \phantom{-}0.88{\scriptsize$\pm$0.25} & \phantom{-}\textbf{0.94}{\scriptsize$\pm$0.21} & \phantom{-}0.55{\scriptsize$\pm$0.34} & \phantom{-}0.47{\scriptsize$\pm$0.54} & \phantom{-}0.73{\scriptsize$\pm$0.58} & \phantom{-}0.86{\scriptsize$\pm$0.11} & \phantom{-}0.36{\scriptsize$\pm$0.45} & \phantom{-}0.54{\scriptsize$\pm$0.27}\\
\hspace{1.2em}\texttt{\scriptsize jnk1} & 3 & \phantom{-}0.86{\scriptsize$\pm$0.84} & \phantom{-}0.92{\scriptsize$\pm$0.04} & \phantom{-}0.08{\scriptsize$\pm$0.83} & \phantom{-}0.69{\scriptsize$\pm$0.83} & \phantom{-}0.11{\scriptsize$\pm$0.84} & \phantom{-}\textbf{0.97}{\scriptsize$\pm$0.01} & -0.97{\scriptsize$\pm$0.01} & -0.97{\scriptsize$\pm$0.01} & -0.64{\scriptsize$\pm$0.82}\\
\hspace{1.2em}\texttt{\scriptsize ptp1b} & 3 & \phantom{-}0.93{\scriptsize$\pm$0.03} & \phantom{-}0.92{\scriptsize$\pm$0.03} & \phantom{-}\textbf{0.99}{\scriptsize$\pm$0.00} & \phantom{-}0.92{\scriptsize$\pm$0.03} & \phantom{-}0.72{\scriptsize$\pm$0.83} & \phantom{-}0.93{\scriptsize$\pm$0.03} & \phantom{-}0.98{\scriptsize$\pm$0.01} & \phantom{-}0.95{\scriptsize$\pm$0.02} & \phantom{-}0.95{\scriptsize$\pm$0.02}\\
\hspace{1.2em}\texttt{\scriptsize thrombin} & 6 & \phantom{-}0.94{\scriptsize$\pm$0.47} & \phantom{-}\textbf{0.97}{\scriptsize$\pm$0.12} & \phantom{-}0.70{\scriptsize$\pm$0.49} & \phantom{-}0.94{\scriptsize$\pm$0.08} & \phantom{-}0.90{\scriptsize$\pm$0.39} & \phantom{-}0.90{\scriptsize$\pm$0.35} & \phantom{-}0.41{\scriptsize$\pm$0.45} & \phantom{-}0.86{\scriptsize$\pm$0.70} & \phantom{-}0.76{\scriptsize$\pm$0.61}\\
\hspace{1.2em}\texttt{\scriptsize tyk2} & 5 & \phantom{-}0.69{\scriptsize$\pm$0.39} & \phantom{-}0.89{\scriptsize$\pm$0.53} & \phantom{-}0.94{\scriptsize$\pm$0.30} & \phantom{-}\textbf{0.97}{\scriptsize$\pm$0.53} & \phantom{-}0.89{\scriptsize$\pm$0.25} & \phantom{-}0.03{\scriptsize$\pm$0.57} & \phantom{-}0.29{\scriptsize$\pm$0.53} & \phantom{-}0.12{\scriptsize$\pm$0.60} & -0.29{\scriptsize$\pm$0.52}\\
\multicolumn{11}{@{}l}{\hspace{0.4em}\textit{Fragments}}\\*
\hspace{1.2em}\texttt{\scriptsize hsp90\_2rings} & 6 & \phantom{-}0.29{\scriptsize$\pm$0.56} & \phantom{-}0.50{\scriptsize$\pm$0.62} & \phantom{-}0.43{\scriptsize$\pm$0.43} & \phantom{-}0.66{\scriptsize$\pm$0.45} & \phantom{-}0.60{\scriptsize$\pm$0.51} & \phantom{-}0.37{\scriptsize$\pm$0.41} & \phantom{-}0.37{\scriptsize$\pm$0.41} & \phantom{-}\textbf{0.72}{\scriptsize$\pm$0.33} & \phantom{-}0.19{\scriptsize$\pm$0.51}\\
\hspace{1.2em}\texttt{\scriptsize hsp90\_single\_ring} & 7 & \phantom{-}0.54{\scriptsize$\pm$0.26} & \phantom{-}0.73{\scriptsize$\pm$0.23} & \phantom{-}0.45{\scriptsize$\pm$0.31} & \phantom{-}0.42{\scriptsize$\pm$0.40} & \phantom{-}\textbf{0.91}{\scriptsize$\pm$0.12} & \phantom{-}0.62{\scriptsize$\pm$0.21} & \phantom{-}0.78{\scriptsize$\pm$0.25} & -0.13{\scriptsize$\pm$0.42} & \phantom{-}0.73{\scriptsize$\pm$0.27}\\
\hspace{1.2em}\texttt{\scriptsize jak2\_set1} & 10 & \phantom{-}0.88{\scriptsize$\pm$0.09} & \phantom{-}0.87{\scriptsize$\pm$0.09} & \phantom{-}0.64{\scriptsize$\pm$0.19} & \phantom{-}0.86{\scriptsize$\pm$0.13} & \phantom{-}0.72{\scriptsize$\pm$0.16} & \phantom{-}\textbf{0.95}{\scriptsize$\pm$0.03} & \phantom{-}0.57{\scriptsize$\pm$0.21} & \phantom{-}0.68{\scriptsize$\pm$0.22} & \phantom{-}0.35{\scriptsize$\pm$0.33}\\
\hspace{1.2em}\texttt{\scriptsize jak2\_set2} & 8 & \phantom{-}\textbf{0.85}{\scriptsize$\pm$0.09} & \phantom{-}0.64{\scriptsize$\pm$0.21} & \phantom{-}0.48{\scriptsize$\pm$0.22} & \phantom{-}0.71{\scriptsize$\pm$0.20} & \phantom{-}0.76{\scriptsize$\pm$0.21} & \phantom{-}0.83{\scriptsize$\pm$0.09} & \phantom{-}0.45{\scriptsize$\pm$0.27} & \phantom{-}0.77{\scriptsize$\pm$0.16} & -0.53{\scriptsize$\pm$0.23}\\
\hspace{1.2em}\texttt{\scriptsize liga} & 11 & \phantom{-}\textbf{0.98}{\scriptsize$\pm$0.01} & \phantom{-}0.97{\scriptsize$\pm$0.04} & \phantom{-}0.95{\scriptsize$\pm$0.02} & \phantom{-}0.96{\scriptsize$\pm$0.05} & \phantom{-}\textbf{0.98}{\scriptsize$\pm$0.02} & \phantom{-}0.90{\scriptsize$\pm$0.08} & \phantom{-}0.96{\scriptsize$\pm$0.04} & \phantom{-}0.96{\scriptsize$\pm$0.07} & \phantom{-}0.80{\scriptsize$\pm$0.14}\\
\hspace{1.2em}\texttt{\scriptsize mcl1} & 12 & \phantom{-}0.52{\scriptsize$\pm$0.18} & \phantom{-}\textbf{0.56}{\scriptsize$\pm$0.23} & \phantom{-}0.23{\scriptsize$\pm$0.25} & \phantom{-}0.52{\scriptsize$\pm$0.22} & \phantom{-}0.42{\scriptsize$\pm$0.17} & \phantom{-}0.39{\scriptsize$\pm$0.23} & \phantom{-}0.26{\scriptsize$\pm$0.27} & \phantom{-}0.20{\scriptsize$\pm$0.29} & -0.38{\scriptsize$\pm$0.17}\\
\hspace{1.2em}\texttt{\scriptsize mup1} & 6 & \phantom{-}0.81{\scriptsize$\pm$0.18} & \phantom{-}0.96{\scriptsize$\pm$0.13} & -0.49{\scriptsize$\pm$0.28} & \phantom{-}\textbf{0.97}{\scriptsize$\pm$0.13} & -0.59{\scriptsize$\pm$0.29} & \phantom{-}0.51{\scriptsize$\pm$0.37} & \phantom{-}\textbf{0.97}{\scriptsize$\pm$0.05} & -0.26{\scriptsize$\pm$0.41} & \phantom{-}0.96{\scriptsize$\pm$0.13}\\
\hspace{1.2em}\texttt{\scriptsize p38} & 6 & \phantom{-}0.82{\scriptsize$\pm$0.37} & \phantom{-}\textbf{0.95}{\scriptsize$\pm$0.07} & \phantom{-}0.82{\scriptsize$\pm$0.27} & \phantom{-}0.70{\scriptsize$\pm$0.40} & \phantom{-}0.72{\scriptsize$\pm$0.36} & \phantom{-}0.89{\scriptsize$\pm$0.11} & \phantom{-}0.80{\scriptsize$\pm$0.32} & \phantom{-}0.66{\scriptsize$\pm$0.29} & \phantom{-}0.77{\scriptsize$\pm$0.19}\\
\hspace{1.2em}\texttt{\scriptsize t4\_lysozyme} & 12 & \phantom{-}0.64{\scriptsize$\pm$0.23} & \phantom{-}0.65{\scriptsize$\pm$0.17} & \phantom{-}0.46{\scriptsize$\pm$0.29} & \phantom{-}0.79{\scriptsize$\pm$0.09} & \phantom{-}0.57{\scriptsize$\pm$0.20} & \phantom{-}\textbf{0.94}{\scriptsize$\pm$0.04} & \phantom{-}0.49{\scriptsize$\pm$0.23} & -0.05{\scriptsize$\pm$0.44} & \phantom{-}0.44{\scriptsize$\pm$0.25}\\
\multicolumn{11}{@{}l}{\hspace{0.4em}\textit{JACS}}\\*
\hspace{1.2em}\texttt{\scriptsize bace} & 36 & \phantom{-}0.75{\scriptsize$\pm$0.08} & \phantom{-}0.62{\scriptsize$\pm$0.11} & \phantom{-}\textbf{0.85}{\scriptsize$\pm$0.05} & \phantom{-}0.55{\scriptsize$\pm$0.12} & \phantom{-}0.26{\scriptsize$\pm$0.15} & \phantom{-}0.80{\scriptsize$\pm$0.05} & -0.15{\scriptsize$\pm$0.13} & -0.14{\scriptsize$\pm$0.12} & -0.40{\scriptsize$\pm$0.11}\\
\hspace{1.2em}\texttt{\scriptsize cdk2} & 16 & \phantom{-}\textbf{0.94}{\scriptsize$\pm$0.02} & \phantom{-}0.65{\scriptsize$\pm$0.14} & \phantom{-}0.77{\scriptsize$\pm$0.08} & \phantom{-}0.77{\scriptsize$\pm$0.08} & \phantom{-}0.86{\scriptsize$\pm$0.05} & \phantom{-}0.79{\scriptsize$\pm$0.09} & \phantom{-}0.48{\scriptsize$\pm$0.19} & \phantom{-}0.71{\scriptsize$\pm$0.11} & \phantom{-}0.62{\scriptsize$\pm$0.15}\\
\hspace{1.2em}\texttt{\scriptsize jnk1} & 21 & \phantom{-}\textbf{0.92}{\scriptsize$\pm$0.04} & \phantom{-}0.84{\scriptsize$\pm$0.05} & \phantom{-}0.70{\scriptsize$\pm$0.11} & \phantom{-}0.76{\scriptsize$\pm$0.07} & \phantom{-}0.76{\scriptsize$\pm$0.05} & \phantom{-}0.63{\scriptsize$\pm$0.15} & \phantom{-}0.39{\scriptsize$\pm$0.25} & \phantom{-}0.52{\scriptsize$\pm$0.11} & -0.54{\scriptsize$\pm$0.18}\\
\hspace{1.2em}\texttt{\scriptsize mcl1} & 42 & \phantom{-}0.88{\scriptsize$\pm$0.03} & \phantom{-}0.70{\scriptsize$\pm$0.09} & \phantom{-}\textbf{0.92}{\scriptsize$\pm$0.02} & \phantom{-}0.70{\scriptsize$\pm$0.07} & \phantom{-}0.89{\scriptsize$\pm$0.03} & \phantom{-}0.65{\scriptsize$\pm$0.07} & \phantom{-}0.55{\scriptsize$\pm$0.11} & \phantom{-}0.61{\scriptsize$\pm$0.10} & \phantom{-}0.26{\scriptsize$\pm$0.14}\\
\hspace{1.2em}\texttt{\scriptsize p38} & 34 & \phantom{-}0.79{\scriptsize$\pm$0.07} & \phantom{-}0.70{\scriptsize$\pm$0.08} & \phantom{-}\textbf{0.87}{\scriptsize$\pm$0.04} & \phantom{-}0.43{\scriptsize$\pm$0.16} & \phantom{-}0.67{\scriptsize$\pm$0.08} & \phantom{-}0.34{\scriptsize$\pm$0.18} & \phantom{-}0.57{\scriptsize$\pm$0.11} & \phantom{-}0.53{\scriptsize$\pm$0.16} & \phantom{-}0.05{\scriptsize$\pm$0.22}\\
\hspace{1.2em}\texttt{\scriptsize ptp1b} & 23 & \phantom{-}0.83{\scriptsize$\pm$0.07} & \phantom{-}\textbf{0.92}{\scriptsize$\pm$0.02} & \phantom{-}0.84{\scriptsize$\pm$0.05} & \phantom{-}0.78{\scriptsize$\pm$0.09} & \phantom{-}0.66{\scriptsize$\pm$0.11} & \phantom{-}0.74{\scriptsize$\pm$0.07} & \phantom{-}0.74{\scriptsize$\pm$0.10} & \phantom{-}0.77{\scriptsize$\pm$0.10} & \phantom{-}0.67{\scriptsize$\pm$0.18}\\
\hspace{1.2em}\texttt{\scriptsize thrombin} & 11 & \phantom{-}\textbf{0.92}{\scriptsize$\pm$0.06} & \phantom{-}0.82{\scriptsize$\pm$0.14} & \phantom{-}0.83{\scriptsize$\pm$0.10} & \phantom{-}0.76{\scriptsize$\pm$0.18} & \phantom{-}0.74{\scriptsize$\pm$0.13} & \phantom{-}0.76{\scriptsize$\pm$0.17} & \phantom{-}0.48{\scriptsize$\pm$0.25} & \phantom{-}\textbf{0.92}{\scriptsize$\pm$0.07} & \phantom{-}0.71{\scriptsize$\pm$0.17}\\
\hspace{1.2em}\texttt{\scriptsize tyk2} & 16 & \phantom{-}0.85{\scriptsize$\pm$0.09} & \phantom{-}\textbf{0.93}{\scriptsize$\pm$0.03} & \phantom{-}0.86{\scriptsize$\pm$0.08} & \phantom{-}0.89{\scriptsize$\pm$0.04} & \phantom{-}0.86{\scriptsize$\pm$0.08} & \phantom{-}0.70{\scriptsize$\pm$0.15} & \phantom{-}0.06{\scriptsize$\pm$0.24} & \phantom{-}0.19{\scriptsize$\pm$0.26} & -0.08{\scriptsize$\pm$0.17}\\
\multicolumn{11}{@{}l}{\hspace{0.4em}\textit{Janssen BACE}}\\*
\hspace{1.2em}\texttt{\scriptsize bace\_ciordia\_prospective} & 9 & \phantom{-}0.70{\scriptsize$\pm$0.23} & \phantom{-}0.77{\scriptsize$\pm$0.16} & \phantom{-}0.78{\scriptsize$\pm$0.14} & \phantom{-}0.76{\scriptsize$\pm$0.15} & -0.05{\scriptsize$\pm$0.30} & \phantom{-}\textbf{0.80}{\scriptsize$\pm$0.15} & \phantom{-}0.57{\scriptsize$\pm$0.29} & \phantom{-}0.02{\scriptsize$\pm$0.30} & \phantom{-}0.35{\scriptsize$\pm$0.29}\\
\hspace{1.2em}\texttt{\scriptsize bace\_p3\_arg368\_in} & 21 & \phantom{-}0.54{\scriptsize$\pm$0.12} & \phantom{-}0.79{\scriptsize$\pm$0.08} & \phantom{-}\textbf{0.86}{\scriptsize$\pm$0.05} & \phantom{-}0.63{\scriptsize$\pm$0.14} & \phantom{-}0.46{\scriptsize$\pm$0.14} & \phantom{-}0.69{\scriptsize$\pm$0.12} & \phantom{-}0.26{\scriptsize$\pm$0.22} & \phantom{-}0.16{\scriptsize$\pm$0.22} & \phantom{-}0.55{\scriptsize$\pm$0.18}\\
\hspace{1.2em}\texttt{\scriptsize ciordia\_retro} & 32 & \phantom{-}0.81{\scriptsize$\pm$0.06} & \phantom{-}0.75{\scriptsize$\pm$0.09} & \phantom{-}\textbf{0.95}{\scriptsize$\pm$0.02} & \phantom{-}0.83{\scriptsize$\pm$0.05} & \phantom{-}0.80{\scriptsize$\pm$0.08} & \phantom{-}0.83{\scriptsize$\pm$0.05} & \phantom{-}0.52{\scriptsize$\pm$0.12} & \phantom{-}0.20{\scriptsize$\pm$0.19} & \phantom{-}0.56{\scriptsize$\pm$0.15}\\
\hspace{1.2em}\texttt{\scriptsize keranen\_p2} & 12 & \phantom{-}0.47{\scriptsize$\pm$0.17} & \phantom{-}0.26{\scriptsize$\pm$0.34} & \phantom{-}0.17{\scriptsize$\pm$0.23} & \phantom{-}0.46{\scriptsize$\pm$0.24} & \phantom{-}0.51{\scriptsize$\pm$0.25} & \phantom{-}\textbf{0.67}{\scriptsize$\pm$0.25} & -0.03{\scriptsize$\pm$0.31} & \phantom{-}0.37{\scriptsize$\pm$0.25} & -0.41{\scriptsize$\pm$0.24}\\
\multicolumn{11}{@{}l}{\hspace{0.4em}\textit{MCS docking}}\\*
\hspace{1.2em}\texttt{\scriptsize hne} & 17 & \phantom{-}\textbf{0.92}{\scriptsize$\pm$0.07} & \phantom{-}0.87{\scriptsize$\pm$0.14} & \phantom{-}0.91{\scriptsize$\pm$0.07} & \phantom{-}0.74{\scriptsize$\pm$0.10} & \phantom{-}0.83{\scriptsize$\pm$0.17} & \phantom{-}0.76{\scriptsize$\pm$0.24} & \phantom{-}0.06{\scriptsize$\pm$0.35} & \phantom{-}0.79{\scriptsize$\pm$0.16} & -0.03{\scriptsize$\pm$0.18}\\
\hspace{1.2em}\texttt{\scriptsize renin} & 29 & \phantom{-}0.74{\scriptsize$\pm$0.09} & \phantom{-}0.68{\scriptsize$\pm$0.09} & \phantom{-}0.83{\scriptsize$\pm$0.08} & \phantom{-}0.74{\scriptsize$\pm$0.10} & \phantom{-}0.75{\scriptsize$\pm$0.10} & \phantom{-}\textbf{0.84}{\scriptsize$\pm$0.06} & \phantom{-}0.33{\scriptsize$\pm$0.17} & \phantom{-}0.59{\scriptsize$\pm$0.12} & \phantom{-}0.53{\scriptsize$\pm$0.14}\\
\multicolumn{11}{@{}l}{\hspace{0.4em}\textit{Merck}}\\*
\hspace{1.2em}\texttt{\scriptsize cdk8} & 32 & \phantom{-}\textbf{0.83}{\scriptsize$\pm$0.08} & \phantom{-}0.75{\scriptsize$\pm$0.10} & \phantom{-}0.81{\scriptsize$\pm$0.06} & \phantom{-}0.80{\scriptsize$\pm$0.06} & \phantom{-}0.68{\scriptsize$\pm$0.09} & \phantom{-}0.72{\scriptsize$\pm$0.07} & \phantom{-}0.57{\scriptsize$\pm$0.14} & \phantom{-}0.27{\scriptsize$\pm$0.18} & \phantom{-}0.71{\scriptsize$\pm$0.11}\\
\hspace{1.2em}\texttt{\scriptsize cmet} & 24 & \phantom{-}0.87{\scriptsize$\pm$0.06} & \phantom{-}\textbf{0.91}{\scriptsize$\pm$0.03} & \phantom{-}0.86{\scriptsize$\pm$0.06} & \phantom{-}0.79{\scriptsize$\pm$0.09} & \phantom{-}0.80{\scriptsize$\pm$0.07} & \phantom{-}0.81{\scriptsize$\pm$0.06} & \phantom{-}0.71{\scriptsize$\pm$0.11} & \phantom{-}0.13{\scriptsize$\pm$0.15} & -0.27{\scriptsize$\pm$0.14}\\
\hspace{1.2em}\texttt{\scriptsize eg5} & 28 & \phantom{-}\textbf{0.77}{\scriptsize$\pm$0.11} & \phantom{-}0.59{\scriptsize$\pm$0.10} & \phantom{-}0.29{\scriptsize$\pm$0.14} & \phantom{-}0.61{\scriptsize$\pm$0.16} & \phantom{-}0.51{\scriptsize$\pm$0.16} & \phantom{-}0.68{\scriptsize$\pm$0.10} & \phantom{-}0.59{\scriptsize$\pm$0.16} & \phantom{-}0.06{\scriptsize$\pm$0.20} & -0.22{\scriptsize$\pm$0.19}\\
\hspace{1.2em}\texttt{\scriptsize hif2a} & 41 & \phantom{-}0.39{\scriptsize$\pm$0.10} & \phantom{-}\textbf{0.78}{\scriptsize$\pm$0.08} & \phantom{-}0.40{\scriptsize$\pm$0.12} & \phantom{-}0.63{\scriptsize$\pm$0.11} & \phantom{-}0.44{\scriptsize$\pm$0.11} & \phantom{-}0.50{\scriptsize$\pm$0.12} & \phantom{-}0.20{\scriptsize$\pm$0.10} & \phantom{-}0.12{\scriptsize$\pm$0.14} & \phantom{-}0.45{\scriptsize$\pm$0.15}\\
\hspace{1.2em}\texttt{\scriptsize pfkfb3} & 40 & \phantom{-}\textbf{0.81}{\scriptsize$\pm$0.05} & \phantom{-}0.76{\scriptsize$\pm$0.06} & \phantom{-}0.64{\scriptsize$\pm$0.09} & \phantom{-}0.64{\scriptsize$\pm$0.10} & \phantom{-}0.22{\scriptsize$\pm$0.16} & \phantom{-}0.39{\scriptsize$\pm$0.12} & \phantom{-}0.28{\scriptsize$\pm$0.15} & -0.04{\scriptsize$\pm$0.13} & \phantom{-}0.53{\scriptsize$\pm$0.09}\\
\hspace{1.2em}\texttt{\scriptsize shp2} & 26 & \phantom{-}0.88{\scriptsize$\pm$0.05} & \phantom{-}0.71{\scriptsize$\pm$0.08} & \phantom{-}\textbf{0.92}{\scriptsize$\pm$0.03} & \phantom{-}0.06{\scriptsize$\pm$0.24} & \phantom{-}0.35{\scriptsize$\pm$0.15} & \phantom{-}0.44{\scriptsize$\pm$0.14} & \phantom{-}0.29{\scriptsize$\pm$0.14} & \phantom{-}0.03{\scriptsize$\pm$0.20} & \phantom{-}0.41{\scriptsize$\pm$0.12}\\
\hspace{1.2em}\texttt{\scriptsize syk} & 44 & \phantom{-}\textbf{0.71}{\scriptsize$\pm$0.07} & \phantom{-}\textbf{0.71}{\scriptsize$\pm$0.08} & \phantom{-}0.46{\scriptsize$\pm$0.14} & \phantom{-}0.67{\scriptsize$\pm$0.10} & \phantom{-}0.42{\scriptsize$\pm$0.16} & \phantom{-}0.44{\scriptsize$\pm$0.16} & \phantom{-}0.28{\scriptsize$\pm$0.11} & \phantom{-}0.05{\scriptsize$\pm$0.15} & \phantom{-}0.53{\scriptsize$\pm$0.13}\\
\hspace{1.2em}\texttt{\scriptsize tnks2} & 27 & \phantom{-}0.56{\scriptsize$\pm$0.10} & \phantom{-}0.54{\scriptsize$\pm$0.18} & \phantom{-}0.77{\scriptsize$\pm$0.08} & \phantom{-}0.20{\scriptsize$\pm$0.20} & \phantom{-}0.64{\scriptsize$\pm$0.13} & \phantom{-}\textbf{0.83}{\scriptsize$\pm$0.07} & \phantom{-}0.44{\scriptsize$\pm$0.18} & \phantom{-}0.51{\scriptsize$\pm$0.18} & \phantom{-}0.39{\scriptsize$\pm$0.16}\\
\multicolumn{11}{@{}l}{\hspace{0.4em}\textit{Miscellaneous}}\\*
\hspace{1.2em}\texttt{\scriptsize btk} & 6 & \phantom{-}0.82{\scriptsize$\pm$0.25} & \phantom{-}0.88{\scriptsize$\pm$0.29} & \phantom{-}0.56{\scriptsize$\pm$0.45} & \phantom{-}\textbf{0.94}{\scriptsize$\pm$0.24} & \phantom{-}0.69{\scriptsize$\pm$0.36} & \phantom{-}0.32{\scriptsize$\pm$0.52} & \phantom{-}0.74{\scriptsize$\pm$0.29} & \phantom{-}0.54{\scriptsize$\pm$0.31} & \phantom{-}0.78{\scriptsize$\pm$0.26}\\
\hspace{1.2em}\texttt{\scriptsize cdk8} & 10 & -0.23{\scriptsize$\pm$0.32} & \phantom{-}0.66{\scriptsize$\pm$0.15} & \phantom{-}0.51{\scriptsize$\pm$0.33} & \phantom{-}0.64{\scriptsize$\pm$0.27} & \phantom{-}0.10{\scriptsize$\pm$0.37} & \phantom{-}\textbf{0.72}{\scriptsize$\pm$0.14} & \phantom{-}0.35{\scriptsize$\pm$0.44} & \phantom{-}0.50{\scriptsize$\pm$0.25} & -0.36{\scriptsize$\pm$0.32}\\
\hspace{1.2em}\texttt{\scriptsize faah} & 24 & \phantom{-}0.79{\scriptsize$\pm$0.10} & \phantom{-}\textbf{0.85}{\scriptsize$\pm$0.08} & \phantom{-}0.40{\scriptsize$\pm$0.15} & \phantom{-}0.77{\scriptsize$\pm$0.11} & \phantom{-}0.53{\scriptsize$\pm$0.12} & \phantom{-}0.52{\scriptsize$\pm$0.18} & \phantom{-}0.61{\scriptsize$\pm$0.15} & \phantom{-}0.50{\scriptsize$\pm$0.10} & \phantom{-}0.27{\scriptsize$\pm$0.21}\\
\hspace{1.2em}\texttt{\scriptsize galectin} & 26 & \phantom{-}\textbf{0.89}{\scriptsize$\pm$0.04} & \phantom{-}0.87{\scriptsize$\pm$0.04} & \phantom{-}0.72{\scriptsize$\pm$0.12} & \phantom{-}0.73{\scriptsize$\pm$0.11} & \phantom{-}0.72{\scriptsize$\pm$0.09} & \phantom{-}0.30{\scriptsize$\pm$0.18} & \phantom{-}0.42{\scriptsize$\pm$0.21} & \phantom{-}0.73{\scriptsize$\pm$0.13} & \phantom{-}0.65{\scriptsize$\pm$0.13}\\
\hspace{1.2em}\texttt{\scriptsize hiv1\_protease} & 13 & \phantom{-}\textbf{0.52}{\scriptsize$\pm$0.18} & -0.01{\scriptsize$\pm$0.34} & -0.14{\scriptsize$\pm$0.25} & -0.14{\scriptsize$\pm$0.26} & -0.44{\scriptsize$\pm$0.24} & \phantom{-}0.38{\scriptsize$\pm$0.28} & -0.08{\scriptsize$\pm$0.31} & -0.47{\scriptsize$\pm$0.20} & \phantom{-}0.15{\scriptsize$\pm$0.28}\\
\multicolumn{11}{@{}l}{\hspace{0.4em}\textit{Scaffold hopping}}\\*
\hspace{1.2em}\texttt{\scriptsize bace1} & 3 & \phantom{-}0.91{\scriptsize$\pm$0.84} & \phantom{-}\textbf{0.96}{\scriptsize$\pm$0.02} & -0.94{\scriptsize$\pm$0.03} & \phantom{-}0.82{\scriptsize$\pm$0.08} & -0.44{\scriptsize$\pm$0.81} & \phantom{-}0.93{\scriptsize$\pm$0.84} & -0.62{\scriptsize$\pm$0.82} & -0.42{\scriptsize$\pm$0.81} & -0.88{\scriptsize$\pm$0.05}\\
\hspace{1.2em}\texttt{\scriptsize factor\_xa} & 3 & -0.94{\scriptsize$\pm$0.03} & -0.98{\scriptsize$\pm$0.01} & \phantom{-}0.98{\scriptsize$\pm$0.85} & \phantom{-}\textbf{1.00}{\scriptsize$\pm$0.00} & -0.95{\scriptsize$\pm$0.84} & -0.95{\scriptsize$\pm$0.02} & -0.23{\scriptsize$\pm$0.82} & -0.43{\scriptsize$\pm$0.82} & -0.96{\scriptsize$\pm$0.84}\\
\multicolumn{11}{@{}l}{\hspace{0.4em}\textit{Water}}\\*
\hspace{1.2em}\texttt{\scriptsize brd4} & 8 & \phantom{-}\textbf{0.94}{\scriptsize$\pm$0.04} & \phantom{-}0.05{\scriptsize$\pm$0.39} & \phantom{-}0.83{\scriptsize$\pm$0.24} & \phantom{-}0.66{\scriptsize$\pm$0.23} & \phantom{-}0.77{\scriptsize$\pm$0.25} & \phantom{-}0.92{\scriptsize$\pm$0.22} & -0.64{\scriptsize$\pm$0.43} & \phantom{-}0.62{\scriptsize$\pm$0.28} & -0.03{\scriptsize$\pm$0.52}\\
\hspace{1.2em}\texttt{\scriptsize chk1} & 13 & \phantom{-}0.71{\scriptsize$\pm$0.10} & \phantom{-}0.78{\scriptsize$\pm$0.11} & \phantom{-}0.76{\scriptsize$\pm$0.10} & \phantom{-}0.86{\scriptsize$\pm$0.09} & \phantom{-}0.84{\scriptsize$\pm$0.08} & \phantom{-}\textbf{0.95}{\scriptsize$\pm$0.03} & \phantom{-}0.71{\scriptsize$\pm$0.25} & \phantom{-}0.56{\scriptsize$\pm$0.18} & \phantom{-}0.46{\scriptsize$\pm$0.20}\\
\hspace{1.2em}\texttt{\scriptsize hsp90\_kung} & 11 & \phantom{-}0.81{\scriptsize$\pm$0.09} & \phantom{-}\textbf{0.83}{\scriptsize$\pm$0.14} & \phantom{-}0.74{\scriptsize$\pm$0.17} & \phantom{-}0.68{\scriptsize$\pm$0.22} & \phantom{-}0.78{\scriptsize$\pm$0.12} & \phantom{-}0.81{\scriptsize$\pm$0.13} & \phantom{-}0.69{\scriptsize$\pm$0.20} & \phantom{-}0.62{\scriptsize$\pm$0.20} & -0.36{\scriptsize$\pm$0.22}\\
\hspace{1.2em}\texttt{\scriptsize hsp90\_woodhead} & 4 & \phantom{-}0.97{\scriptsize$\pm$0.02} & \phantom{-}\textbf{1.00}{\scriptsize$\pm$0.00} & \phantom{-}0.25{\scriptsize$\pm$0.75} & \phantom{-}0.72{\scriptsize$\pm$0.61} & \phantom{-}0.71{\scriptsize$\pm$0.50} & \phantom{-}0.51{\scriptsize$\pm$0.71} & -0.74{\scriptsize$\pm$0.75} & \phantom{-}0.95{\scriptsize$\pm$0.49} & -0.34{\scriptsize$\pm$0.61}\\
\hspace{1.2em}\texttt{\scriptsize scyt\_dehyd} & 7 & \phantom{-}0.66{\scriptsize$\pm$0.27} & \phantom{-}\textbf{0.88}{\scriptsize$\pm$0.22} & -0.31{\scriptsize$\pm$0.42} & \phantom{-}0.81{\scriptsize$\pm$0.17} & -0.27{\scriptsize$\pm$0.37} & \phantom{-}0.58{\scriptsize$\pm$0.30} & \phantom{-}0.79{\scriptsize$\pm$0.14} & -0.09{\scriptsize$\pm$0.40} & \phantom{-}0.06{\scriptsize$\pm$0.43}\\
\hspace{1.2em}\texttt{\scriptsize taf12} & 8 & \phantom{-}0.84{\scriptsize$\pm$0.16} & \phantom{-}\textbf{0.91}{\scriptsize$\pm$0.22} & \phantom{-}0.62{\scriptsize$\pm$0.21} & \phantom{-}0.86{\scriptsize$\pm$0.42} & \phantom{-}0.82{\scriptsize$\pm$0.11} & \phantom{-}0.61{\scriptsize$\pm$0.29} & -0.40{\scriptsize$\pm$0.40} & \phantom{-}0.45{\scriptsize$\pm$0.31} & \phantom{-}0.21{\scriptsize$\pm$0.37}\\
\hspace{1.2em}\texttt{\scriptsize thrombin} & 21 & \phantom{-}0.70{\scriptsize$\pm$0.11} & \phantom{-}0.75{\scriptsize$\pm$0.08} & \phantom{-}0.67{\scriptsize$\pm$0.12} & \phantom{-}0.46{\scriptsize$\pm$0.17} & \phantom{-}0.59{\scriptsize$\pm$0.16} & \phantom{-}\textbf{0.77}{\scriptsize$\pm$0.09} & \phantom{-}0.50{\scriptsize$\pm$0.13} & \phantom{-}0.41{\scriptsize$\pm$0.19} & -0.02{\scriptsize$\pm$0.23}\\
\hspace{1.2em}\texttt{\scriptsize urokinase} & 4 & \phantom{-}0.65{\scriptsize$\pm$0.50} & \phantom{-}0.93{\scriptsize$\pm$0.06} & \phantom{-}0.70{\scriptsize$\pm$0.86} & \phantom{-}0.77{\scriptsize$\pm$0.09} & \phantom{-}0.52{\scriptsize$\pm$0.79} & \phantom{-}0.72{\scriptsize$\pm$0.84} & \phantom{-}0.83{\scriptsize$\pm$0.44} & \phantom{-}\textbf{0.94}{\scriptsize$\pm$0.46} & \phantom{-}0.92{\scriptsize$\pm$0.05}\\
\midrule
\multicolumn{11}{@{}l}{\textbf{PLB / FEP+ 4}}\\
\hspace{0.4em}\textit{compound-weighted} & 87 & \phantom{-}\textbf{0.86}{\scriptsize$\pm$0.03} & \phantom{-}0.77{\scriptsize$\pm$0.04} & \phantom{-}0.81{\scriptsize$\pm$0.04} & \phantom{-}0.66{\scriptsize$\pm$0.07} & \phantom{-}0.77{\scriptsize$\pm$0.04} & \phantom{-}0.54{\scriptsize$\pm$0.09} & \phantom{-}0.41{\scriptsize$\pm$0.10} & \phantom{-}0.50{\scriptsize$\pm$0.08} & \phantom{-}0.31{\scriptsize$\pm$0.10}\\
\hspace{0.4em}\texttt{\scriptsize cdk2} & 16 & \phantom{-}\textbf{0.94}{\scriptsize$\pm$0.02} & \phantom{-}0.65{\scriptsize$\pm$0.14} & \phantom{-}0.82{\scriptsize$\pm$0.09} & \phantom{-}0.77{\scriptsize$\pm$0.08} & \phantom{-}0.83{\scriptsize$\pm$0.06} & \phantom{-}0.82{\scriptsize$\pm$0.08} & \phantom{-}0.48{\scriptsize$\pm$0.19} & \phantom{-}0.59{\scriptsize$\pm$0.13} & \phantom{-}0.73{\scriptsize$\pm$0.09}\\
\hspace{0.4em}\texttt{\scriptsize jnk1} & 21 & \phantom{-}\textbf{0.92}{\scriptsize$\pm$0.04} & \phantom{-}0.84{\scriptsize$\pm$0.05} & \phantom{-}0.64{\scriptsize$\pm$0.11} & \phantom{-}0.76{\scriptsize$\pm$0.07} & \phantom{-}0.81{\scriptsize$\pm$0.05} & \phantom{-}0.36{\scriptsize$\pm$0.23} & \phantom{-}0.39{\scriptsize$\pm$0.26} & \phantom{-}0.53{\scriptsize$\pm$0.18} & -0.16{\scriptsize$\pm$0.26}\\
\hspace{0.4em}\texttt{\scriptsize p38} & 34 & \phantom{-}0.79{\scriptsize$\pm$0.07} & \phantom{-}0.70{\scriptsize$\pm$0.08} & \phantom{-}\textbf{0.87}{\scriptsize$\pm$0.04} & \phantom{-}0.43{\scriptsize$\pm$0.16} & \phantom{-}0.68{\scriptsize$\pm$0.08} & \phantom{-}0.40{\scriptsize$\pm$0.16} & \phantom{-}0.56{\scriptsize$\pm$0.11} & \phantom{-}0.55{\scriptsize$\pm$0.10} & \phantom{-}0.58{\scriptsize$\pm$0.12}\\
\hspace{0.4em}\texttt{\scriptsize tyk2} & 16 & \phantom{-}0.84{\scriptsize$\pm$0.08} & \phantom{-}\textbf{0.93}{\scriptsize$\pm$0.03} & \phantom{-}0.87{\scriptsize$\pm$0.08} & \phantom{-}0.89{\scriptsize$\pm$0.04} & \phantom{-}0.83{\scriptsize$\pm$0.09} & \phantom{-}0.81{\scriptsize$\pm$0.09} & \phantom{-}0.05{\scriptsize$\pm$0.25} & \phantom{-}0.23{\scriptsize$\pm$0.26} & -0.05{\scriptsize$\pm$0.29}\\
\midrule
\multicolumn{11}{@{}l}{\textbf{CASP16}}\\
\hspace{0.4em}\textit{compound-weighted} & 140 & \phantom{-}\textbf{0.77}{\scriptsize$\pm$0.03} & -- & \phantom{-}0.65{\scriptsize$\pm$0.05} & -- & \phantom{-}0.59{\scriptsize$\pm$0.05} & \phantom{-}0.56{\scriptsize$\pm$0.06} & \phantom{-}0.56{\scriptsize$\pm$0.05} & \phantom{-}0.49{\scriptsize$\pm$0.06} & --\\
\hspace{0.4em}\texttt{\scriptsize autotaxin} & 123 & \phantom{-}\textbf{0.80}{\scriptsize$\pm$0.03} & -- & \phantom{-}0.69{\scriptsize$\pm$0.05} & -- & \phantom{-}0.60{\scriptsize$\pm$0.06} & \phantom{-}0.58{\scriptsize$\pm$0.06} & \phantom{-}0.60{\scriptsize$\pm$0.06} & \phantom{-}0.56{\scriptsize$\pm$0.05} & --\\
\hspace{0.4em}\texttt{\scriptsize chymase} & 17 & \phantom{-}\textbf{0.55}{\scriptsize$\pm$0.10} & -- & \phantom{-}0.41{\scriptsize$\pm$0.27} & -- & \phantom{-}0.54{\scriptsize$\pm$0.19} & \phantom{-}0.37{\scriptsize$\pm$0.21} & \phantom{-}0.30{\scriptsize$\pm$0.13} & \phantom{-}0.00{\scriptsize$\pm$0.23} & --\\
\end{longtable}
}
\endgroup
\begingroup\def\pptablecaption{Additional public-potency metrics: Spearman $\rho$ for OpenBind EV-A71 2A, as shown in Figure~\ref{fig:series}, and Kendall $\tau$ for \caspsixteen. The \caspsixteen mean weights targets by compound count, and its target rows show autotaxin and chymase separately. Values carry one bootstrap standard error where per-compound predictions are available. OpenBind Nesso-1 and ApherisAffinity are cited aggregates without matched errors.
Its MolWt, GNINA, Boltz-2 and AQAffinity values use published per-compound predictions, and LigUnity was run on the same cohort. OpenBind uses three decimals and \caspsixteen two.}%
{\footnotesize
\setlength{\tabcolsep}{1.2pt}
\begin{longtable}{@{}lrrrrrrrrr@{}}
\caption{\pptablecaption{} \statustag{series}}\label{tab:a2secondary}\\
\toprule
Series / target & $n$ & \multicolumn{1}{c}{\rotatebox{70}{\ruby}} & \multicolumn{1}{c}{\rotatebox{70}{Nesso-1}} & \multicolumn{1}{c}{\rotatebox{70}{Boltz-2}} & \multicolumn{1}{c}{\rotatebox{70}{LigUnity}} & \multicolumn{1}{c}{\rotatebox{70}{Mol. weight}} & \multicolumn{1}{c}{\rotatebox{70}{AQAffinity}} & \multicolumn{1}{c}{\rotatebox{70}{GNINA}} & \multicolumn{1}{c}{\rotatebox{70}{ApherisAffinity}}\\
\midrule
\endfirsthead
\multicolumn{10}{@{}l}{\textit{continued}}\\
\toprule
Series / target & $n$ & \multicolumn{1}{c}{\rotatebox{70}{\ruby}} & \multicolumn{1}{c}{\rotatebox{70}{Nesso-1}} & \multicolumn{1}{c}{\rotatebox{70}{Boltz-2}} & \multicolumn{1}{c}{\rotatebox{70}{LigUnity}} & \multicolumn{1}{c}{\rotatebox{70}{Mol. weight}} & \multicolumn{1}{c}{\rotatebox{70}{AQAffinity}} & \multicolumn{1}{c}{\rotatebox{70}{GNINA}} & \multicolumn{1}{c}{\rotatebox{70}{ApherisAffinity}}\\
\midrule
\endhead
\bottomrule
\endfoot
\multicolumn{10}{@{}l}{\textbf{OpenBind EV-A71 2A}}\\*
\hspace{0.4em}EV-A71 2A & 494 & \textbf{0.574}{\scriptsize$\pm$0.034} & 0.512{\scriptsize\phantom{$\pm$0.000}} & 0.403{\scriptsize$\pm$0.038} & 0.112{\scriptsize$\pm$0.044} & 0.487{\scriptsize$\pm$0.036} & 0.114{\scriptsize$\pm$0.044} & 0.454{\scriptsize$\pm$0.037} & 0.537{\scriptsize\phantom{$\pm$0.000}}\\
\midrule
\multicolumn{10}{@{}l}{\textbf{CASP16}}\\*
\hspace{0.4em}\textit{compound-weighted} & 140 & \textbf{0.57}\phantom{0}{\scriptsize$\pm$0.04\phantom{0}} & 0.47\phantom{0}{\scriptsize$\pm$0.05\phantom{0}} & 0.38\phantom{0}{\scriptsize$\pm$0.05\phantom{0}} & 0.39\phantom{0}{\scriptsize$\pm$0.05\phantom{0}} & 0.42\phantom{0}{\scriptsize$\pm$0.05\phantom{0}} & 0.35\phantom{0}{\scriptsize$\pm$0.05\phantom{0}} & -- & --\\
\hspace{0.4em}\texttt{\scriptsize autotaxin} & 123 & \textbf{0.60}\phantom{0}{\scriptsize$\pm$0.04\phantom{0}} & 0.49\phantom{0}{\scriptsize$\pm$0.04\phantom{0}} & 0.40\phantom{0}{\scriptsize$\pm$0.05\phantom{0}} & 0.42\phantom{0}{\scriptsize$\pm$0.05\phantom{0}} & 0.42\phantom{0}{\scriptsize$\pm$0.05\phantom{0}} & 0.39\phantom{0}{\scriptsize$\pm$0.05\phantom{0}} & -- & --\\
\hspace{0.4em}\texttt{\scriptsize chymase} & 17 & \textbf{0.41}\phantom{0}{\scriptsize$\pm$0.14\phantom{0}} & 0.26\phantom{0}{\scriptsize$\pm$0.19\phantom{0}} & 0.25\phantom{0}{\scriptsize$\pm$0.19\phantom{0}} & 0.18\phantom{0}{\scriptsize$\pm$0.20\phantom{0}} & 0.40\phantom{0}{\scriptsize$\pm$0.14\phantom{0}} & 0.07\phantom{0}{\scriptsize$\pm$0.20\phantom{0}} & -- & --\\
\end{longtable}
}
\endgroup

\FloatBarrier

\Needspace{.45\textheight}
\section{Partner-program analyses}
\label{app:partner-analyses}

\subsection{Per-assay performance}
\begingroup\def\pptablecaption{Per-assay performance on the eight \partnercohort assays underlying Figure~\ref{fig:programs}, with methods in columns and metrics in rows.
The first row group gives the equally weighted assay average, with $n$ summed across assays.
Each setting reports Spearman $\rho$, Pearson $r$, and Kendall $\tau_b$, with MAE in \pX units additionally reported after recalibration.
The $n$ column distinguishes the full zero-shot cohort, the shared fitted cohort for recalibrated correlations, and eligible exact measurements for MAE.
Values show point estimates and one standard error from 1{,}000 within-assay compound bootstrap resamples (seed 0), rounded to two decimals.
Assay-average errors use the Figure~\ref{fig:programs} bootstrap estimates where available, with independent assay errors propagated for Kendall $\tau_b$ and molecular weight.
Recalibrated correlations use pooled rolling predictions for every method.}%
{\scriptsize
\setlength{\tabcolsep}{3.5pt}
\begin{longtable}{@{}l@{\hspace{6pt}}l@{\hspace{7pt}}l@{\hspace{7pt}}r@{\hspace{10pt}}r@{\hspace{11pt}}r@{\hspace{11pt}}r@{\hspace{11pt}}r@{\hspace{11pt}}r@{\hspace{11pt}}r@{\hspace{11pt}}r@{}}
\caption{\pptablecaption{} \statustag{incyte_detail}}\label{tab:incyte-detail}\\
\toprule
\shortstack[l]{Program\\task} & Mode & Metric & $n$ & \multicolumn{1}{c}{\rotatebox{70}{\ruby}} & \multicolumn{1}{c}{\rotatebox{70}{Nesso-1}} & \multicolumn{1}{c}{\rotatebox{70}{Boltz-2}} & \multicolumn{1}{c}{\rotatebox{70}{GNINA}} & \multicolumn{1}{c}{\rotatebox{70}{LigUnity}} & \multicolumn{1}{c}{\rotatebox{70}{AQAffinity}} & \multicolumn{1}{c}{\rotatebox{70}{Mol. weight}}\\
\midrule
\endfirsthead
\multicolumn{11}{@{}l}{\textit{continued}}\\
\toprule
\shortstack[l]{Program\\task} & Mode & Metric & $n$ & \multicolumn{1}{c}{\rotatebox{70}{\ruby}} & \multicolumn{1}{c}{\rotatebox{70}{Nesso-1}} & \multicolumn{1}{c}{\rotatebox{70}{Boltz-2}} & \multicolumn{1}{c}{\rotatebox{70}{GNINA}} & \multicolumn{1}{c}{\rotatebox{70}{LigUnity}} & \multicolumn{1}{c}{\rotatebox{70}{AQAffinity}} & \multicolumn{1}{c}{\rotatebox{70}{Mol. weight}}\\
\midrule
\endhead
\bottomrule
\endfoot
\noalign{\needspace{9\baselineskip}}
\multirow{7}{1.35cm}{\raggedright\textbf{\shortstack[l]{Assay\\average}{}}} & \multirow{3}{*}{Zero-shot} & $\rho$ & 22{,}036 & \textbf{0.61}{\scriptsize$\pm$0.01} & 0.39{\scriptsize$\pm$0.01} & 0.46{\scriptsize$\pm$0.01} & \phantom{-}0.19{\scriptsize$\pm$0.01} & \phantom{-}0.11{\scriptsize$\pm$0.01} & 0.30{\scriptsize$\pm$0.01} & 0.43{\scriptsize$\pm$0.01}\\*
 &  & $r$ & 22{,}036 & \textbf{0.59}{\scriptsize$\pm$0.01} & 0.36{\scriptsize$\pm$0.01} & 0.45{\scriptsize$\pm$0.01} & \phantom{-}0.19{\scriptsize$\pm$0.01} & \phantom{-}0.11{\scriptsize$\pm$0.01} & 0.29{\scriptsize$\pm$0.01} & 0.41{\scriptsize$\pm$0.01}\\*
 &  & $\tau$ & 22{,}036 & \textbf{0.44}{\scriptsize$\pm$0.00} & 0.27{\scriptsize$\pm$0.01} & 0.32{\scriptsize$\pm$0.01} & \phantom{-}0.13{\scriptsize$\pm$0.01} & \phantom{-}0.08{\scriptsize$\pm$0.01} & 0.21{\scriptsize$\pm$0.01} & 0.30{\scriptsize$\pm$0.01}\\*
\cmidrule(lr){2-11}
 & \multirow{4}{*}{Recalibrated} & $\rho$ & 20{,}619 & \textbf{0.71}{\scriptsize$\pm$0.01} & 0.44{\scriptsize$\pm$0.01} & 0.51{\scriptsize$\pm$0.01} & \phantom{-}0.39{\scriptsize$\pm$0.01} & \phantom{-}0.41{\scriptsize$\pm$0.01} & 0.45{\scriptsize$\pm$0.01} & 0.49{\scriptsize$\pm$0.01}\\*
 &  & $r$ & 20{,}619 & \textbf{0.70}{\scriptsize$\pm$0.01} & 0.42{\scriptsize$\pm$0.01} & 0.49{\scriptsize$\pm$0.01} & \phantom{-}0.36{\scriptsize$\pm$0.01} & \phantom{-}0.40{\scriptsize$\pm$0.01} & 0.42{\scriptsize$\pm$0.01} & 0.47{\scriptsize$\pm$0.01}\\*
 &  & $\tau$ & 20{,}619 & \textbf{0.53}{\scriptsize$\pm$0.01} & 0.31{\scriptsize$\pm$0.01} & 0.36{\scriptsize$\pm$0.01} & \phantom{-}0.27{\scriptsize$\pm$0.01} & \phantom{-}0.29{\scriptsize$\pm$0.01} & 0.31{\scriptsize$\pm$0.01} & 0.34{\scriptsize$\pm$0.01}\\*
 &  & MAE & 17{,}631 & \textbf{0.53}{\scriptsize$\pm$0.01} & 0.76{\scriptsize$\pm$0.01} & 0.74{\scriptsize$\pm$0.01} & \phantom{-}0.81{\scriptsize$\pm$0.01} & \phantom{-}0.81{\scriptsize$\pm$0.01} & 0.80{\scriptsize$\pm$0.01} & 0.77{\scriptsize$\pm$0.01}\\
\midrule
\noalign{\needspace{9\baselineskip}}
\multirow{7}{1.35cm}{\raggedright\textbf{\shortstack[l]{Incyte-A\\biochem}{}}} & \multirow{3}{*}{Zero-shot} & $\rho$ & 3{,}560 & \textbf{0.73}{\scriptsize$\pm$0.01} & 0.61{\scriptsize$\pm$0.01} & 0.63{\scriptsize$\pm$0.01} & -0.01{\scriptsize$\pm$0.02} & -0.23{\scriptsize$\pm$0.02} & 0.34{\scriptsize$\pm$0.01} & 0.41{\scriptsize$\pm$0.02}\\*
 &  & $r$ & 3{,}560 & \textbf{0.72}{\scriptsize$\pm$0.01} & 0.60{\scriptsize$\pm$0.01} & 0.62{\scriptsize$\pm$0.01} & -0.01{\scriptsize$\pm$0.02} & -0.19{\scriptsize$\pm$0.02} & 0.35{\scriptsize$\pm$0.01} & 0.41{\scriptsize$\pm$0.02}\\*
 &  & $\tau$ & 3{,}560 & \textbf{0.54}{\scriptsize$\pm$0.01} & 0.43{\scriptsize$\pm$0.01} & 0.45{\scriptsize$\pm$0.01} & -0.01{\scriptsize$\pm$0.01} & -0.15{\scriptsize$\pm$0.01} & 0.23{\scriptsize$\pm$0.01} & 0.29{\scriptsize$\pm$0.01}\\*
\cmidrule(lr){2-11}
 & \multirow{4}{*}{Recalibrated} & $\rho$ & 3{,}387 & \textbf{0.83}{\scriptsize$\pm$0.01} & 0.63{\scriptsize$\pm$0.01} & 0.67{\scriptsize$\pm$0.01} & \phantom{-}0.44{\scriptsize$\pm$0.02} & \phantom{-}0.49{\scriptsize$\pm$0.01} & 0.50{\scriptsize$\pm$0.01} & 0.57{\scriptsize$\pm$0.01}\\*
 &  & $r$ & 3{,}387 & \textbf{0.83}{\scriptsize$\pm$0.01} & 0.61{\scriptsize$\pm$0.01} & 0.67{\scriptsize$\pm$0.01} & \phantom{-}0.40{\scriptsize$\pm$0.02} & \phantom{-}0.47{\scriptsize$\pm$0.01} & 0.48{\scriptsize$\pm$0.01} & 0.55{\scriptsize$\pm$0.02}\\*
 &  & $\tau$ & 3{,}387 & \textbf{0.64}{\scriptsize$\pm$0.01} & 0.45{\scriptsize$\pm$0.01} & 0.49{\scriptsize$\pm$0.01} & \phantom{-}0.30{\scriptsize$\pm$0.01} & \phantom{-}0.34{\scriptsize$\pm$0.01} & 0.34{\scriptsize$\pm$0.01} & 0.41{\scriptsize$\pm$0.01}\\*
 &  & MAE & 3{,}001 & \textbf{0.68}{\scriptsize$\pm$0.01} & 1.04{\scriptsize$\pm$0.01} & 1.03{\scriptsize$\pm$0.01} & \phantom{-}1.30{\scriptsize$\pm$0.01} & \phantom{-}1.28{\scriptsize$\pm$0.01} & 1.23{\scriptsize$\pm$0.01} & 1.20{\scriptsize$\pm$0.01}\\
\midrule
\noalign{\needspace{9\baselineskip}}
\multirow{7}{1.35cm}{\raggedright\textbf{\shortstack[l]{Incyte-A\\cellular}{}}} & \multirow{3}{*}{Zero-shot} & $\rho$ & 3{,}360 & \textbf{0.54}{\scriptsize$\pm$0.01} & 0.36{\scriptsize$\pm$0.02} & 0.40{\scriptsize$\pm$0.01} & \phantom{-}0.03{\scriptsize$\pm$0.02} & -0.30{\scriptsize$\pm$0.02} & 0.20{\scriptsize$\pm$0.02} & 0.21{\scriptsize$\pm$0.02}\\*
 &  & $r$ & 3{,}360 & \textbf{0.52}{\scriptsize$\pm$0.01} & 0.38{\scriptsize$\pm$0.02} & 0.36{\scriptsize$\pm$0.01} & \phantom{-}0.02{\scriptsize$\pm$0.02} & -0.28{\scriptsize$\pm$0.02} & 0.20{\scriptsize$\pm$0.02} & 0.21{\scriptsize$\pm$0.02}\\*
 &  & $\tau$ & 3{,}360 & \textbf{0.37}{\scriptsize$\pm$0.01} & 0.25{\scriptsize$\pm$0.01} & 0.27{\scriptsize$\pm$0.01} & \phantom{-}0.02{\scriptsize$\pm$0.01} & -0.20{\scriptsize$\pm$0.01} & 0.14{\scriptsize$\pm$0.01} & 0.14{\scriptsize$\pm$0.01}\\*
\cmidrule(lr){2-11}
 & \multirow{4}{*}{Recalibrated} & $\rho$ & 3{,}260 & \textbf{0.81}{\scriptsize$\pm$0.01} & 0.63{\scriptsize$\pm$0.01} & 0.65{\scriptsize$\pm$0.01} & \phantom{-}0.60{\scriptsize$\pm$0.01} & \phantom{-}0.61{\scriptsize$\pm$0.01} & 0.61{\scriptsize$\pm$0.01} & 0.63{\scriptsize$\pm$0.01}\\*
 &  & $r$ & 3{,}260 & \textbf{0.81}{\scriptsize$\pm$0.01} & 0.62{\scriptsize$\pm$0.01} & 0.64{\scriptsize$\pm$0.01} & \phantom{-}0.59{\scriptsize$\pm$0.01} & \phantom{-}0.60{\scriptsize$\pm$0.01} & 0.60{\scriptsize$\pm$0.01} & 0.62{\scriptsize$\pm$0.01}\\*
 &  & $\tau$ & 3{,}260 & \textbf{0.61}{\scriptsize$\pm$0.01} & 0.45{\scriptsize$\pm$0.01} & 0.46{\scriptsize$\pm$0.01} & \phantom{-}0.43{\scriptsize$\pm$0.01} & \phantom{-}0.44{\scriptsize$\pm$0.01} & 0.44{\scriptsize$\pm$0.01} & 0.44{\scriptsize$\pm$0.01}\\*
 &  & MAE & 3{,}109 & \textbf{0.48}{\scriptsize$\pm$0.01} & 0.80{\scriptsize$\pm$0.01} & 0.80{\scriptsize$\pm$0.01} & \phantom{-}0.86{\scriptsize$\pm$0.01} & \phantom{-}0.84{\scriptsize$\pm$0.01} & 0.85{\scriptsize$\pm$0.01} & 0.85{\scriptsize$\pm$0.01}\\
\midrule
\noalign{\needspace{9\baselineskip}}
\multirow{7}{1.35cm}{\raggedright\textbf{\shortstack[l]{Incyte-B\\assay 1}{}}} & \multirow{3}{*}{Zero-shot} & $\rho$ & 1{,}418 & \textbf{0.69}{\scriptsize$\pm$0.01} & 0.45{\scriptsize$\pm$0.02} & 0.42{\scriptsize$\pm$0.02} & \phantom{-}0.22{\scriptsize$\pm$0.03} & \phantom{-}0.19{\scriptsize$\pm$0.02} & 0.46{\scriptsize$\pm$0.02} & 0.43{\scriptsize$\pm$0.02}\\*
 &  & $r$ & 1{,}418 & \textbf{0.63}{\scriptsize$\pm$0.01} & 0.42{\scriptsize$\pm$0.02} & 0.41{\scriptsize$\pm$0.02} & \phantom{-}0.27{\scriptsize$\pm$0.02} & \phantom{-}0.13{\scriptsize$\pm$0.02} & 0.45{\scriptsize$\pm$0.02} & 0.41{\scriptsize$\pm$0.02}\\*
 &  & $\tau$ & 1{,}418 & \textbf{0.50}{\scriptsize$\pm$0.01} & 0.32{\scriptsize$\pm$0.02} & 0.31{\scriptsize$\pm$0.02} & \phantom{-}0.16{\scriptsize$\pm$0.02} & \phantom{-}0.13{\scriptsize$\pm$0.02} & 0.33{\scriptsize$\pm$0.02} & 0.31{\scriptsize$\pm$0.02}\\*
\cmidrule(lr){2-11}
 & \multirow{4}{*}{Recalibrated} & $\rho$ & 1{,}166 & \textbf{0.54}{\scriptsize$\pm$0.02} & 0.40{\scriptsize$\pm$0.03} & 0.38{\scriptsize$\pm$0.03} & \phantom{-}0.27{\scriptsize$\pm$0.03} & \phantom{-}0.23{\scriptsize$\pm$0.03} & 0.40{\scriptsize$\pm$0.03} & 0.35{\scriptsize$\pm$0.03}\\*
 &  & $r$ & 1{,}166 & \textbf{0.53}{\scriptsize$\pm$0.02} & 0.34{\scriptsize$\pm$0.02} & 0.31{\scriptsize$\pm$0.02} & \phantom{-}0.19{\scriptsize$\pm$0.03} & \phantom{-}0.20{\scriptsize$\pm$0.03} & 0.33{\scriptsize$\pm$0.03} & 0.29{\scriptsize$\pm$0.03}\\*
 &  & $\tau$ & 1{,}166 & \textbf{0.39}{\scriptsize$\pm$0.02} & 0.28{\scriptsize$\pm$0.02} & 0.27{\scriptsize$\pm$0.02} & \phantom{-}0.18{\scriptsize$\pm$0.02} & \phantom{-}0.16{\scriptsize$\pm$0.02} & 0.28{\scriptsize$\pm$0.02} & 0.25{\scriptsize$\pm$0.02}\\*
 &  & MAE & 846 & \textbf{0.57}{\scriptsize$\pm$0.02} & 0.71{\scriptsize$\pm$0.02} & 0.71{\scriptsize$\pm$0.02} & \phantom{-}0.73{\scriptsize$\pm$0.02} & \phantom{-}0.76{\scriptsize$\pm$0.02} & 0.70{\scriptsize$\pm$0.02} & 0.72{\scriptsize$\pm$0.02}\\
\midrule
\noalign{\needspace{9\baselineskip}}
\multirow{7}{1.35cm}{\raggedright\textbf{\shortstack[l]{Incyte-B\\assay 2}{}}} & \multirow{3}{*}{Zero-shot} & $\rho$ & 1{,}418 & \textbf{0.72}{\scriptsize$\pm$0.01} & 0.49{\scriptsize$\pm$0.02} & 0.47{\scriptsize$\pm$0.02} & \phantom{-}0.29{\scriptsize$\pm$0.03} & \phantom{-}0.24{\scriptsize$\pm$0.02} & 0.49{\scriptsize$\pm$0.02} & 0.51{\scriptsize$\pm$0.02}\\*
 &  & $r$ & 1{,}418 & \textbf{0.63}{\scriptsize$\pm$0.01} & 0.43{\scriptsize$\pm$0.02} & 0.42{\scriptsize$\pm$0.02} & \phantom{-}0.29{\scriptsize$\pm$0.02} & \phantom{-}0.19{\scriptsize$\pm$0.02} & 0.44{\scriptsize$\pm$0.02} & 0.44{\scriptsize$\pm$0.02}\\*
 &  & $\tau$ & 1{,}418 & \textbf{0.53}{\scriptsize$\pm$0.01} & 0.35{\scriptsize$\pm$0.01} & 0.34{\scriptsize$\pm$0.02} & \phantom{-}0.21{\scriptsize$\pm$0.02} & \phantom{-}0.17{\scriptsize$\pm$0.02} & 0.34{\scriptsize$\pm$0.01} & 0.36{\scriptsize$\pm$0.02}\\*
\cmidrule(lr){2-11}
 & \multirow{4}{*}{Recalibrated} & $\rho$ & 1{,}112 & \textbf{0.60}{\scriptsize$\pm$0.02} & 0.43{\scriptsize$\pm$0.03} & 0.40{\scriptsize$\pm$0.03} & \phantom{-}0.29{\scriptsize$\pm$0.03} & \phantom{-}0.31{\scriptsize$\pm$0.03} & 0.39{\scriptsize$\pm$0.03} & 0.39{\scriptsize$\pm$0.03}\\*
 &  & $r$ & 1{,}112 & \textbf{0.56}{\scriptsize$\pm$0.02} & 0.33{\scriptsize$\pm$0.03} & 0.31{\scriptsize$\pm$0.03} & \phantom{-}0.21{\scriptsize$\pm$0.03} & \phantom{-}0.26{\scriptsize$\pm$0.03} & 0.31{\scriptsize$\pm$0.03} & 0.31{\scriptsize$\pm$0.03}\\*
 &  & $\tau$ & 1{,}112 & \textbf{0.44}{\scriptsize$\pm$0.02} & 0.30{\scriptsize$\pm$0.02} & 0.28{\scriptsize$\pm$0.02} & \phantom{-}0.19{\scriptsize$\pm$0.02} & \phantom{-}0.21{\scriptsize$\pm$0.02} & 0.27{\scriptsize$\pm$0.02} & 0.27{\scriptsize$\pm$0.02}\\*
 &  & MAE & 837 & \textbf{0.45}{\scriptsize$\pm$0.01} & 0.56{\scriptsize$\pm$0.01} & 0.57{\scriptsize$\pm$0.01} & \phantom{-}0.58{\scriptsize$\pm$0.02} & \phantom{-}0.58{\scriptsize$\pm$0.01} & 0.57{\scriptsize$\pm$0.01} & 0.57{\scriptsize$\pm$0.01}\\
\midrule
\noalign{\needspace{9\baselineskip}}
\multirow{7}{1.35cm}{\raggedright\textbf{Incyte-C{}}} & \multirow{3}{*}{Zero-shot} & $\rho$ & 2{,}302 & \textbf{0.41}{\scriptsize$\pm$0.02} & 0.24{\scriptsize$\pm$0.02} & 0.34{\scriptsize$\pm$0.02} & \phantom{-}0.07{\scriptsize$\pm$0.02} & \phantom{-}0.26{\scriptsize$\pm$0.02} & 0.31{\scriptsize$\pm$0.02} & 0.34{\scriptsize$\pm$0.02}\\*
 &  & $r$ & 2{,}302 & \textbf{0.43}{\scriptsize$\pm$0.02} & 0.23{\scriptsize$\pm$0.02} & 0.33{\scriptsize$\pm$0.02} & \phantom{-}0.10{\scriptsize$\pm$0.02} & \phantom{-}0.26{\scriptsize$\pm$0.02} & 0.30{\scriptsize$\pm$0.02} & 0.31{\scriptsize$\pm$0.02}\\*
 &  & $\tau$ & 2{,}302 & \textbf{0.28}{\scriptsize$\pm$0.01} & 0.16{\scriptsize$\pm$0.01} & 0.23{\scriptsize$\pm$0.01} & \phantom{-}0.05{\scriptsize$\pm$0.02} & \phantom{-}0.19{\scriptsize$\pm$0.01} & 0.22{\scriptsize$\pm$0.01} & 0.24{\scriptsize$\pm$0.01}\\*
\cmidrule(lr){2-11}
 & \multirow{4}{*}{Recalibrated} & $\rho$ & 2{,}153 & \textbf{0.70}{\scriptsize$\pm$0.01} & 0.34{\scriptsize$\pm$0.02} & 0.43{\scriptsize$\pm$0.02} & \phantom{-}0.33{\scriptsize$\pm$0.02} & \phantom{-}0.37{\scriptsize$\pm$0.02} & 0.38{\scriptsize$\pm$0.02} & 0.43{\scriptsize$\pm$0.02}\\*
 &  & $r$ & 2{,}153 & \textbf{0.71}{\scriptsize$\pm$0.01} & 0.33{\scriptsize$\pm$0.02} & 0.45{\scriptsize$\pm$0.02} & \phantom{-}0.34{\scriptsize$\pm$0.02} & \phantom{-}0.39{\scriptsize$\pm$0.02} & 0.39{\scriptsize$\pm$0.02} & 0.43{\scriptsize$\pm$0.02}\\*
 &  & $\tau$ & 2{,}153 & \textbf{0.52}{\scriptsize$\pm$0.01} & 0.23{\scriptsize$\pm$0.01} & 0.30{\scriptsize$\pm$0.01} & \phantom{-}0.22{\scriptsize$\pm$0.01} & \phantom{-}0.26{\scriptsize$\pm$0.01} & 0.26{\scriptsize$\pm$0.01} & 0.30{\scriptsize$\pm$0.01}\\*
 &  & MAE & 1{,}714 & \textbf{0.60}{\scriptsize$\pm$0.01} & 0.79{\scriptsize$\pm$0.01} & 0.77{\scriptsize$\pm$0.01} & \phantom{-}0.79{\scriptsize$\pm$0.01} & \phantom{-}0.78{\scriptsize$\pm$0.01} & 0.76{\scriptsize$\pm$0.01} & 0.78{\scriptsize$\pm$0.01}\\
\midrule
\noalign{\needspace{9\baselineskip}}
\multirow{7}{1.35cm}{\raggedright\textbf{Incyte-D{}}} & \multirow{3}{*}{Zero-shot} & $\rho$ & 451 & 0.42{\scriptsize$\pm$0.04} & 0.16{\scriptsize$\pm$0.05} & 0.35{\scriptsize$\pm$0.04} & \phantom{-}0.24{\scriptsize$\pm$0.05} & \phantom{-}0.28{\scriptsize$\pm$0.04} & 0.35{\scriptsize$\pm$0.04} & \textbf{0.45}{\scriptsize$\pm$0.04}\\*
 &  & $r$ & 451 & \textbf{0.45}{\scriptsize$\pm$0.03} & 0.08{\scriptsize$\pm$0.05} & 0.37{\scriptsize$\pm$0.04} & \phantom{-}0.23{\scriptsize$\pm$0.04} & \phantom{-}0.29{\scriptsize$\pm$0.04} & 0.34{\scriptsize$\pm$0.04} & \textbf{0.45}{\scriptsize$\pm$0.04}\\*
 &  & $\tau$ & 451 & 0.29{\scriptsize$\pm$0.03} & 0.11{\scriptsize$\pm$0.03} & 0.25{\scriptsize$\pm$0.03} & \phantom{-}0.17{\scriptsize$\pm$0.03} & \phantom{-}0.19{\scriptsize$\pm$0.03} & 0.24{\scriptsize$\pm$0.03} & \textbf{0.32}{\scriptsize$\pm$0.03}\\*
\cmidrule(lr){2-11}
 & \multirow{4}{*}{Recalibrated} & $\rho$ & 318 & \textbf{0.65}{\scriptsize$\pm$0.04} & 0.09{\scriptsize$\pm$0.05} & 0.35{\scriptsize$\pm$0.05} & \phantom{-}0.25{\scriptsize$\pm$0.05} & \phantom{-}0.28{\scriptsize$\pm$0.05} & 0.31{\scriptsize$\pm$0.05} & 0.44{\scriptsize$\pm$0.05}\\*
 &  & $r$ & 318 & \textbf{0.66}{\scriptsize$\pm$0.03} & 0.08{\scriptsize$\pm$0.05} & 0.33{\scriptsize$\pm$0.04} & \phantom{-}0.21{\scriptsize$\pm$0.05} & \phantom{-}0.26{\scriptsize$\pm$0.05} & 0.28{\scriptsize$\pm$0.05} & 0.44{\scriptsize$\pm$0.04}\\*
 &  & $\tau$ & 318 & \textbf{0.48}{\scriptsize$\pm$0.03} & 0.07{\scriptsize$\pm$0.03} & 0.24{\scriptsize$\pm$0.03} & \phantom{-}0.17{\scriptsize$\pm$0.03} & \phantom{-}0.19{\scriptsize$\pm$0.03} & 0.21{\scriptsize$\pm$0.03} & 0.30{\scriptsize$\pm$0.03}\\*
 &  & MAE & 232 & \textbf{0.55}{\scriptsize$\pm$0.03} & 0.85{\scriptsize$\pm$0.04} & 0.77{\scriptsize$\pm$0.03} & \phantom{-}0.78{\scriptsize$\pm$0.04} & \phantom{-}0.84{\scriptsize$\pm$0.04} & 0.80{\scriptsize$\pm$0.04} & 0.79{\scriptsize$\pm$0.03}\\
\midrule
\noalign{\needspace{9\baselineskip}}
\multirow{7}{1.35cm}{\raggedright\textbf{\shortstack[l]{Incyte-E\\biochem}{}}} & \multirow{3}{*}{Zero-shot} & $\rho$ & 5{,}181 & \textbf{0.69}{\scriptsize$\pm$0.01} & 0.43{\scriptsize$\pm$0.01} & 0.58{\scriptsize$\pm$0.01} & \phantom{-}0.31{\scriptsize$\pm$0.01} & \phantom{-}0.27{\scriptsize$\pm$0.01} & 0.13{\scriptsize$\pm$0.01} & 0.57{\scriptsize$\pm$0.01}\\*
 &  & $r$ & 5{,}181 & \textbf{0.69}{\scriptsize$\pm$0.01} & 0.41{\scriptsize$\pm$0.01} & 0.58{\scriptsize$\pm$0.01} & \phantom{-}0.30{\scriptsize$\pm$0.01} & \phantom{-}0.29{\scriptsize$\pm$0.01} & 0.16{\scriptsize$\pm$0.01} & 0.56{\scriptsize$\pm$0.01}\\*
 &  & $\tau$ & 5{,}181 & \textbf{0.50}{\scriptsize$\pm$0.01} & 0.30{\scriptsize$\pm$0.01} & 0.41{\scriptsize$\pm$0.01} & \phantom{-}0.21{\scriptsize$\pm$0.01} & \phantom{-}0.18{\scriptsize$\pm$0.01} & 0.09{\scriptsize$\pm$0.01} & 0.40{\scriptsize$\pm$0.01}\\*
\cmidrule(lr){2-11}
 & \multirow{4}{*}{Recalibrated} & $\rho$ & 4{,}994 & \textbf{0.76}{\scriptsize$\pm$0.01} & 0.55{\scriptsize$\pm$0.01} & 0.64{\scriptsize$\pm$0.01} & \phantom{-}0.55{\scriptsize$\pm$0.01} & \phantom{-}0.53{\scriptsize$\pm$0.01} & 0.52{\scriptsize$\pm$0.01} & 0.58{\scriptsize$\pm$0.01}\\*
 &  & $r$ & 4{,}994 & \textbf{0.76}{\scriptsize$\pm$0.01} & 0.56{\scriptsize$\pm$0.01} & 0.63{\scriptsize$\pm$0.01} & \phantom{-}0.52{\scriptsize$\pm$0.01} & \phantom{-}0.56{\scriptsize$\pm$0.01} & 0.54{\scriptsize$\pm$0.01} & 0.59{\scriptsize$\pm$0.01}\\*
 &  & $\tau$ & 4{,}994 & \textbf{0.57}{\scriptsize$\pm$0.01} & 0.38{\scriptsize$\pm$0.01} & 0.45{\scriptsize$\pm$0.01} & \phantom{-}0.38{\scriptsize$\pm$0.01} & \phantom{-}0.37{\scriptsize$\pm$0.01} & 0.36{\scriptsize$\pm$0.01} & 0.41{\scriptsize$\pm$0.01}\\*
 &  & MAE & 4{,}444 & \textbf{0.56}{\scriptsize$\pm$0.01} & 0.85{\scriptsize$\pm$0.01} & 0.77{\scriptsize$\pm$0.01} & \phantom{-}0.89{\scriptsize$\pm$0.01} & \phantom{-}0.91{\scriptsize$\pm$0.01} & 0.92{\scriptsize$\pm$0.01} & 0.78{\scriptsize$\pm$0.01}\\
\midrule
\noalign{\needspace{9\baselineskip}}
\multirow{7}{1.35cm}{\raggedright\textbf{\shortstack[l]{Incyte-E\\cellular}{}}} & \multirow{3}{*}{Zero-shot} & $\rho$ & 4{,}346 & \textbf{0.67}{\scriptsize$\pm$0.01} & 0.36{\scriptsize$\pm$0.01} & 0.49{\scriptsize$\pm$0.01} & \phantom{-}0.36{\scriptsize$\pm$0.01} & \phantom{-}0.18{\scriptsize$\pm$0.01} & 0.11{\scriptsize$\pm$0.02} & 0.50{\scriptsize$\pm$0.01}\\*
 &  & $r$ & 4{,}346 & \textbf{0.64}{\scriptsize$\pm$0.01} & 0.32{\scriptsize$\pm$0.01} & 0.48{\scriptsize$\pm$0.01} & \phantom{-}0.35{\scriptsize$\pm$0.01} & \phantom{-}0.19{\scriptsize$\pm$0.01} & 0.10{\scriptsize$\pm$0.02} & 0.48{\scriptsize$\pm$0.01}\\*
 &  & $\tau$ & 4{,}346 & \textbf{0.48}{\scriptsize$\pm$0.01} & 0.24{\scriptsize$\pm$0.01} & 0.34{\scriptsize$\pm$0.01} & \phantom{-}0.25{\scriptsize$\pm$0.01} & \phantom{-}0.12{\scriptsize$\pm$0.01} & 0.07{\scriptsize$\pm$0.01} & 0.35{\scriptsize$\pm$0.01}\\*
\cmidrule(lr){2-11}
 & \multirow{4}{*}{Recalibrated} & $\rho$ & 4{,}229 & \textbf{0.75}{\scriptsize$\pm$0.01} & 0.49{\scriptsize$\pm$0.01} & 0.59{\scriptsize$\pm$0.01} & \phantom{-}0.44{\scriptsize$\pm$0.01} & \phantom{-}0.48{\scriptsize$\pm$0.01} & 0.46{\scriptsize$\pm$0.01} & 0.54{\scriptsize$\pm$0.01}\\*
 &  & $r$ & 4{,}229 & \textbf{0.74}{\scriptsize$\pm$0.01} & 0.47{\scriptsize$\pm$0.01} & 0.56{\scriptsize$\pm$0.01} & \phantom{-}0.41{\scriptsize$\pm$0.01} & \phantom{-}0.45{\scriptsize$\pm$0.01} & 0.43{\scriptsize$\pm$0.01} & 0.52{\scriptsize$\pm$0.01}\\*
 &  & $\tau$ & 4{,}229 & \textbf{0.56}{\scriptsize$\pm$0.01} & 0.33{\scriptsize$\pm$0.01} & 0.41{\scriptsize$\pm$0.01} & \phantom{-}0.30{\scriptsize$\pm$0.01} & \phantom{-}0.32{\scriptsize$\pm$0.01} & 0.31{\scriptsize$\pm$0.01} & 0.38{\scriptsize$\pm$0.01}\\*
 &  & MAE & 3{,}448 & \textbf{0.36}{\scriptsize$\pm$0.01} & 0.50{\scriptsize$\pm$0.01} & 0.48{\scriptsize$\pm$0.01} & \phantom{-}0.51{\scriptsize$\pm$0.01} & \phantom{-}0.53{\scriptsize$\pm$0.01} & 0.53{\scriptsize$\pm$0.01} & 0.48{\scriptsize$\pm$0.01}\\
\end{longtable}
}
\endgroup

\subsection{Paired comparisons}
\begin{table}[htbp]
\centering\footnotesize
\caption{Paired bootstrap comparisons on eight partner assays.
Each assay receives equal weight, and compounds are resampled within assays on shared draws.
Zero-shot correlations use 22{,}036 scored registrations.
Recalibrated correlations use 20{,}619 shared fitted predictions, and MAE uses 17{,}631 eligible exact measurements.
The comparison method is selected separately for each metric by its aggregate point estimate.
Advantage is \ruby minus the comparison method for correlations and the comparison method minus \ruby for MAE.
Intervals are bootstrap percentiles, and tests condition on the selected \ruby readout and recalibration fits.
With 1{,}000 resamples, zero unfavorable draws are reported as $p<0.001$.}
\label{tab:partner-paired}
\begin{tabular}{@{}lllrrrl@{}}
\toprule
Mode & Metric & Comparison method & \ruby & Value & Advantage [95\% CI] & One-sided $p$\\
\midrule
Zero-shot & Spearman $\rho$ & Boltz-2 & 0.61 & 0.46 & 0.15 [0.13, 0.16] & $<0.001$\\
 & Pearson $r$ & Boltz-2 & 0.59 & 0.45 & 0.14 [0.13, 0.16] & $<0.001$\\
\midrule
Recalibrated & Spearman $\rho$ & Boltz-2 & 0.71 & 0.51 & 0.19 [0.18, 0.21] & $<0.001$\\
 & Pearson $r$ & Boltz-2 & 0.70 & 0.49 & 0.21 [0.20, 0.23] & $<0.001$\\
 & MAE (pX) & Boltz-2 & 0.53 & 0.74 & 0.20 [0.19, 0.21] & $<0.001$\\
\bottomrule
\end{tabular}
\end{table}

\FloatBarrier

\subsection{\partnercohort chemical similarity}
\label{app:partner-similarity}
For each scored rolling test compound, we take the maximum Tanimoto similarity of its nonchiral Morgan fingerprint (radius 3, 2{,}048 bits) to an earlier exactly measured compound from the same assay.
This temporal prefix describes novelty relative to earlier same-assay measurements, not to a comparator's pretraining corpus.
We use fixed 0.1-wide bins and display a bin only when all eight \partnercohort assays have at least six exact compounds.
Both MAE and Spearman $\rho$ use each method's rolling-recalibrated predictions.
We retain only records with finite fitted predictions from every method, matching the recalibrated cohort described in Appendix~\ref{app:partner-cohorts}.
MAE uses exact measurements, and Spearman additionally includes censored measurements at their reported bounds.
We pool predictions across all rolling test windows within each assay and similarity bin before computing metrics, then average assays equally.
The five bins spanning 0.4--0.9 meet the coverage criterion for all eight assays.

The bins are descriptive: the compounds and potency ranges differ between them.
The equal-assay curves are smoother than individual assays, which need not be monotone.
Within-bin $\rho$ also omits comparisons between bins, so it cannot be compared directly with the whole-cohort correlation in Figure~\ref{fig:programs}.

\FloatBarrier

\FloatBarrier

\subsection{Data scaling for fine-tuning}

\label{app:scaling}

We first tested whether updating the final potency-prediction layers of \ruby and the comparison models improved performance under the rolling-temporal evaluation used in the main paper.
For most model--assay combinations, fine-tuning did not consistently improve on the zero-shot model over the duration of each program.
Potency distributions shifted between the earlier validation data and later test compounds, and training settings selected on validation data did not reliably transfer to the subsequent test window.
We therefore use the more robust assay-specific recalibration for the main evaluation and examine fine-tuning separately under a split based on chemical clustering as a way to probe each model's internal generalization capabilities.

For the data-scaling analysis, we retain up to 1{,}000 compounds per assay, sampled evenly across its measurement history.

\begin{figure}[t!]
  \centering
  \includegraphics[width=\linewidth]{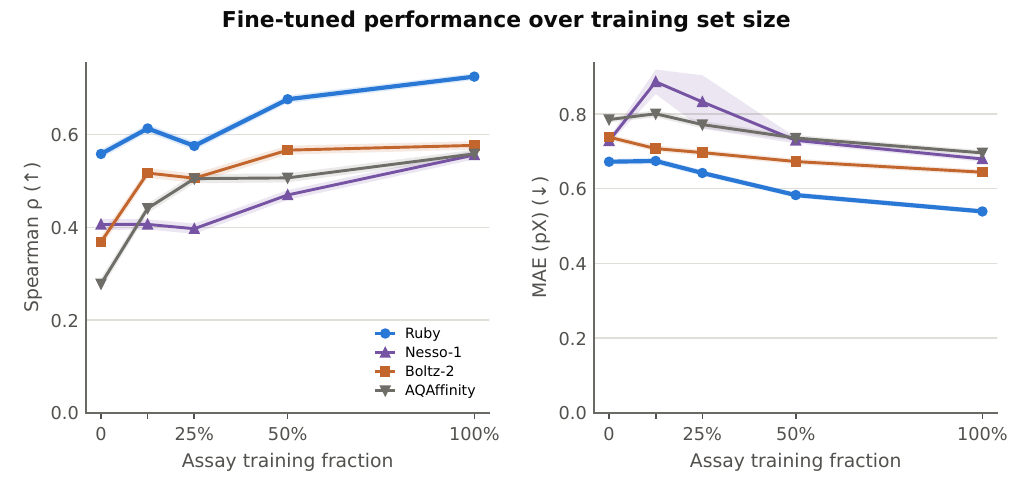}
  \caption{Data-scaling of assay fine-tuning, averaged with equal assay weight over the eight \partnercohort assays.
  Each model is fine-tuned on nested fractions of up to 1000 measurements from each assay training pool, so metrics are not directly comparable to those in Figure~\ref{fig:partner-similarity}. Shading represents $\pm$ one bootstrap standard error.
  \statustag{scaling}}
  \label{fig:scaling}
\end{figure}

\paragraph{Train--test split.}
For each assay, we group compounds by chemical similarity using Butina clustering with a Tanimoto threshold of 0.6 and assign the groups to four folds.
We use a 4-fold evaluation setting, testing on each fold and using the other three as the training pool.
Within each training pool, we repeat the clustering to create four inner folds for selecting training settings without using the test measurements.
We evaluate training budgets of 12.5\%, 25\%, 50\%, and 100\% of the available training pool, selecting settings separately at each budget.

\paragraph{Fine-tuning and model selection.}
For \ruby, \boltz{2}, \nesso, and \aqaffinity, we update the last one, two, or three layers of the potency prediction head while keeping the rest of the model fixed.
We precompute the inputs to these layers so that training does not require repeatedly running the full model.
For each model and data budget, we select the number of layers to update and the other training settings using the two-stage validation search in Table~\ref{tab:hparam-search}.
We rank configurations using a score that combines Spearman correlation, MAE, and average precision for identifying the highest-potency 20\% of compounds.
The three best configurations are refitted on the training pool at that data budget, and their predictions are averaged for the held-out test fold.

We combine held-out predictions across the four outer folds, compute metrics separately for each assay, and report the equally weighted mean across the eight assays.

\ruby outperforms every comparison method on both metrics at every training fraction, widening its margin as the fraction approaches 100\%.
These results suggest that \ruby may have a richer internal representation than the other models.

\begin{table}[!htbp]
\centering
\caption{Two-stage hyperparameter search for the affinity models.
Stage~1 is an exhaustive grid search.
Stage~2 varies one parameter at a time for the two best configurations per loss function rather than forming a second Cartesian product.
Both stages are scored only on inner-fold validation data, and the outer test fold is excluded from model selection.}
\label{tab:hparam-search}
\begin{minipage}[t]{0.48\linewidth}
\centering
\textbf{Stage 1 — Screening grid}\\[2pt]
\small
\begin{tabular}{@{}ll@{}}
\toprule
Axis & Values \\
\midrule
Last linear layers updated & $1,\ 2,\ 3$ \\
Learning rate              & $3\!\times\!10^{-4},\ 10^{-3},\ 3\!\times\!10^{-3}$ \\
Loss                       & MSE, Huber, pairwise \\
Relative penalty $\lambda$ & $0,\ 0.1$ \\
\midrule
Dropout                    & $0$ (fixed) \\
LR schedule                & constant (fixed) \\
Mean calibration           & off (fixed) \\
\midrule
Configurations             & $3\times3\times3\times2=54$ \\
Update checkpoints         & $10,25,50,100,200,400$ \\
Scored points              & $54\times6=324$ \\
\bottomrule
\end{tabular}
\end{minipage}
\hfill
\begin{minipage}[t]{0.48\linewidth}
\centering
\textbf{Stage 2 — Refinements}\\[2pt]
\small
\begin{tabular}{@{}ll@{}}
\toprule
Axis & Values \\
\midrule
Dropout                    & $0.1$ \\
LR schedule                & linear, cosine \\
Schedule horizon $H$       & $100,\ 400$ \\
Mean calibration           & on (MSE/Huber only) \\

\midrule
Configurations             & $34$ \\
Scored points              & $84$ \\
\midrule
\textbf{Total per fold}    & \textbf{88 cfg} \\
\bottomrule
\end{tabular}
\end{minipage}
\end{table}

\FloatBarrier

\end{document}